\documentclass{article} 
\usepackage{booktabs}  
\usepackage{tabularx}  
\usepackage{longtable} 
\usepackage{adjustbox}
\usepackage{hyperref}  
\usepackage{makecell} 
\usepackage{xcolor}  
\usepackage{pdflscape}
\usepackage{caption}
\usepackage{PRIMEarxiv}
\usepackage[utf8]{inputenc}
\usepackage[T1]{fontenc}
\usepackage{hyperref}
\usepackage{url}
\usepackage{rotating}
\usepackage{amsfonts}
\usepackage{nicefrac}
\usepackage{microtype}
\usepackage{fancyhdr}
\usepackage{graphicx}
\usepackage{lettrine} 
\usepackage{type1cm}
\usepackage{enumitem}

\title{From Pixels to Digital Agents: An Empirical Study on the Taxonomy and Technological Trends of Reinforcement Learning Environments}

\author{
  Lijing Luo \\
  Sun Yat-sen University \\
  Shenzhen, China\\
  \texttt{luolijingfuze@gmail.com} \\
  \And
  Yiben Luo \\
  Yancheng Institute of Technology \\
  Yancheng, China\\
  \texttt{yiben0011@163.com} \\
  \And
  Alexey Gorbatovski\\
  Central University \\
  Moscow, Russia\\
  \texttt{alexey.gorbatovski@gmail.com} \\
  \And
  Sergey Kovalchuk \\
  ITMO University \\
  Saint Petersburg, Russia\\
  \texttt{sergey.v.kovalchuk@gmail.com} \\
  \And
  Xiaodan Liang \\
  Sun Yat-sen University \\
  Shenzhen, China\\
  \texttt{xdliang328@gmail.com}
}

\begin{document}

\twocolumn[
  \begin{@twocolumnfalse}
    \maketitle
    \begin{abstract}
    The remarkable progress of reinforcement learning (RL) is intrinsically tied to the environments used to train and evaluate artificial agents. Moving beyond traditional qualitative reviews, this work presents a large-scale, data-driven empirical investigation into the evolution of RL environments. By programmatically processing a massive corpus of academic literature and rigorously distilling over 2,000 core publications, we propose a quantitative methodology to map the transition from isolated physical simulations to generalist, language-driven foundation agents. Implementing a novel, multi-dimensional taxonomy, we systematically analyze benchmarks against diverse application domains and requisite cognitive capabilities. Our automated semantic and statistical analysis reveals a profound, data-verified paradigm shift: the bifurcation of the field into a "Semantic Prior" ecosystem dominated by Large Language Models (LLMs) and a "Domain-Specific Generalization" ecosystem. Furthermore, we characterize the "cognitive fingerprints" of these distinct domains to uncover the underlying mechanisms of cross-task synergy, multi-domain interference, and zero-shot generalization. Ultimately, this study offers a rigorous, quantitative roadmap for designing the next generation of Embodied Semantic Simulators, bridging the gap between continuous physical control and high-level logical reasoning.
    \end{abstract}
    \vspace{0.5cm}
    \textbf{Keywords: Environment Taxonomy \textbullet{} Large Language Models (LLMs) \textbullet{} Agent Capabilities \textbullet{} Cross-Domain Generalization \textbullet{} Benchmarks \textbullet{} Reinforcement learning \textbullet{} RL environment \textbullet{} Community Evolution}
    \vspace{0.5cm}
  \end{@twocolumnfalse}
]

\section{Introduction}

Reinforcement Learning (RL) has established itself as a foundational pillar of modern Artificial Intelligence, providing the theoretical mechanism for agents to learn optimal behaviors through trial-and-error interaction. Unlike supervised learning, which operates on static, annotated datasets, RL is inherently dynamic; it relies on the \textit{Agent-Environment Interface} to facilitate a cyclical exchange of states, actions, and rewards~\cite{sutton2018reinforcement}. While the agent represents the adaptive learner—encapsulating the policy and value functions—the \textbf{environment} constitutes the physical or virtual reality that defines the problem boundaries, transition dynamics, and success criteria.

In recent years, the symbiosis between agent-side architectures and environmental complexity has driven remarkable breakthroughs. We have witnessed RL agents mastering high-dimensional strategy games like Go~\cite{silver2016mastering} and StarCraft II~\cite{vinyals2019grandmaster}, solving complex continuous control tasks in robotics~\cite{levine2016end, mnih2015human}, and optimizing large-scale industrial operations. However, this progress has unveiled a critical paradox: while policy optimization techniques have become increasingly sophisticated, they are often brittle, over-fitting to the specific idiosyncrasies of their training environments. A growing body of literature points to a ``crisis of reproducibility'' and generalization, where agents achieving superhuman scores on one benchmark fail catastrophically when subjected to minor environmental perturbations~\cite{henderson2018deep}. Thus, the environment is not merely a backdrop for training; it is the decisive factor that determines the robustness, safety, and generalizability of intelligent systems. 

\textbf{This paper advances a central thesis:} the historical progress of Reinforcement Learning has been fundamentally driven not only by methodological advancements, but by a continuous escalation in the \emph{structural and cognitive complexity of environments}. We identify a unifying pattern: the evolution of RL environments follows a consistent trajectory toward higher levels of \emph{cognitive abstraction}. Specifically, we observe a paradigm shift from environments dominated by low-dimensional physical dynamics to those requiring high-level semantic reasoning, long-horizon planning, and multi-modal integration. This transition is accompanied by an increase in the coupling of agent capabilities, a shift in reward structures from dense heuristics to sparse, preference-based signals, and a fundamental change in transfer dynamics—from shared physical regularities to shared abstract representations.

Despite its critical importance, the landscape of RL environments remains fragmented. The rapid proliferation of testbeds—from the Arcade Learning Environment (ALE)~\cite{bellemare2013arcade} to physics-based simulators like MuJoCo~\cite{todorov2012mujoco} and photorealistic platforms like Unity~\cite{juliani2018unity}—has created a ``Wild West'' scenario. Concurrently, with the rapid advancements in Large Language Models (LLMs), the application of RL has aggressively expanded into abstract semantic environments~\cite{brown2020language, openai2023gpt4}. Researchers often select environments based on popularity rather than task suitability, leading to inconsistent benchmarking and a lack of clarity regarding which task properties actually drive capability emergence. 

However, the vast majority of review literature focuses almost exclusively on model architectures and optimization algorithms, treating the simulation environments as static, secondary components. This critical oversight leaves fundamental questions unanswered regarding the empirical foundations of RL. To address this gap, this research systematically investigates the following core Research Questions (RQs):
\begin{itemize}[leftmargin=*, itemsep=2pt, topsep=4pt]
    \item \textbf{RQ1 (Taxonomy \& Distribution):} What are the fundamental dimensions that characterize RL environments, and how are modern benchmarks distributed across different application domains and required capabilities?
    \item \textbf{RQ2 (Evolutionary Trajectory):} How has the structural and cognitive complexity of these environments evolved in tandem with AI breakthroughs, transitioning from spatial physics to semantic reasoning?
    \item \textbf{RQ3 (Transfer Dynamics):} How do different environmental domains and task modalities interact, and what underlying mechanisms govern synergistic transfer versus catastrophic interference across multi-domain learning?
\end{itemize}

\textbf{To systematically answer these questions, we structure our research and contributions across three critical dimensions:}

\begin{figure*}[h!]
    \centering
    \includegraphics[width=0.9\textwidth]{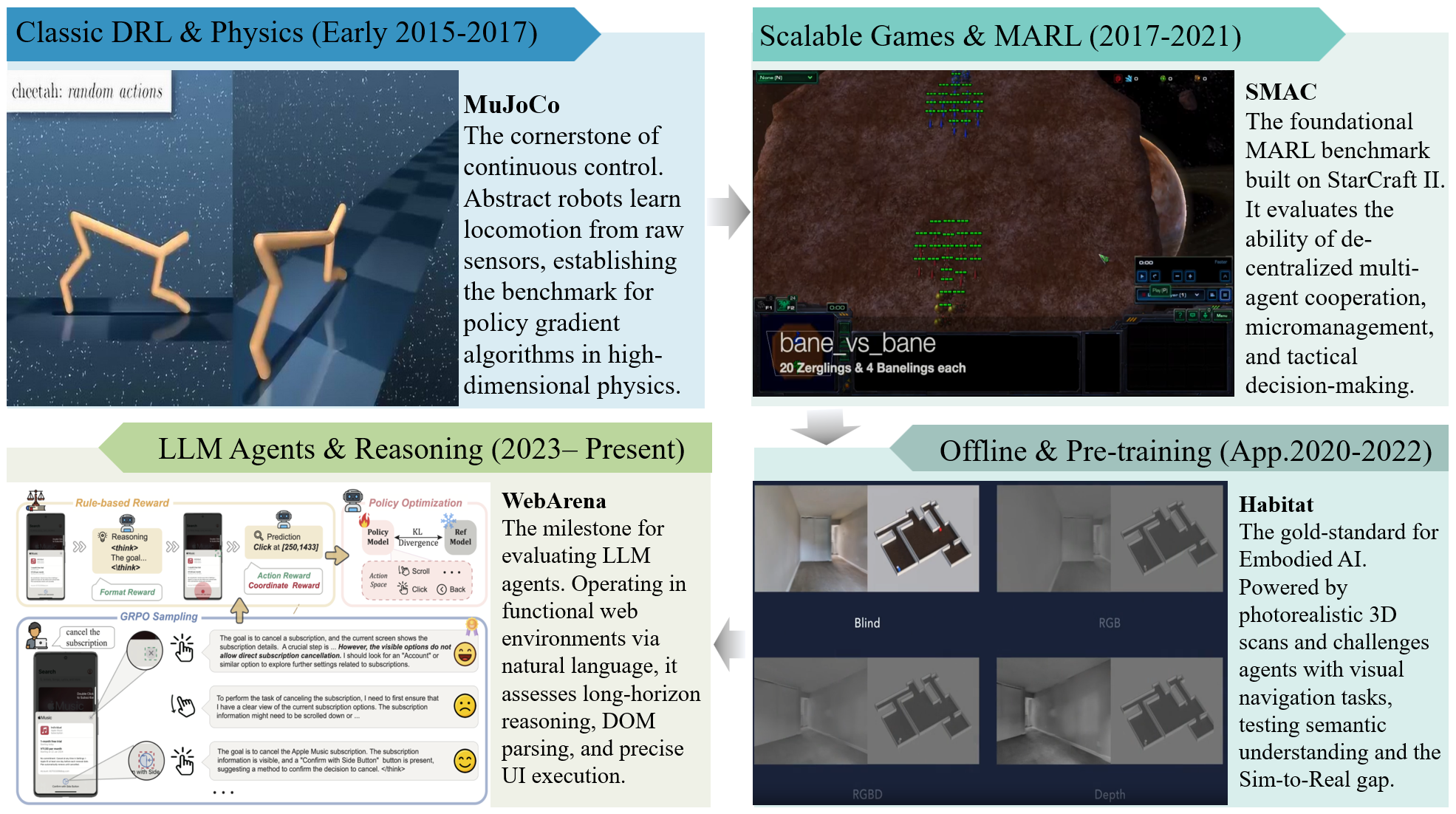}
    \caption{The Evolution of Reinforcement Learning Environments: A chronological visual timeline illustrating the paradigm shifts from classic continuous control and multi-agent coordination, to data-driven embodied AI, and ultimately to semantic reasoning via autonomous LLM agents.}
    \label{fig:rl_evolution_timeline}
\end{figure*}

\begin{figure*}[p] 
    \centering
    \includegraphics[width=0.95\textwidth, height=0.95\textheight, keepaspectratio]{images/rivermap.pdf}

    \caption{The Evolutionary Tree of Reinforcement Learning Environments: The Ascent of Cognitive Abstraction.}
    \label{fig:evolution_tree_vertical}
\end{figure*}

\paragraph{1. Systematic Taxonomy and Attribute Analysis.}
Addressing \textbf{RQ1}, the inherent diversity of RL tasks necessitates a rigorous taxonomy to evaluate algorithmic robustness. We categorize environments across seven strategic dimensions: 
\begin{itemize}[leftmargin=*, itemsep=2pt, topsep=4pt]
    \item \textbf{Agent Population:} Distinguishes single-agent paradigms from Multi-Agent Reinforcement Learning (MARL), encompassing both cooperative coordination and adversarial game-theoretic dynamics.
    \item \textbf{Application Domains:} Categorizes environments by their practical or simulated focus, spanning classical control, robotics, strategic gaming, and modern software ecosystems.
    \item \textbf{Agent Capabilities:} Evaluates the specific cognitive and algorithmic demands placed on the agent, ranging from explicit logical deduction to temporal memory retention for non-Markovian settings.
    \item \textbf{Observability:} Quantifies environmental transparency, contrasting fully observable Markov environments with Partially Observable Markov Decision Processes (POMDPs) that necessitate historical context encoding.
    \item \textbf{Multi-modal Span:} Assesses the requirement for fusing heterogeneous information channels, such as raw visual pixels, natural language instructions, and proprioceptive telemetry.
    \item \textbf{Action Space Modality:} Defines the mathematical and structural nature of the agent's interventions, capturing the transition from classical discrete/continuous motor control to semantic, auto-regressive token generation.
    \item \textbf{Reward Formulation:} Examines the density, granularity, and origin of the optimization signal, contrasting dense programmatic heuristics with sparse terminal goals and human-aligned preference models.
\end{itemize}

\paragraph{2. The Evolutionary Trajectory of Environments.}
To answer \textbf{RQ2}, we utilize milestones in RL advancements as key dividing points to analyze the evolution of environments across four distinct periods (Figure \ref{fig:rl_evolution_timeline}). By tracing this chronological trajectory, we reveal how benchmark designs have mirrored and catalyzed broader shifts in artificial intelligence (Figure \ref{fig:evolution_tree_vertical}). This research highlights the transition from early ``toy problems'' (e.g., GridWorld~\cite{sutton2018reinforcement}, CartPole~\cite{barto1983neuronlike}) designed to verify theoretical convergence, to the era of Deep RL characterized by visual complexity, and finally to the current frontier of multimodal Foundation Models, Embodied AI, and Sim-to-Real transfer~\cite{tobin2017domain}. The focus has explicitly shifted from maximizing scores in deterministic games to mastering procedural generation~\cite{cobbe2020leveraging} and open-ended exploration.

\paragraph{3. Task Synergy, Interference, and Transfer Dynamics.}
Addressing \textbf{RQ3}, as the field advances towards Multi-Task and Meta-Reinforcement Learning, treating environments in isolation is no longer sufficient. A central contribution of this research is the systematic analysis of \textbf{Inter-Task Dynamics}. We investigate the underlying mechanisms of:
\begin{itemize}[leftmargin=*, itemsep=2pt, topsep=4pt]
    \item \textbf{Synergistic Effects (Positive Transfer):} Where learning a source task accelerates the mastery of a target task through shared sub-skills or representation learning.
    \item \textbf{Negative Capabilities (Interference):} Where competing objectives or conflicting gradients across tasks lead to performance degradation or ``catastrophic forgetting.''
\end{itemize}
Understanding which task clusters enable generalization is pivotal for designing effective curricula and pre-training strategies. This research identifies the specific structural conflicts and cognitive alignments that govern cross-domain capability transfer.

By synthesizing these perspectives, this research aims to establish a unified framework for understanding the ``World'' side of the RL equation. The data collection, pre-processing protocol, and analysis methods are detailed in Appendix \ref{app:methodology_details}. From an initial pool of 2,183 quantitatively evaluated papers, a core set of over 200 milestone environments was selected for in-depth taxonomic analysis. By anchoring our taxonomy in this rigorously filtered dataset, we trace the genuine paradigm shifts in benchmark design, providing a definitive reference for researchers seeking to select appropriate testbeds and design the next generation of environments for adaptive intelligence.

\section{Conceptual Framework: the Agent-Environment Interface of Reinforcement Learning}

At its conceptual core, Reinforcement Learning (RL) represents a shift from static pattern recognition to dynamic, sequential decision-making. Unlike supervised learning, which relies on externally curated datasets, RL is grounded in the paradigm of \textit{active interaction}. The learner, termed the \textbf{agent}, must discover optimal behavioral strategies solely through feedback signals elicited from its actions within a dynamic system~\cite{sutton2018reinforcement}. 

While the MDP formalism provides a convenient abstraction\cite{bellman1957markovian}, we emphasize that it is not a strict requirement for defining reinforcement learning environments. In practice, many modern environments—particularly those involving language, memory, or tool use—violate the Markov assumption. Therefore, we adopt a more general perspective in which the environment may be partially observable, history-dependent, or even implicitly defined within the agent itself. While standard literature often uses the term ``environment'' to encompass the entire MDP tuple, for this study—and to rigorously categorize existing benchmarks—it is essential to distinguish between the \textbf{Environment} (the physical world and its dynamics) and the \textbf{Task} (the specific objective). Accordingly, we decompose the MDP $\mathcal{M}$ into two constituent components: the Environmental Dynamics $\mathcal{E}$ and the Task Specification $\mathcal{T}$.

In this broader view, the state space $\mathcal{S}$ need not be explicitly defined or fully observable, but may instead correspond to latent variables, historical context, or internal representations within the agent. Similarly, the transition function $\mathcal{P}$ may be stochastic, learned, or even implicitly defined by external systems such as simulators, human feedback, or language models.

\subsection{Environments \& Tasks}
The \textbf{Environment} $\mathcal{E}$ defines the immutable laws of the world in which the agent operates. It is formally characterized by the tuple $\mathcal{E} = \langle \mathcal{S}, \mathcal{A}, \mathcal{P} \rangle$:
\begin{itemize}[leftmargin=*, itemsep=2pt, topsep=4pt]
    \item $\mathcal{S}$ is the \textbf{State Space}, encapsulating all possible configurations (e.g., robot joint angles, pixel inputs, or textual dialogue histories).
    \item $\mathcal{A}$ is the \textbf{Action Space}, defining the set of control primitives available to the agent (e.g., motor torques, API calls, or generated language tokens).
    \item $\mathcal{P}: \mathcal{S} \times \mathcal{A} \times \mathcal{S} \to [0,1]$ is the \textbf{Transition Function}, representing the causal laws or ``physics'' of the domain. It dictates the probability $P(s'|s,a)$ of evolving to state $s'$ given action $a$ in state $s$.
\end{itemize}
Conceptually, $\mathcal{E}$ represents \textit{where} the agent is and \textit{what} it can do, independent of what it \textit{should} do. For instance, in a robotic simulation, $\mathcal{E}$ encompasses the robot's kinematics, gravity, and friction~\cite{todorov2012mujoco}. Crucially, in modern reinforcement learning, the formulation of $\mathcal{E}$ extends far beyond spatial and physical physics to encompass \textit{semantic and digital spaces}. In language-driven or web-based environments, the environment dictates the logic of a text parser, the state of a graphical user interface (GUI), or the response of an external tool, treating natural language itself as the interactive physics of the domain~\cite{cote2018textworld, yao2023reactsynergizingreasoningacting}.

Within the environment suite, another crucial concept is the Task. \textbf{Task} $\mathcal{T}$ superimposes a goal upon the environment. It is defined by the tuple $\mathcal{T} = \langle \mathcal{R}, \rho_0, \gamma, H \rangle$:
\begin{itemize}[leftmargin=*, itemsep=2pt, topsep=4pt]
    \item $\mathcal{R}: \mathcal{S} \times \mathcal{A} \to \mathbb{R}$ is the \textbf{Reward Function}, a scalar signal quantifying the immediate utility of a transition. This is the primary driver of behavior, encoding the goal (e.g., +1 for reaching a target, -1 for falling, or a BLEU score for text generation).
    \item $\rho_0: \mathcal{S} \to [0,1]$ denotes the \textbf{Initial State Distribution}, specifying where the agent begins an episode.
    \item $\gamma \in [0,1]$ is the \textbf{Discount Factor}, determining the agent's foresight horizon.
    \item $H$ represents the \textbf{Horizon} or termination condition, distinguishing between episodic and continuing tasks.
\end{itemize}
The Task $\mathcal{T}$ represents the semantic intent of the problem. A single environment $\mathcal{E}$ can support infinite tasks $\mathcal{T}$ by varying the reward structure $\mathcal{R}$~\cite{taylor2009transfer}.

\subsection{Distinction and Interplay}
The distinction between environment and task is not merely semantic but fundamental to understanding generalization in RL. 
\begin{itemize}[leftmargin=*, itemsep=2pt, topsep=4pt]
    \item \textbf{One-to-Many Mapping:} A single environment (e.g., a maze layout) can host multiple distinct tasks (e.g., finding the exit, patrolling corridors, or avoiding traps). This relationship is central to \textit{Multi-Task RL} and \textit{Meta-RL}, where the goal is to learn a policy that can adapt to new rewards $\mathcal{R}$ within a fixed dynamic $\mathcal{P}$~\cite{finn2017model}.
    \item \textbf{Domain Adaptation:} Conversely, the task may remain constant (e.g., ``grasp the cup'') while the environment changes (e.g., friction coefficients shift, or visual textures change). This is the domain of \textit{Sim-to-Real} transfer~\cite{tobin2017domain}.
\end{itemize}

Throughout this research, we will utilize this distinction to analyze benchmarks, separating those that test an agent's ability to master complex dynamics ($\mathcal{P}$) from those that test the ability to solve complex cognitive objectives ($\mathcal{R}$).

\subsection{Theoretical basis of Agent Capabilities in cognitive psychology}

To provide a rigorous taxonomy of reinforcement learning (RL) environments, we ground the required agent capabilities in established constructs from cognitive psychology and neurobiology. We move beyond a purely task-oriented view, framing the challenges of the environment as benchmarks for specific higher-order cognitive functions.

\begin{itemize}[leftmargin=*, itemsep=2pt, topsep=4pt]
    \item \textbf{Memory \& Retrieval} are rooted in the concepts of \textit{Working Memory} and \textit{Episodic Buffer}. Psychologically, this involves the transient maintenance and manipulation of information necessary for complex cognitive tasks \cite{baddeley2000episodic}. In RL, this maps to the agent's ability to maintain internal states to resolve Partial Observability and long-term temporal dependencies.
    
    \item \textbf{Deduction \& Inference} corresponds to \textit{Relational Reasoning}---the capacity to identify and manipulate the relationships between mental representations \cite{penn2008darwin}. This transcends simple associative learning, requiring the agent to deduce latent causal structures and perform multi-step reasoning to navigate hierarchical dependencies.
    
    \item \textbf{Induction \& Generalization} reflect \textit{Inductive Logic} and \textit{Abstract Categorization}. In cognitive science, this refers to the ability to synthesize generalizable rules from sparse data \cite{tenenbaum2011how}. This capability is essential for Meta-RL scenarios where the agent must "learn-to-learn" across novel task distributions.
    
    \item \textbf{Strategy \& Game Play} are grounded in \textit{Theory of Mind (ToM)} and \textit{Social Cognition}. Strategic tasks require agents to model the mental states, intents, and beliefs of other entities \cite{premack1978does}. This is the psychological basis for mastering adversarial dynamics and seeking Nash Equilibria in multi-agent systems.
    
    \item \textbf{Planning \& Search} align with \textit{Prospective Memory} and \textit{Mental Simulation}. Cognitively, this involves "looking ahead" by simulating future outcomes within an internal world model before execution \cite{gilbert2007prospective}. This integration of learned value functions with tree search allows for optimized decision-making in complex state spaces.
    
    \item \textbf{Numerical Computation} relates to \textit{Numerical Cognition} and \textit{Symbolic Processing}. This specialized faculty allows for the precise quantification of resources and the execution of arithmetic rules \cite{dehaene1998abstract}, framing calculation as a sequential, goal-directed cognitive process rather than simple pattern matching.
    
    \item \textbf{Control \& Manipulation} are manifestations of \textit{Sensorimotor Coordination} and \textit{Proprioception}. These involve the seamless integration of sensory inputs with motor outputs to manage high-degree-of-freedom (DoF) systems \cite{wolpert2003motor}, requiring robust policy optimization under complex physical and contact dynamics.
    
    \item \textbf{Structural Analysis \& Evaluation} correspond to \textit{Syntactic and Hierarchical Processing}. Just as humans process language or spatial layouts through hierarchical structures \cite{chomsky1956three}, agents in these environments must recognize and optimize underlying graphs, netlists, or grammars to ensure logical and structural correctness.
\end{itemize}

\subsection{Theoretical basis of Observability}
The nature of the information available to an agent fundamentally dictates the mathematical framework and the requisite cognitive capabilities (e.g., memory) for solving a task.

\begin{itemize}[leftmargin=*, itemsep=2pt, topsep=4pt]
    \item \textbf{Full Observability (Perfect Information):} In these environments, the agent's observation $O_t$ is equivalent to the true environment state $S_t$. Formalized through Markov Decision Processes (MDPs), tasks like Chess or Go provide perfect information, where the optimal policy depends only on the current state without requiring historical context \cite{silver2016mastering}.
    
    \item \textbf{Partial Observability (Incomplete Information):} Modeled as Partially Observable MDPs (POMDPs), these environments provide noisy or incomplete data \cite{kaelbling1998planning}. Agents must maintain an internal "belief state" or utilize recurrent architectures (e.g., LSTMs) to infer hidden variables from past sequences of observations and actions.
    
    \item \textbf{Imperfect Information in Strategic Games:} Specifically in multi-agent adversarial settings, imperfect information arises when certain state elements (such as an opponent's cards or intent) are hidden \cite{brown2019superhuman}. Success in these domains requires identifying Nash Equilibria and modeling the hidden strategies of competitors through Theory of Mind.
\end{itemize}

\subsection{Action Space Modality}
The action space $\mathcal{A}$ dictates the output probability distribution structure of the policy network $\pi(a|s)$. To comprehensively cover the diverse applications of Reinforcement Learning (RL), we categorize the action space into three primary theoretical paradigms:

\begin{itemize}[leftmargin=*, itemsep=2pt, topsep=4pt]
    \item \textbf{Discrete Action Spaces:} Under this paradigm, the agent's interventions belong to a finite set ($|\mathcal{A}| < \infty$). In low-dimensional scenarios (e.g., classic Atari games), the policy can directly output action probabilities via a Softmax layer \cite{mnih2015human}. However, in board games or complex graphic control environments, actions exhibit a \textit{Combinatorial \& Multi-discrete} tendency. This induces the ``curse of dimensionality,'' typically necessitating autoregressive action generation or branching network architectures to mitigate the exponentially large exploration space \cite{vinyals2019grandmaster}.
    
    \item \textbf{Continuous Action Spaces:} Primarily targeting robotics and physical simulations, the action space here is a real-valued vector space $\mathcal{A} \subseteq \mathbb{R}^n$. Unlike discrete spaces, continuous spaces cannot exhaustively enumerate action values (Q-values) and must rely on Policy Gradient or Actor-Critic architectures (e.g., DDPG, PPO) to directly output the mean and variance of multivariate Gaussian distributions \cite{lillicrap2015continuous}. As the degrees of freedom in robotic manipulators increase, the space transitions from low-dimensional control to \textit{High-dimensional Kinematics}.
    
    \item \textbf{Hybrid \& Non-standard Spaces:} This serves as the theoretical bridge connecting classical RL to modern Embodied AI and LLMs. On one hand, \textit{Parameterized Actions} allow the agent to output a discrete action category alongside continuous action parameters (e.g., the command ``Move to coordinates $(x, y)$'' in StarCraft) \cite{hausknecht2015deep}. On the other hand, with the proliferation of LLMs acting as the core cognitive brain of agents, \textit{Text \& Token-based Output} has emerged as a novel non-standard action space, where the agent's actions manifest as autoregressive token generation within a linguistic space \cite{yao2023reactsynergizingreasoningacting}.
\end{itemize}

More generally, we define the action space as any mechanism through which the agent can influence the future trajectory of the system. This includes not only physical actions, but also symbolic outputs, natural language generation, API calls, and latent decisions that are not directly observable but affect downstream computation or interaction.

\subsection{Reward Formulation and Density}
The reward function $\mathcal{R}(s, a, s')$ is the sole target-driven signal in RL. According to the ``Reward Hypothesis,'' all goals of an agent can be described by the maximization of the expected cumulative scalar reward \cite{sutton2018reinforcement}. We theoretically decompose this formulation along two orthogonal dimensions: temporal density and source structure.

\begin{itemize}[leftmargin=*, itemsep=2pt, topsep=4pt]
    \item \textbf{Temporal Density:} This dimension directly determines the severity of the Credit Assignment Problem (CAP). \textit{Dense Rewards} provide high-frequency, distance- or progress-based signals (Reward Shaping), allowing the policy to converge efficiently \cite{ng1999policy}. In contrast, \textit{Sparse \& Delayed Rewards} (e.g., a binary win/loss signal at the end of a long episode) pose one of the most formidable challenges in RL, often requiring Intrinsic Motivation or advanced search strategies (e.g., MCTS) to resolve effectively \cite{burda2018exploration}.
    
    \item \textbf{Source \& Structure:} Traditional RL relies heavily on an \textit{Environment-defined Scalar} (e.g., game score or physical distance). However, complex real-world tasks often demand balancing conflicting optimization objectives (e.g., speed vs. energy efficiency), giving rise to \textit{Multi-objective / Vector} reward systems \cite{roijers2013survey}. Crucially, in the era of foundation models, the source of rewards is undergoing a profound paradigm shift: to circumvent ``reward hacking,'' modern RL incorporates \textit{Human-aligned \& Learned} rewards. Through Reinforcement Learning from Human Feedback (RLHF) \cite{ouyang2022training} and preference models, rewards are no longer pre-defined, hard-coded formulas, but rather high-dimensional representations implicitly learned from human values.
\end{itemize}

From a unified perspective, the reward function can be interpreted as an externalization of preference over trajectories, rather than a fixed scalar function. This perspective naturally encompasses learned reward models, human feedback, and implicit optimization signals used in modern alignment methods.

\section{Functional Motivation: Why Environments Drive Algorithmic Advancement?}

Following the fundamental definitions of reinforcement learning (RL) and its environmental components, it is crucial to articulate the functional necessity of environments in the broader research landscape. The environment is not merely a passive container for the agent; it constitutes the \textbf{generative crucible} of the data distribution itself. Unlike supervised learning, where data is static, identically distributed, and pre-collected, RL environments provide a dynamic, interactive manifold where the agent's policy actively shapes its own training distribution. This unique bidirectional coupling positions the environment as the primary evolutionary driver of algorithmic advancement, fulfilling three indispensable functions: establishing standardized benchmarking paradigms, ensuring scalable and safe exploration, and catalyzing the leap towards high-order cognitive complexity.

\subsection{Standardization and the Paradigm Shift of Benchmarking}
From an empirical research perspective, environments serve as the definitive ``control variables'' in the scientific method of algorithmic evaluation. The historical breakthroughs in Deep RL were inextricably linked to the introduction of standardized testbeds, such as the \textbf{Arcade Learning Environment (ALE)} \cite{bellemare2013arcade} and \textbf{OpenAI Gym} \cite{brockman2016openai}. By formalizing the interaction interfaces---standardizing observation spaces, transition dynamics, and reward scales---these platforms allowed researchers to isolate and quantify the exact contributions of algorithmic innovations like Deep Q-Networks (DQN) or Proximal Policy Optimization (PPO).

Beyond mere convenience, standardized environments form the primary defense against the notoriously pervasive ``reproducibility crisis'' in RL \cite{henderson2018deep}. Historically, physics engines like \textbf{MuJoCo} \cite{todorov2012mujoco} provided the rigorous baselines needed for continuous control. Today, as RL expands into the Foundation Model era, this standardizing function has shifted towards digital and cognitive domains. Modern benchmarks like \textbf{SWE-bench} \cite{jimenez2023swebench} and \textbf{WebArena} \cite{zhou2023webarena} provide immutable, standardized metrics to evaluate the otherwise opaque reasoning and tool-use capabilities of Large Language Model (LLM) agents, proving that as algorithms evolve, the benchmarking environments must co-evolve to maintain empirical rigor.

\subsection{Safety, Cost-Efficiency, and the Reality Gap}
In applied domains, the environment functions as an indispensable safety buffer and an economic accelerator. Training agents directly in the physical or production world is often prohibitive due to the immense ``sample complexity'' of modern RL algorithms, which may require millions of trial-and-error interactions to converge. In high-stakes fields---such as autonomous driving, industrial robotics, or live financial trading---unconstrained exploration poses unacceptable risks to human safety, hardware integrity, and economic stability \cite{garcia2015comprehensive}.

Simulated environments completely neutralize this risk, transforming catastrophic physical failures into benign digital reset signals. Furthermore, these environments are essential for implementing \textbf{Domain Randomization} \cite{tobin2017domain}, a technique where environmental parameters (e.g., friction, mass, lighting, or network latency) are heavily perturbed during training. This forces the agent to learn robust, invariant policies capable of traversing the ``Sim-to-Real'' gap. More recently, this concept has expanded into ``Digital-to-Real'' safety, where sandboxed executable environments (e.g., code interpreters) allow LLM agents to safely test and compile generated code before deployment, preventing critical software failures.

\subsection{Catalyzing Cognitive Evolution and System 2 Generalization}
Ultimately, the most profound function of the environment is its role as an evolutionary curriculum that actively shapes the cognitive architecture of the agent. As demonstrated by our capability taxonomy, early environments dominated by low-dimensional states or raw pixels naturally catalyzed the development of reactive policies and spatial perception (e.g., via Convolutional Neural Networks). However, to drive intelligence further, modern environments systematically introduce profound complexities such as partial observability, vast combinatorial action spaces, and abstract logical constraints.

This environmental pressure is directly responsible for the recent algorithmic shift from reactive pattern matching to deliberative ``System 2'' thinking. By introducing benchmarks that require open-ended procedural adaptation (e.g., \textbf{Procgen} \cite{cobbe2020leveraging}), long-horizon inductive abstraction (e.g., \textbf{ARC-AGI} \cite{chollet2019measure}), and step-level mathematical verification (e.g., \textbf{ProcessBench} \cite{zheng2025processbenchidentifyingprocesserrors}), the environment explicitly forces the agent to develop coupled cognitive chains. It is within these rigorous, structured environments that algorithms are compelled to fuse \textit{Deduction \& Inference} with \textit{Planning \& Search} (such as via Monte Carlo Tree Search). In this sense, the environment is not merely an evaluation metric; it is the fundamental curriculum that dictates the upper bound of artificial general intelligence \cite{silver2021reward}.

\section{Taxonomic Landscape: A Multi-dimensional Spectrum of Environment \& Task Types}
Drawing upon an extensive repository of thousands of heterogeneous reinforcement learning (RL) environments and tasks, and by synthesizing taxonomies from extant literature, we categorize these tasks across several key dimensions: multi-modal span, application domains, agent capabilities, observability (information completeness), action space, reward formulation, and agent population (multi-agent vs. single-agent).

\begin{figure*}[h]
    \centering
    \includegraphics[width=0.75\textwidth]{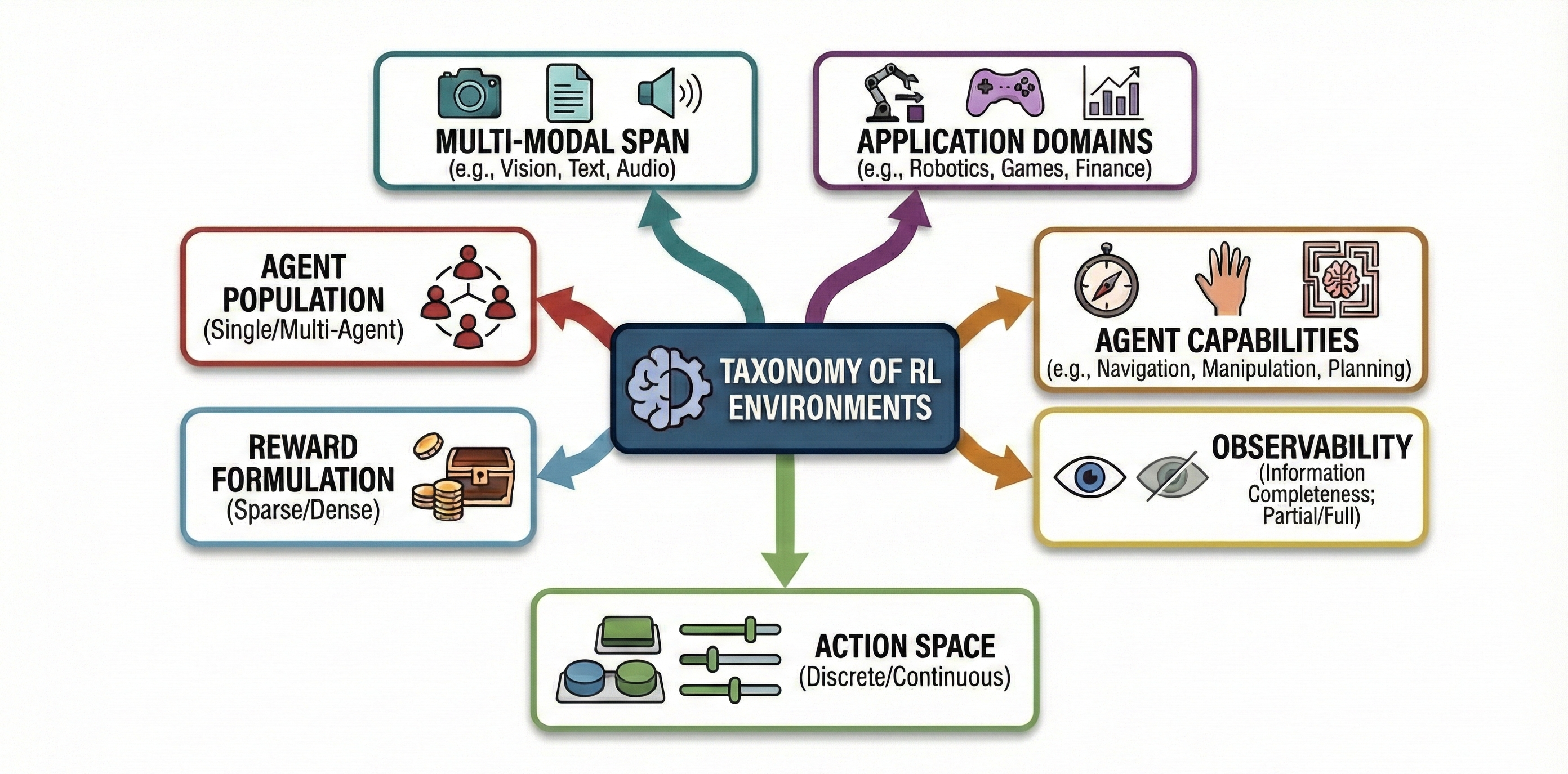}
    \caption{The taxonomy of multi-dimensional spectrum of reinforcement learning task types}
    \label{fig:1}
\end{figure*}

\subsection{Agent Population}
To categorize tasks based on agent population, we adopt the formal definitions established in foundational reinforcement learning literature. The distinction is rooted in the mathematical framework used to model the environmental dynamics: the Markov Decision Process (MDP) for single-agent settings and the Stochastic Game (SG) for multi-agent settings.

\begin{table*}[htbp]

  \centering

  \small 

  \caption{Comprehensive Taxonomy of Representative Single-Agent Reinforcement Learning Environments}

  \label{tab:sarl_envs_comprehensive}

  \renewcommand{\arraystretch}{1.15} 

  \begin{tabularx}{\linewidth}{@{} l l c X @{}}

    \toprule

    \textbf{Environment} & \textbf{Task / Domain} & \textbf{Year$^{\mathrm{a}}$} & \textbf{DOI / Source} \\

    \midrule

    \multicolumn{4}{@{}l}{\textbf{Part I: Classic Control, Arcade \& 3D Perception (Pre-2018)}} \\

    \midrule

    Arcade Learning Env (ALE)\cite{bellemare2013arcade} & Atari 2600 (Visual Pixel Control) & 2013 & \href{https://doi.org/10.1613/jair.3912}{10.1613/jair.3912} \\

    MuJoCo (Gym Control)\cite{todorov2012mujoco} & Continuous Physics Control & 2015 & \href{https://doi.org/10.1109/IROS.2012.6386109}{IROS'2012} \\

    OpenAI Gym (Classic Control)\cite{brockman2016openai} & Standardized Tabular \& Physics MDPs & 2016 & \href{https://doi.org/10.48550/arXiv.1606.01540}{arXiv:1606.01540} \\

    ViZDoom\cite{kempka2016vizdoom} & 3D First-Person Visual Perception & 2016 & \href{https://doi.org/10.1109/CIG.2016.7860433}{CIG'2016} \\

    DeepMind Lab (DMLab)\cite{beattie2016deepmind} & 3D Navigation \& Spatial Puzzles & 2016 & \href{https://doi.org/10.48550/arXiv.1612.03801}{arXiv:1612.03801} \\

    \midrule

    \multicolumn{4}{@{}l}{\textbf{Part II: Procedural Generation, Meta-Learning \& Embodied AI (2018--2022)}} \\

    \midrule

    MiniGrid\cite{chevalier2018minimalistic} & Procedural 2D Grid-world Navigation & 2018 & \href{https://github.com/Farama-Foundation/Minigrid}{GitHub: Minigrid} \\

    Habitat\cite{savva2019habitat} & Photorealistic 3D Visual Navigation & 2019 & \href{https://doi.org/10.48550/arXiv.1904.01201}{arXiv:1904.01201} \\

    Meta-World\cite{yu2020meta_world} & Meta-RL Robotic Manipulation & 2019 & \href{https://doi.org/10.48550/arXiv.1910.10897}{CoRL'19} \\

    ProcGen\cite{cobbe2020leveraging} & Procedurally Generated 2D Games & 2019 & \href{https://doi.org/10.48550/arXiv.1912.01588}{ICML'20} \\

    ALFWorld\cite{shridhar2020alfworld} & Text-aligned Embodied Household Tasks & 2020 & \href{https://doi.org/10.48550/arXiv.2010.03768}{ICLR'21} \\

    Brax\cite{freeman2021brax} & Hardware-Accelerated Rigid Physics & 2021 & \href{https://doi.org/10.48550/arXiv.2106.13281}{NeurIPS'21} \\

    MineDojo (Minecraft)\cite{fan2022minedojobuildingopenendedembodied} & Open-Ended Embodied Survival & 2022 & \href{https://doi.org/10.48550/arXiv.2206.08853}{NeurIPS'22} \\

    \midrule

    \multicolumn{4}{@{}l}{\textbf{Part III: The Foundation Era: Web, GUI \& Software Agents (2022--Present)}} \\

    \midrule

    WebShop\cite{yao2022webshop} & Simulated E-commerce Web Navigation & 2022 & \href{https://doi.org/10.48550/arXiv.2207.01206}{NeurIPS'22} \\

    WebArena\cite{zhou2023webarena} & Highly Realistic Web Environment & 2023 & \href{https://doi.org/10.48550/arXiv.2307.13854}{ICLR'24} \\

    SWE-bench\cite{jimenez2023swebench} / SWE-Gym\cite{pan2025trainingsoftwareengineeringagents} & Real-World Software Engineering & 2023 & \href{https://doi.org/10.48550/arXiv.2310.06770}{arXiv:2310.06770} \\
    Code Interpreter\cite{bubeck2023sparksartificialgeneralintelligence} & Tool-Integrated Sandboxed Execution & 2023 & \href{https://doi.org/10.48550/arXiv.2303.12712}{arXiv:2303.12712} \\

    OSWorld\cite{xie2024osworld} & Multimodal Desktop OS Automation & 2024 & \href{https://doi.org/10.48550/arXiv.2404.07972}{arXiv:2404.07972} \\

    AndroidWorld\cite{rawles2024androidworld} / AndroidLab & Mobile GUI Control \& Interaction & 2024 & \href{https://doi.org/10.48550/arXiv.2405.14573}{arXiv:2405.14573} \\

    MLE-bench (Kaggle)\cite{chan2024mle} & Autonomous Machine Learning Tasks & 2024 & \href{https://doi.org/10.48550/arXiv.2410.07095}{arXiv:2410.07095} \\

    \midrule

    \multicolumn{4}{@{}l}{\textbf{Part IV: LLM Alignment \& Verification Benchmarks (2021--Present)}} \\

    \midrule

    GSM8K\cite{cobbe2021training} & Grade School Math Word Problems & 2021 & \href{https://doi.org/10.48550/arXiv.2110.14168}{arXiv:2110.14168} \\

    MATH / MATH-500\cite{hendrycks2021measuring} & Competition-Level Math Reasoning & 2021 & \href{https://doi.org/10.48550/arXiv.2103.03874}{NeurIPS'21} \\

    HumanEval\cite{chen2021evaluating} \& MBPP & Python Code Generation \& Logic & 2021 & \href{https://doi.org/10.48550/arXiv.2107.03374}{arXiv:2107.03374} \\

    Lean 4\cite{moura2021lean} / MiniF2F\cite{zheng2021minif2f} & Formal Theorem Proving & 2021 & \href{https://doi.org/10.48550/arXiv.2109.00110}{ICLR'22} \\

    GPQA\cite{rein2023gpqagraduatelevelgoogleproofqa} & Graduate-Level Scientific Reasoning & 2023 & \href{https://doi.org/10.48550/arXiv.2311.12022}{arXiv:2311.12022} \\

    ARC-AGI\cite{chollet2019measure} & Abstraction \& Inductive Reasoning & 2019 & \href{https://doi.org/10.48550/arXiv.1911.01547}{arXiv:1911.01547} \\

    LiveCodeBench\cite{jain2024livecodebenchholisticcontaminationfree} & Contamination-Free Code Reasoning & 2024 & \href{https://doi.org/10.48550/arXiv.2403.07974}{arXiv:2403.07974} \\

    OlympiadBench\cite{he2024olympiadbench} & Advanced Mathematical Problem-Solving & 2024 & \href{https://doi.org/10.48550/arXiv.2402.14008}{arXiv:2402.14008} \\

    \midrule

    \multicolumn{4}{@{}l}{\textbf{Part V: System 2 Thinking, Search \& Logic (The PRM \& GRPO Era) (2024--Present)}} \\

    \midrule

    Game24\cite{yao2023tree} & Logic \& Search Tree Exploration & 2023 & \href{https://doi.org/10.48550/arXiv.2305.10601}{arXiv:2305.10601} \\

    Countdown Game\cite{deepseek2025r1} & Arithmetic Planning \& Target Search & 2025 & \href{https://doi.org/10.48550/arXiv.2503.09512}{arXiv:2503.09512} \\

    ProcessBench\cite{zheng2025processbenchidentifyingprocesserrors} & Step-level Reasoning Verification & 2024 & \href{https://doi.org/10.48550/arXiv.2412.06559}{arXiv:2412.06559} \\

    GraphRAG\cite{edge2025localglobalgraphrag} & Multi-hop QA via Search APIs & 2024 & \href{https://doi.org/10.48550/arXiv.2404.16130}{arXiv:2404.16130} \\

    BIRD (Text-to-SQL)\cite{li2023bird} & Database Schema Logical Mapping & 2023 & \href{https://doi.org/10.48550/arXiv.2305.03111}{NeurIPS'23} \\

    AlphaCode (CodeForces)\footnote{*}\cite{doi:10.1126/science.abq1158} & Complex Algorithmic Text Processing & 2022 & \href{https://doi.org/10.1126/science.abq1158}{Science}\\   VisualWebArena\cite{koh2024visualwebarenaevaluatingmultimodalagents} & Multimodal Reasoning \& Visual Web Tasks & 2024 & \href{https://arxiv.org/abs/2401.13649}{arxiv:2401.13649} \\
    Sokoban\cite{pastukhov2025solvingsokobanusinghierarchical,stojanovski2025reasoninggymreasoningenvironments} / Rubik's Cube\cite{stojanovski2025reasoninggymreasoningenvironments} & Planning \& Search in Reasoning Gym & 2025 & \href{https://doi.org/10.48550/arXiv.2504.04366}{arXiv:2504.04366} \\

    \bottomrule

  \end{tabularx}

  \vspace{0ex}

  \footnotesize{\textbf{Note:} Environments are categorized chronologically and thematically. Early eras focused on pixel inputs and physical control. In contrast, modern single-agent RL benchmarks explicitly test logical deduction, tool use, and System 2 thinking (e.g., GRPO/RLHF) in LLMs and Vision-Language Models. $^{\mathrm{a}}$The year indicates the formal introduction/publication. (Same applies below)}

\end{table*}

\paragraph{Single-Agent: The Evolving MDP Formulation}
Following the canonical definition by Sutton and Barto \cite{sutton2018reinforcement}, Single-Agent RL is formulated as a Markov Decision Process (MDP). An MDP is defined as a tuple $\langle \mathcal{S}, \mathcal{A}, \mathcal{P}, \mathcal{R}, \gamma \rangle$, where the environment is characterized by a state space $\mathcal{S}$ and a transition function $\mathcal{P}: \mathcal{S} \times \mathcal{A} \rightarrow \Delta(\mathcal{S})$. As noted by Bertsekas \cite{bertsekas2012dynamic}, a fundamental property of this formulation is that the environment is treated as a \textit{stationary} system, where state transitions depend exclusively on the agent's action and the current state, independent of time-varying external agents.

However, as cataloged comprehensively in Table \ref{tab:sarl_envs_comprehensive}, while this underlying mathematical invariant remains strict, the practical instantiations of the state space $\mathcal{S}$, action space $\mathcal{A}$, and reward signal $\mathcal{R}$ have undergone a profound paradigm shift across different research eras. 

In early physical and spatial simulators (Parts I \& II, e.g., ALE~\cite{bellemare2013arcade}, MuJoCo~\cite{todorov2012mujoco}), the MDP was strictly grounded in physics: $\mathcal{S}$ was heavily constrained to low-dimensional kinematic vectors or dense pixel arrays, and $\mathcal{A}$ comprised primitive, high-frequency motor controls. Conversely, the emergence of Foundation Models has drastically expanded the empirical boundaries of the single-agent MDP (Parts III -- V). In modern digital and System 2 reasoning benchmarks (e.g., WebArena~\cite{zhou2023webarena}, SWE-bench~\cite{jimenez2023swebench}, and ProcessBench~\cite{zheng2025processbenchidentifyingprocesserrors}), $\mathcal{S}$ is fundamentally semantic, spanning massive textual contexts or HTML DOM trees. The action space $\mathcal{A}$ has concurrently evolved from physical torque generation to discrete, auto-regressive token generation, encompassing high-level cognitive operations such as tool-use, API calls, and step-level logical deductions.

This chronological evolution demonstrates a critical realization: while the modern LLM agent still operates within an isolated, stationary MDP loop, the complexity of its interaction has migrated entirely from embodied physical entanglement to open-ended cognitive reasoning.

\begin{table*}[htbp]

  \centering

  \small 

  \caption{Comprehensive Taxonomy of Representative Multi-Agent Reinforcement Learning Environments}

  \label{tab:marl_envs_comprehensive}

  \renewcommand{\arraystretch}{1.15} 

  \begin{tabularx}{\linewidth}{@{} l l c X @{}}

    \toprule

    \textbf{Environment} & \textbf{Task / Domain} & \textbf{Year$^{\mathrm{a}}$} & \textbf{DOI / Source} \\

    \midrule

    \multicolumn{4}{@{}l}{\textbf{Part I: Classic Board Games \& Grid-World Micro-Management (2016--2019)}} \\

    \midrule

    Switch (DIAL)\cite{foerster2016learningcommunicatedeepmultiagent} & Partially-Observable Coordination & 2016 & \href{https://doi.org/10.48550/arXiv.1605.06676}{arXiv:1605.06676} \\

    Checkers (DIAL)\cite{foerster2016learningcommunicatedeepmultiagent} & Role Specialization (Collect vs. Clear) & 2016 & \href{https://doi.org/10.48550/arXiv.1605.06676}{arXiv:1605.06676} \\

    AlphaZero (Go/Chess/Shogi)\cite{silver2018general} & Self-play Board Games & 2017 & \href{https://doi.org/10.1038/nature24270}{10.1038/nature24270} \\

    Multi-Agent Particle Env (MPE)\cite{lowe2020multiagentactorcriticmixedcooperativecompetitive} & Partially-Observable Coordination & 2017 & \href{https://doi.org/10.48550/arXiv.1706.02275}{arXiv:1706.02275} \\

    MAgent\cite{zheng2017magentmanyagentreinforcementlearning} & Many-Agent Grid-world Combat & 2018 & \href{https://arxiv.org/abs/1712.00600}{arxiv.org:1712.00600} \\

    SMAC (StarCraft II)\cite{vinyals2019grandmaster} & Cooperative Micromanagement & 2019 & \href{https://doi.org/10.48550/arXiv.1902.04043}{arXiv:1902.04043} \\

    Hanabi Learning Env\cite{BARD2020103216} & Partially Observable Cooperative Game & 2019 & \href{https://doi.org/10.1016/j.artint.2019.103216}{Artificial Intelligence} \\

    Big 2\cite{charlesworth2018applicationselfplayreinforcementlearning} & Self-learning Card Game Agents & 2018 & \href{ https://doi.org/10.48550/arXiv.1808.10442}{arXiv:1808.10442} \\

    \midrule

    \multicolumn{4}{@{}l}{\textbf{Part II: Complex Simulation, Physics \& Logistics (2019--2023)}} \\

    \midrule

    Google Research Football\cite{kurach2020google} & Simulated Soccer \& Strategy & 2019 & \href{https://doi.org/10.48550/arXiv.1907.11180}{arXiv:1907.11180} \\

    Overcooked-AI\cite{carroll2019utility} & Human-AI Coordination \& Puzzles & 2019 & \href{https://doi.org/10.48550/arXiv.1910.05789}{arXiv:1910.05789} \\    EPyMARL\cite{papoudakis2021benchmarkingmultiagentdeepreinforcement} & Grid-world Foraging & 2020 & \href{https://doi.org/10.48550/arXiv.2006.07869}{arXiv:2006.07869} \\

    Robot Warehouse (RWARE)\cite{papoudakis2020benchmarking} & Multi-Robot Warehouse Logistics & 2020 & \href{https://doi.org/10.48550/arXiv.2006.07869}{arXiv:2006.07869} \\

    Habitat 3.0\cite{puig2023habitat30cohabitathumans} & Interactive \& Human-Robot Synergy & 2023& \href{https://doi.org/10.48550/arXiv.2310.13724}{arXiv:2310.13724} \\

    MA-Gym & Cooperative Grid-world Settings & 2021 & \href{https://github.com/koulanurag/ma-gym}{GitHub: ma-gym} \\

    VMAS\cite{bettini2022vmas} & Vectorized 2D Physics Control & 2022 & \href{https://doi.org/10.48550/arXiv.2207.03530}{arXiv:2207.03530} \\

    Isaac Gym\cite{makoviychuk2021isaac} & GPU-accelerated Physics Simulation & 2021 & \href{https://arxiv.org/abs/2108.10470}{arXiv:2108.10470} \\

    \midrule

    \multicolumn{4}{@{}l}{\textbf{Part III: Standardized Suites \& Hardware Acceleration (2020--2023)}} \\

    \midrule

    PettingZoo\cite{terry2021pettingzoogymmultiagentreinforcement} & General MARL Benchmark Suite & 2021 & \href{https://doi.org/10.48550/arXiv.2009.14471}{arXiv:2009.14471} \\

    Isaac Gym (MARL suite)\cite{makoviychuk2021isaac} & GPU-accelerated 3D Physics MARL & 2021 & \href{https://doi.org/10.48550/arXiv.2108.10470}{arXiv:2108.10470} \\

    JaxMARL\cite{rutherford2024jaxmarlmultiagentrlenvironments} & Hardware-Accelerated MARL on JAX & 2023 & \href{https://doi.org/10.48550/arXiv.2311.10090}{arXiv:2311.10090} \\

    \midrule

    \multicolumn{4}{@{}l}{\textbf{Part IV: LLM-Based Multi-Agent (Software, Tool-Use \& Planning) (2023--Present)}} \\

    \midrule

    Generative Agents\cite{park2023generativeagentsinteractivesimulacra} & Human Behavior Simulation & 2023 & \href{https://doi.org/10.48550/arXiv.2304.03442}{arXiv:2304.03442} \\ 

    AgentVerse\cite{chen2023agentversefacilitatingmultiagentcollaboration} & Multi-Agent Problem Decomposition & 2023 & \href{https://doi.org/10.48550/arXiv.2308.10848}{arXiv:2308.10848} \\

    MetaGPT\cite{hong2023metagpt} & Joint Evolution (Product, QA, Engineer) & 2023 & \href{https://doi.org/10.48550/arXiv.2308.00352}{arXiv:2308.00352} \\

    MindAgent\cite{gong2023mindagentemergentgaminginteraction} & Multi-Agent Text Search \& Defusal & 2023 & \href{https://doi.org/10.48550/arXiv.2309.09971}{arXiv:2309.09971} \\

    Mobile-Agent-v2\cite{khaniki2024novelapproachchestxray} & Mobile GUI (Navigator \& Interactor) & 2024 & \href{https://arxiv.org/abs/2406.01014}{arXiv:2404.14322} \\

    TAU-Bench (Airline/Telecom)\cite{fang2024taubench} & Multi-Agent Interacting Tool Use & 2024 & \href{https://doi.org/10.48550/arXiv.2406.12045}{arXiv:2406.12045} \\

    ColBench (SweetRL)\cite{zhou2025sweetrltrainingmultiturnllm} & LLM Collaborative Software Engineering & 2025 & \href{https://doi.org/10.48550/arXiv.2503.15478}{arXiv:2503.15478} \\

    Mobile GUI Agents (AMEX)\cite{chai-etal-2025-amex} & Multi-Agent GUI Interaction & 2025 & \href{https://aclanthology.org/2025.findings-acl.110}{ACL'25 Findings} \\

    ReSo\cite{zhou2025resorewarddrivenselforganizingllmbased} & Reward-driven Self-organizing MAS & 2025 & \href{https://doi.org/10.48550/arXiv.2503.02390}{arXiv:2503.02390} \\

    HiMA-Ecom (Former JoyAgents-R1)\cite{hu2026himaecomenablingjointtraining} & Multi-Agent E-commerce Assistants & 2025 & \href{https://doi.org/10.48550/arXiv.2506.19846}{arXiv:2506.19846} \\

    \midrule

    \multicolumn{4}{@{}l}{\textbf{Part V: LLM-Based Multi-Agent (Game Theory, Social \& Negotiation) (2024--Present)}} \\

    \midrule

    TextArena: Suite\cite{guertler2025textarena} & Interactive Text Collaboration & 2024 & \href{https://doi.org/10.48550/arXiv.2504.11442}{	arXiv:2504.11442} \\

    TextArena: Diplomacy\cite{guertler2025textarena} & Complex Strategy, Trust \& Betrayal & 2024 & \href{https://doi.org/10.48550/arXiv.2504.11442}{	arXiv:2504.11442} \\

    TextArena: Game Theory\cite{guertler2025textarena} & Iterated Prisoner's Dilemma, Stag Hunt & 2024 & \href{https://doi.org/10.48550/arXiv.2504.11442}{	arXiv:2504.11442} \\

    SPIRAL (Kuhn Poker)\cite{liu2026spiralselfplayzerosumgames} & Probabilistic Reasoning \& Self-Play & 2025 & \href{https://doi.org/10.48550/arXiv.2506.24119}{arXiv:2506.24119} \\

    Divide-Fuse-Conquer\cite{zhang2025dividefuseconquerelicitingahamoments} & ConnectFour \& Multi-Scenario Games & 2025 & \href{https://doi.org/10.48550/arXiv.2505.16401}{arXiv:2505.16401} \\

    SynLogic\cite{liu2025synlogicsynthesizingverifiablereasoning} & Logic \& Controllable Data Synthesis & 2025 & \href{https://doi.org/10.48550/arXiv.2505.19641}{arXiv:2505.19641} \\

    gg-bench\cite{verma2025measuringgeneralintelligencegenerated} & Generated Games \& LLM Intelligence & 2025 & \href{https://doi.org/10.48550/arXiv.2505.07215}{arXiv:2505.07215} \\

    MAGRPO \& MARFT\cite{liao2025marftmultiagentreinforcementfinetuning} & Multi-Agent RL Alignment Frameworks & 2025 & \href{https://doi.org/10.48550/arXiv.2504.16129}{arXiv:2504.16129} \\

    \bottomrule

  \end{tabularx}

  \vspace{0ex} 

  \footnotesize{\textbf{Note:} Environments are categorized chronologically and thematically to illustrate the paradigm shift from physical/pixel-based multi-agent control towards LLM-driven cognitive and social interaction benchmarks.}

\end{table*}

\paragraph{Multi-Agent: The Stochastic Game Formulation}

When an environment is populated by more than one adaptive entity, the foundational single-agent MDP formulation breaks down. As formalized by Shapley \cite{shapley1953stochastic} and Littman \cite{littman1994markov}, multi-agent reinforcement learning (MARL) is fundamentally modeled as a Markov Game, or Stochastic Game (SG), defined by the tuple $\langle \mathcal{N}, \mathcal{S}, \{\mathcal{A}_i\}_{i \in \mathcal{N}}, \mathcal{P}, \{\mathcal{R}_i\}_{i \in \mathcal{N}}, \gamma \rangle$, where $\mathcal{N}$ represents the set of $N$ agents. 

The critical divergence from the MDP framework lies in the transition function $\mathcal{P}: \mathcal{S} \times \mathcal{A}_1 \times \dots \times \mathcal{A}_N \rightarrow \Delta(\mathcal{S})$ and the reward function $\mathcal{R}_i$. Here, the environment's state transition and the reward received by agent $i$ are contingent not only on its own action but upon the \textit{joint action} profile of all agents. This interdependence shatters the \textit{stationary} assumption of the MDP. From the perspective of any single agent, the environment becomes highly \textit{non-stationary} as other agents continuously update their policies. Furthermore, this topology introduces complex game-theoretic dynamics, ranging from pure cooperation (where $\mathcal{R}_1 = \dots = \mathcal{R}_N$) to zero-sum competition ($\sum \mathcal{R}_i = 0$), and mixed-motive scenarios necessitating negotiation.

Similar to the single-agent trajectory, Table \ref{tab:marl_envs_comprehensive} chronicles how the empirical instantiations of the SG framework have radically evolved, mirroring the field's shift from physical micromanagement to high-level cognitive sociology.

In the foundational MARL era (Parts I -- III), benchmarks were heavily anchored in spatial competition and embodied coordination. Environments such as \textbf{SMAC (StarCraft II)}~\cite{samvelyan2019starcraft} and \textbf{Google Research Football}~\cite{kurach2020google} challenged agents to optimize high-frequency spatial maneuvers under severe partial observability (e.g., fog of war). The algorithmic focus during this period was primarily on resolving the credit assignment problem in cooperative tasks (e.g., through value decomposition methods like QMIX~\cite{rashid2018qmix}) and managing the exponentially growing joint action space in rigid physical simulators like \textbf{Isaac Gym}~\cite{makoviychuk2021isaac} and \textbf{VMAS}~\cite{bettini2022vmas}.

However, the advent of LLM-based multi-agent systems (Parts IV \& V) has transposed the SG framework into purely semantic and socio-cognitive domains. As a significant experiment in mimicking human social behavior, the Generative Agents\cite{park2023generativeagentsinteractivesimulacra} pioneered the LLM-based agent paradigm and marked a shift in traditional multi-agent communication methods from Differentiable Communication to Natural Language Communication. This transformation signifies a shift in the information carrier for multi-agent cooperation in complex environments from early "feature signals", such as SMAC (StarCraft II)\cite{samvelyan2019starcraft} and MAgent\cite{zheng2017magentmanyagentreinforcementlearning}, to semantically meaningful human language. In cooperative scenarios (Part IV), environments like \textbf{ColBench}~\cite{zhou2025sweetrltrainingmultiturnllm} (in training SweetRL) and \textbf{Math Collaboration} frameworks~\cite{du2023improving} elevate joint actions from physical movements to modular cognitive workflows. Agents assume distinct personas (e.g., Coder, Reviewer, Planner) to jointly navigate massive software repositories or decompose complex mathematical theorems, fundamentally shifting the focus from spatial coordination to \textit{semantic alignment}~\cite{hong2023metagpt}.

Most profoundly, modern mixed-motive environments (Part V) such as \textbf{TextArena}~\cite{guertler2025textarena} and \textbf{SPIRAL}~\cite{liu2026spiralselfplayzerosumgames} utilize the SG topology to benchmark sophisticated human-like social interactions. Here, the challenge of non-stationarity is no longer about predicting a physical opponent's movement, but rather about modeling their mental state (Theory of Mind). These LLM agents must execute strategies involving trust-building, deception, distributive negotiation, and probabilistic reasoning within text-based constraints. Consequently, the modern MARL benchmark has evolved into a rigorous digital sandbox for evaluating the sociological and game-theoretic alignment of Foundation Models (e.g., through frameworks like MAGRPO~\cite{liu2025llmcollaborationmultiagentreinforcement}). This also signifies that the penetration rate of semantic information carriers in multi-agent systems is increasing.

\subsection{Multi-modal Span}
The dimension of \textit{Multi-modal Span} characterizes the heterogeneity and structural format of the information channels through which an agent perceives its environment. Based on the complexity and fusion of sensory inputs, we categorize RL tasks into three categories: \textit{Single-modality Paradigms} (e.g., purely visual pixel-to-control environments like Atari~\cite{mnih2015human}), \textit{Multi-modal Fusion} (e.g., instruction-guided embodied navigation requiring vision-language grounding~\cite{shridhar2020alfred}), and \textit{Structured \& Non-standard Modalities} (e.g., graph-based relational inputs or point-cloud spatial geometries~\cite{kool2019attention}).

\begin{table*}[htbp]
  \centering 
  \caption{Taxonomy of Representative Single-Modality Reinforcement Learning Environments}
  \label{tab:single_modal_envs}
  \renewcommand{\arraystretch}{1.15} 
  \begin{tabularx}{\linewidth}{@{} l l c l @{}}
    \toprule
    \textbf{Environment} & \textbf{Task / Domain} & \textbf{Year$^{\mathrm{a}}$} & \textbf{DOI / Source} \\
    \midrule
    
    \multicolumn{4}{@{}l}{\textbf{Part I: Low-Dimensional Numerical \& Proprioceptive (Vectors \& Physics)}} \\
    \midrule
    MuJoCo (Gym Control)\cite{todorov2012mujoco} & Continuous Physics Locomotion & 2012 & \href{https://doi.org/10.1109/IROS.2012.6386109}{IROS'2012} \\
    Box2D (LunarLander / Bipedal) & 2D Physics Control \& Actuation & 2016 & \href{https://github.com/erincatto}{github.com/erincatto} \\
    OpenAI Gym (Classic Control)\cite{brockman2016openai} & Standardized Tabular \& Physics MDPs & 2016 & \href{https://doi.org/10.48550/arXiv.1606.01540}{arXiv:1606.01540} \\
    DeepMind Control Suite (DMC)\cite{tassa2018deepmindcontrolsuite} & High-fidelity Continuous Locomotion & 2018 & \href{https://doi.org/10.48550/arXiv.1801.00690}{arXiv:1801.00690} \\
    Meta-World (State-based)\cite{yu2020meta_world} & Multi-task Robotic Manipulation & 2019 & \href{https://doi.org/10.48550/arXiv.1910.10897}{CoRL'19} \\
    SMAC (StarCraft II)\cite{vinyals2019grandmaster} & Low-dim Cooperative Micromanagement & 2019 & \href{https://doi.org/10.48550/arXiv.1902.04043}{arXiv:1902.04043} \\
    Google Research Football\cite{kurach2020google} & Simulated Soccer \& Strategy & 2019 & \href{https://doi.org/10.48550/arXiv.1907.11180}{arXiv:1907.11180} \\
    Hanabi Learning Env\cite{BARD2020103216} & Partially Observable Card Game & 2019 & \href{https://doi.org/10.1016/j.artint.2019.103216}{Artificial Intelligence} \\
    Robosuite\cite{zhu2020robosuite} & Modular Robotic Manipulation & 2020 & \href{https://doi.org/10.48550/arXiv.2009.12293}{arXiv:2009.12293} \\
    Brax\cite{freeman2021brax} & Hardware-Accelerated Rigid Physics & 2021 & \href{https://doi.org/10.48550/arXiv.2106.13281}{NeurIPS'21} \\
    \midrule
    
    \multicolumn{4}{@{}l}{\textbf{Part II: Visual Perception (Pure Pixel Inputs)}} \\
    \midrule
    Arcade Learning Env (ALE)\cite{bellemare2013arcade} & Atari 2600 (Visual Pixel Control) & 2013 & \href{https://doi.org/10.1613/jair.3912}{10.1613/jair.3912} \\
    ViZDoom\cite{kempka2016vizdoom} & 3D First-Person Visual Perception & 2016 & \href{https://doi.org/10.1109/CIG.2016.7860433}{CIG'2016} \\
    DeepMind Lab (DMLab)\cite{beattie2016deepmind} & 3D Navigation \& Spatial Puzzles & 2016 & \href{https://doi.org/10.48550/arXiv.1612.03801}{arXiv:1612.03801} \\
    MiniGrid (Visual)\cite{chevalier2018minimalistic} & Procedural 2D Grid-world Navigation & 2018 & \href{https://doi.org/10.48550/arXiv.2306.13831}{arXiv:2306.13831} \\
    CoinRun\cite{pmlr-v97-cobbe19a} & Procedural Visual Generalization & 2019 & \href{https://proceedings.mlr.press/v97/cobbe19a.html}{ICML'19} \\
    Animal-AI Environment\cite{beyret2019animalaienvironmenttrainingtesting} & Visual Cognitive \& Physics Testing & 2019 & \href{https://doi.org/10.48550/arXiv.1909.07483}{	arXiv:1909.07483} \\
    ProcGen\cite{cobbe2020leveraging} & Procedurally Generated 2D Games & 2019 & \href{https://doi.org/10.48550/arXiv.1912.01588}{arXiv:1912.01588} \\
    Atari 100k\cite{kaiser2024modelbasedreinforcementlearningatari} & Data-efficient Visual RL & 2020 & \href{https://doi.org/10.48550/arXiv.1903.00374}{arXiv:1903.00374} \\
    NetHack Learning Env (NLE)\cite{kuttler2020nethack} & Complex Procedural Roguelike Visuals & 2020 & \href{https://doi.org/10.48550/arXiv.2006.13760}{arXiv:2006.13760} \\
    Crafter\cite{hafner2022benchmarkingspectrumagentcapabilities} & Open-ended Visual Survival & 2021 & \href{https://doi.org/10.48550/arXiv.2109.06780}{arXiv:2109.06780} \\
    \midrule
    
    \multicolumn{4}{@{}l}{\textbf{Part III: Language \& Audio (Textual Reasoning \& Digital Dialog)}} \\
    \midrule
    TextWorld\cite{cote2018textworld} & Procedural Text-Based Games & 2018 & \href{https://doi.org/10.48550/arXiv.1806.11532}{	arXiv:1806.11532} \\
    BabyAI\cite{chevalierboisvert2019babyaiplatformstudysample} & Grounded Language Navigation & 2019 & \href{https://doi.org/10.48550/arXiv.1810.08272}{arXiv:1810.08272} \\
    Jericho\cite{hausknecht2020interactivefictiongamescolossal} & Interactive Fiction \& Semantic Reasoning & 2019 & \href{https://doi.org/10.48550/arXiv.1909.05398}{arXiv:1909.05398} \\
    GSM8K\cite{cobbe2021training} & Grade School Math Word Problems & 2021 & \href{https://doi.org/10.48550/arXiv.2110.14168}{arXiv:2110.14168} \\
    MATH / MATH-500\cite{hendrycks2021measuring} & Competition-Level Math Reasoning & 2021 & \href{https://doi.org/10.48550/arXiv.2103.03874}{arXiv:2103.03874} \\
    HH-RLHF (Anthropic)\cite{bai2022training} & Helpful \& Harmless Dialogue Alignment & 2022 & \href{https://doi.org/10.48550/arXiv.2204.05862}{arXiv:2204.05862} \\
    WebShop\cite{yao2022webshop} & Text-based E-commerce Navigation & 2022 & \href{https://doi.org/10.48550/arXiv.2207.01206}{arXiv:2207.01206} \\
    AlphaCode (CodeForces)\cite{doi:10.1126/science.abq1158} & Complex Algorithmic Text Processing & 2022 & \href{https://doi.org/10.1126/science.abq1158}{Science'22} \\
    ReAct (Interactive QA)\cite{yao2023reactsynergizingreasoningacting} & Multi-hop QA via Text Search Engine & 2022 & \href{https://doi.org/10.48550/arXiv.2210.03629}{arXiv:2210.03629} \\
    ToolLLM\cite{qin2023toolllm} & Complex Tool-Use \& API Execution & 2023 & \href{https://doi.org/10.48550/arXiv.2307.16789}{arXiv:2307.16789} \\
    GPQA\cite{rein2023gpqagraduatelevelgoogleproofqa} & Graduate-Level Scientific Reasoning & 2023 & \href{https://doi.org/10.48550/arXiv.2311.12022}{arXiv:2311.12022} \\
    LiveCodeBench\cite{jain2024livecodebenchholisticcontaminationfree} & Contamination-Free Code Generation & 2024 & \href{https://doi.org/10.48550/arXiv.2403.07974}{arXiv:2403.07974} \\
    MMLU-Pro\cite{wang2024mmluprorobustchallengingmultitask} / SuperGPQA\cite{pteam2025supergpqascalingllmevaluation} & Advanced General Domain QA & 2024 & \href{https://doi.org/10.48550/arXiv.2406.01574}{arXiv:2406.01574} \\
    TextArena\cite{guertler2025textarena} & Multi-Agent Text Games \& Negotiation & 2024 & \href{https://doi.org/10.48550/arXiv.2504.11442}{arXiv:2504.11442} \\
    ProcessBench\cite{zheng2025processbenchidentifyingprocesserrors} & Step-level Reasoning Verification & 2024 & \href{https://doi.org/10.48550/arXiv.2412.06559}{arXiv:2412.06559} \\
    Countdown (DeepSeek-R1)\cite{deepseek2025r1} & Pure Math \& Logic Planning & 2025 & \href{https://doi.org/10.48550/arXiv.2501.12948}{arXiv:2501.12948} \\
    DeepScaleR Dataset\cite{deepscaler2025} & RL Scaling for Mathematical Reasoning & 2025 & \href{https://huggingface.co/datasets/agentica-org/DeepScaleR-Preview-Dataset}{Github: DeepScaleR} \\
    \bottomrule
  \end{tabularx}
  \vspace{0ex}
  \footnotesize{\textbf{Note:} Single-modality paradigms have historically progressed from scalar vectors and raw pixels to natural language tokens, driving the evolution from classic control to modern LLM System 2 reasoning benchmarks.}
\end{table*}

\subsubsection{Single-modality Paradigms}
This category encompasses environments where the agent relies on a homogeneous data source. As delineated in Table \ref{tab:single_modal_envs}, while the dimensionality and semantic density of these environments vary drastically—from raw kinematic vectors to advanced natural language tokens—the input stream remains conceptually uniform. We categorize these paradigms into three primary evolutionary stages, supplemented by specialized temporal signals.

\paragraph{Low-Dimensional Numerical \& Proprioceptive States (Part I)}

The foundational benchmarks in RL typically employ \textit{Feature-based State Representations} grounded in physics and kinematics. In classic control tasks (e.g., CartPole \cite{sutton2018reinforcement}) and continuous locomotion simulators such as MuJoCo \cite{todorov2012mujoco} or the DeepMind Control Suite\cite{tassa2018deepmind}, observations are compact, continuous vectors describing proprioceptive states (e.g., joint angles, angular velocities, and center-of-mass dynamics). 

Beyond passive state vectors, many embodied environments explicitly demand \textit{Actuation-aware Control}. In robotics-oriented platforms like Brax\cite{freeman2021brax} or real-world manipulation simulations, agents must reason over both motor commands and actuator states, including torque limits, friction, and control frequency. In these paradigms, agents must learn stable policies without direct visual perception, relying entirely on low-level physical feedback and temporal coordination.

\paragraph{Visual Perception (Part II)}

With the advent of Deep RL, observations expanded to high-dimensional \textit{Pixel-level Representations}, forcing agents to learn spatial feature extractors (e.g., CNNs) concurrently with control policies.
\begin{itemize}[leftmargin=*, itemsep=2pt, topsep=4pt]
    \item \textbf{RGB \& Procedural Video:} Seminal suites such as the Arcade Learning Environment (ALE) \cite{bellemare2013arcade} utilize raw 2D RGB frames. Later environments like Procgen \cite{cobbe2020leveraging} introduced procedural generation to benchmark visual generalization, while NetHack (NLE) pushed the limits of complex, symbol-rich visual observation arrays\cite{kuttler2020nethack}.
    \item \textbf{Egocentric \& 3D Spatial:} In 3D navigation and survival tasks like ViZDoom \cite{kempka2016vizdoom} and DeepMind Lab\cite{beattie2016deepmind}, agents rely on \textit{First-Person Visual Observations}. In advanced embodied AI (e.g., PointGoal Navigation), these are often augmented with depth-based visual input (RGB-D) to infer spatial occupancy and complex geometry.
\end{itemize}

\paragraph{Language \& Audio Modalities (Part III)}

Beyond spatial and physical inputs, the environment state can be formulated purely through text or acoustic signals. The language modality, in particular, has seen the most dramatic evolution in recent years, driving the transition towards System 2 cognitive benchmarks.
\begin{itemize}[leftmargin=*, itemsep=2pt, topsep=4pt]
    \item \textbf{Textual Reasoning \& Digital Dialog:} Early environments like TextWorld \cite{cote2018textworld} presented states purely as \textit{Natural Language Instructions}, requiring reading comprehension and combinatorial action parsing. Today, this paradigm has exploded into LLM-driven reasoning benchmarks. Environments such as MATH~\cite{hendrycks2021measuring}, GSM8K~\cite{cobbe2021training}, and LiveCodeBench~\cite{jain2024livecodebenchholisticcontaminationfree} cast the state space as formal mathematical theorems or codebase logic, while interactive environments (e.g., ReAct~\cite{yao2023reactsynergizingreasoningacting}, ToolLLM~\cite{qin2023toolllm}) require agents to navigate text-based API queries and auto-regressive token generation.
    \item \textbf{Audio-based Navigation:} Environments such as SoundSpaces \cite{chen2020soundspaces} introduce \textit{Acoustic Signal Input}, in which agents must navigate or interact based solely on the spatial intensity, reverberation, and frequency of sound sources, completely bypassing visual rendering.
\end{itemize}

\paragraph{Time-Series \& Sequential Signals}
In highly specialized applied domains, such as quantitative trading (e.g., FinRL \cite{liu2020finrl}) or healthcare (e.g., sepsis treatment \cite{raghu2017deep}), the environment state is defined by \textit{Temporal Numerical Signals} (e.g., physiological time-series or financial indicators). These environments bypass spatial or semantic complexities, focusing entirely on capturing long-term temporal dependencies and non-stationary stochastic trends.

\begin{table*}[htbp]
  \centering
  \caption{Taxonomy of Representative Multi-Modal and Structured Reinforcement Learning Environments}
  \label{tab:multi_modal_envs}
  \renewcommand{\arraystretch}{1.15} 
  \begin{tabularx}{\linewidth}{@{} l l c X @{}}
    \toprule
    \textbf{Environment} & \textbf{Task / Domain} & \textbf{Year$^{\mathrm{a}}$} & \textbf{DOI / Source} \\
    \midrule
    
    \multicolumn{4}{@{}l}{\textbf{Part I: Vision-Language Fusion (Web, GUI \& Desktop Agents)}} \\
    \midrule
    ChartQA (Chart-to-Code)\cite{masry2022chartqabenchmarkquestionanswering} & Generating Plot Code from Image & 2022 & \href{https://doi.org/10.48550/arXiv.2203.10244}{arXiv:2203.10244} \\
    WebShop\cite{yao2022webshop} & Simulated E-commerce Web Navigation & 2022 & \href{https://doi.org/10.48550/arXiv.2207.01206}{NeurIPS'22} \\
    Mind2Web\cite{deng2023mind2web} & Generalist Web Agent in the Wild & 2023 & \href{https://doi.org/10.48550/arXiv.2306.06070}{NeurIPS'23} \\
    WebArena\cite{zhou2023webarena} & Highly Realistic Web Agent Execution & 2023 & \href{https://doi.org/10.48550/arXiv.2307.13854}{ICLR'24} \\
    MathVista / MathVision\cite{lu2024mathvistaevaluatingmathematicalreasoning} & Visual Mathematical Reasoning & 2023 & \href{https://doi.org/10.48550/arXiv.2310.02255}{ICLR'24} \\
    Ferret (Visual Grounding)\cite{you2023ferret} & Spatial Grounding \& UI Target Selection & 2023 & \href{https://doi.org/10.48550/arXiv.2310.07704}{ICLR'24} \\
    MMMU / MMMU-Pro\cite{yue2024mmmumassivemultidisciplinemultimodal} & Massive Multi-discipline Multimodal QA & 2023 & \href{https://doi.org/10.48550/arXiv.2311.16502}{CVPR'24} \\
    VisualWebArena\cite{koh2024visualwebarenaevaluatingmultimodalagents} & Multimodal Web Environment & 2024 & \href{https://doi.org/10.48550/arXiv.2401.13649}{ACL'24} \\
    V-IRL\cite{yang2024virl} & Visual Navigation with Language Query & 2024 & \href{https://doi.org/10.48550/arXiv.2402.03310}{arXiv:2402.03310} \\
    OSWorld\cite{xie2024osworld} & Multimodal Desktop OS Automation & 2024 & \href{https://doi.org/10.48550/arXiv.2404.07972}{arXiv:2404.07972} \\
    Video-MME (VideoQA)\cite{fu2025videommefirstevercomprehensiveevaluation} & Spatio-temporal Video Reasoning & 2024 & \href{https://doi.org/10.48550/arXiv.2405.21075}{arXiv:2405.21075} \\
    AndroidWorld\cite{rawles2024androidworld} & Mobile GUI Control \& Interaction & 2024 & \href{https://doi.org/10.48550/arXiv.2405.14573}{arXiv:2405.14573} \\
    \midrule
    
    \multicolumn{4}{@{}l}{\textbf{Part II: Visual-Proprioceptive \& Multi-Sensor Fusion (Embodied AI)}} \\
    \midrule
    CARLA Simulator\cite{dosovitskiy2017carla} & Trajectory Planning via Multi-sensor & 2017 & \href{https://doi.org/10.48550/arXiv.1711.03938}{CoRL'17} \\
    AI2-THOR\cite{kolve2017ai2} & Interactive 3D Object Manipulation & 2017 & \href{https://doi.org/10.48550/arXiv.1712.05474}{arXiv:1712.05474} \\
    VirtualHome\cite{puig2018virtualhomesimulatinghouseholdactivities} & Complex Household Activity Sequences & 2018 & \href{https://doi.org/10.48550/arXiv.1806.07011}{CVPR'18} \\
    Habitat\cite{savva2019habitat} & Photorealistic 3D Visual Navigation & 2019 & \href{https://arxiv.org/abs/1904.01201}{ICCV'19} \\
    iGibson\cite{shen2021igibson} & High-Fidelity Interactive Sim \& Physics & 2021 & \href{https://github.com/StanfordVL/iGibson}{Github: iGibson} \\
    MineDojo (Minecraft)\cite{fan2022minedojobuildingopenendedembodied} & Open-Ended Embodied Vision-Action & 2022 & \href{https://doi.org/10.48550/arXiv.2206.08853}{NeurIPS'22} \\
    ManiSkill 2 (Visual)\cite{gu2023maniskill2} & Visual-based Robotic Pick-and-Place & 2023 & \href{https://doi.org/10.48550/arXiv.2302.04659}{ICLR'23} \\
    LIBERO\cite{liu2023libero} & Lifelong Robot Manipulation & 2023 & \href{https://doi.org/10.48550/arXiv.2306.03310}{NeurIPS'23} \\
    Mobile ALOHA\cite{fu2024mobile} & Multi-sensor Mobile Arm Control & 2024 & \href{https://doi.org/10.48550/arXiv.2401.02117}{arXiv:2401.02117} \\
    \midrule
    
    \multicolumn{4}{@{}l}{\textbf{Part III: Symbolic, Logic \& Executable Code (Structured Non-Standard)}} \\
    \midrule
    HOList\cite{bansal2019holistenvironmentmachinelearning} & Higher-Order Logic Theorem Proving & 2019 & \href{https://doi.org/10.48550/arXiv.1904.03241}{ICML'19} \\
    Lean 4\cite{moura2021lean} / MiniF2F\cite{zheng2021minif2f} & Formal Theorem Proving & 2021 & \href{https://doi.org/10.48550/arXiv.2109.00110}{ICLR'22} \\
    CompilerGym\cite{cummins2022compilergym} & Compiler Optimization \& Execution & 2021 & \href{https://doi.org/10.48550/arXiv.2109.08267}{CGO'22} \\
    BIRD (Text-to-SQL)\cite{li2023bird} & Database Schema Logical Mapping & 2023 & \href{https://doi.org/10.48550/arXiv.2305.03111}{NeurIPS'23} \\
    PrOntoQA\cite{saparov2023languagemodelsgreedyreasoners} & Pure Logic \& Symbolic Puzzles & 2023 & \href{https://arxiv.org/abs/2210.01240}{arXiv:2210.01240} \\
    InterCode\cite{yang2023intercode} & Interactive Coding \& Execution Env & 2023 & \href{https://doi.org/10.48550/arXiv.2306.14898}{NeurIPS'23} \\
    VerilogEval (RTL Gen)\cite{liu2023verilogeval} & Hardware Description Logic Synth & 2023 & \href{https://doi.org/10.48550/arXiv.2309.07544}{arXiv:2309.07544} \\
    SWE-bench\cite{jimenez2023swebench} / SWE-Gym\cite{pan2025trainingsoftwareengineeringagents} & Real-World Software Eng. (GitHub) & 2023 & \href{https://doi.org/10.48550/arXiv.2310.06770}{ICLR'24} \\
    OpenCodeInterpreter\cite{zheng2024opencodeinterpreter} & Sandboxed Execution Interaction & 2024 & \href{https://doi.org/10.48550/arXiv.2402.14658}{arXiv:2402.14658} \\
    ColBench (SweetRL)\cite{zhou2025sweetrltrainingmultiturnllm} & LLM Collaborative Software Eng. & 2025 & \href{https://doi.org/10.48550/arXiv.2503.15478}{arXiv.2503.15478} \\
    \midrule
    
    \multicolumn{4}{@{}l}{\textbf{Part IV: Graph-Structured \& Complex Domain-Specific Fusion}} \\
    \midrule
    MolDQN\cite{zhou2019optimization} & Molecule Generation via Graph RL & 2019 & \href{https://doi.org/10.1038/s41598-019-47148-x}{Nature Sci Rep'19} \\
    CityLearn\cite{vazquezcanteli2020citylearnstandardizingresearchmultiagent} & Urban Energy Management Optimization & 2020 & \href{https://arxiv.org/abs/2012.10504}{arXiv:2012.10504} \\
    TSP (Graph-based RL)\cite{kool2019attention} & Reasoning on Routing / TSP Graphs & 2020 & \href{https://doi.org/10.48550/arXiv.1803.08475}{arXiv:1803.08475} \\
    Geneformer\cite{theodoris2023transfer} & Bioinformatics \& Pathway Logic & 2023 & \href{https://doi.org/10.1038/s41586-023-06139-9}{Nature'23} \\
    Med-PaLM (Clinical)\cite{singhal2023large} & Graph-structured / EHR Reasoning & 2023 & \href{https://doi.org/10.48550/arXiv.2305.09617}{Nature'23} \\
    AlphaGeometry\cite{trinh2024alphageometry} & Neuro-symbolic Geometry Theorem & 2024 & \href{https://doi.org/10.1038/s41586-023-06747-5}{Nature'24} \\
    NPPC Gym\cite{yang2025nppc} & Nondeterministic Polynomial-time Problem & 2025 & \href{https://doi.org/10.48550/arXiv.2504.11239}{arXiv:2504.11239} \\
    MLE-bench\cite{chan2024mle} & Kaggle Auto-ML (Tabular \& Mixed) & 2024 & \href{https://doi.org/10.48550/arXiv.2410.07095}{arXiv:2410.07095} \\
    \bottomrule
  \end{tabularx}
  \vspace{0ex}
  \footnotesize{\textbf{Note:} Multi-modal and structured environments test an agent's ability to fuse disparate data types (e.g., grounding language into web HTML DOM trees, parsing visual UI screenshots, or logically proving theorems via formal symbolic execution).}
\end{table*}

\subsubsection{Multi-Modal and Structured Paradigms}
As reinforcement learning progresses toward generalist artificial intelligence, environments increasingly demand the synthesis of heterogeneous data streams and non-Euclidean topologies. As outlined in Table \ref{tab:multi_modal_envs}, these environments force agents to perform cross-modal alignment, semantic grounding, and formal logical deduction. We categorize these advanced paradigms into four distinct domains.

\begin{figure}[htbp] 
    \centering
    \includegraphics[width=\columnwidth]{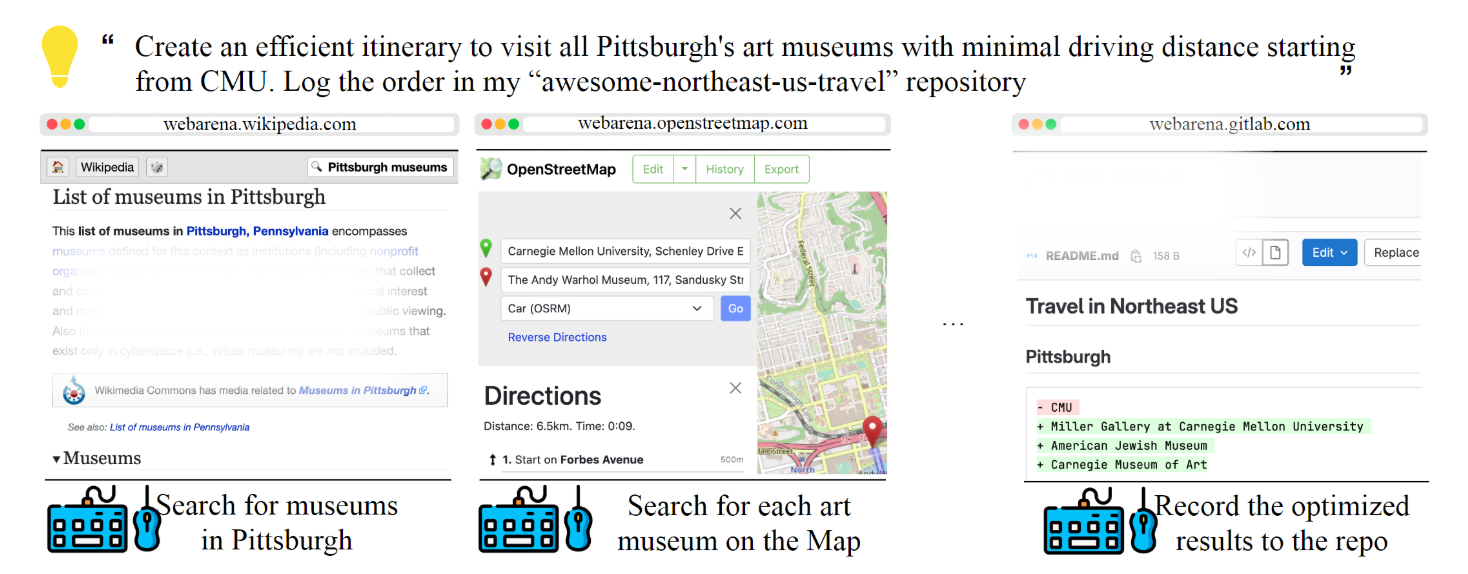} 
    \caption{\textbf{WebArena: The Frontier of Vision-Language-Action (VLA) Fusion.} Representing the modern multi-modal landscape, this environment requires agents to ground open-ended natural language instructions into dense visual interfaces. It forces a complex synthesis of image-based visual reasoning, structural analysis of HTML DOM trees, and auto-regressive text generation to execute executable actions. Source: \href{https://webarena.dev/}{webarena.dev}}
    \label{fig:multi_modal_webarena}
\end{figure}

\paragraph{Vision-Language Fusion: Web, GUI \& Desktop Agents (Part I)}
The frontier of digital AI lies in \textit{Vision-Language-Action (VLA)} models interacting with complex human interfaces. Unlike traditional spatial navigation, environments like WebArena \cite{zhou2023webarena} (See Figure \ref{fig:multi_modal_webarena}), Mind2Web \cite{deng2023mind2web}, and OSWorld \cite{xie2024osworld} require agents to ground abstract natural language instructions into dense visual interfaces (e.g., HTML DOM trees, browser screenshots, or desktop icons). In these settings, such as ChartQA \cite{masry2022chartqabenchmarkquestionanswering} or Ferret \cite{you2023ferret}, the agent must execute \textit{Spatial Grounding \& UI Target Selection}, translating a visual-semantic understanding of the screen into precise actionable coordinates or executable code. This multimodal fusion bridges the gap between passive image captioning and active digital manipulation.

\paragraph{Visual-Proprioceptive \& Multi-Sensor Fusion (Part II)}
In the realm of Embodied AI, agents must navigate and manipulate the physical (or simulated) world by fusing exteroceptive and proprioceptive signals. 
\begin{itemize}[leftmargin=*, itemsep=2pt, topsep=4pt]
    \item \textbf{Visual-Motor Perception:} In robotic manipulation benchmarks (e.g., ManiSkill 2 \cite{gu2023maniskill2}, LIBERO \cite{liu2023libero}, and Mobile ALOHA \cite{fu2024mobile}), the agent receives a composite observation of \textit{Egocentric Vision} (camera feed) and \textit{Proprioceptive State} (gripper position, torque). This demands fusing allocentric visual cues with ego-centric kinematic data to perform long-horizon physical tasks.
    \item \textbf{Heterogeneous Sensor Fusion:} In autonomous driving and interactive 3D simulations (e.g., CARLA \cite{dosovitskiy2017carla}, iGibson \cite{shen2021igibson}, AI2-THOR \cite{kolve2017ai2}), agents process massive \textit{Heterogeneous Modal Inputs}. Fusing LiDAR point clouds, RGB-D cameras, and GPS signals is critical for robust perception under varying physical dynamics and environmental stochasticity.
\end{itemize}

\paragraph{Symbolic, Logic \& Executable Code (Part III)}
This category represents a fundamental departure from continuous numerical inputs, requiring the agent to operate within strict syntactic and logical constraints.
\begin{itemize}[leftmargin=*, itemsep=2pt, topsep=4pt]
    \item \textbf{Interactive Code Execution:} In software engineering benchmarks like SWE-bench \cite{jimenez2023swebench}, OpenCodeInterpreter \cite{zheng2024opencodeinterpreter}, and InterCode \cite{yang2023intercode}, the environment is a sandboxed compiler or terminal. The state is represented by source code and execution error logs, requiring the agent to perform \textit{Programmatic Reasoning} and iterative debugging.
    \item \textbf{Formal Logic \& Symbolic Reasoning:} Environments such as Lean 4 \cite{moura2021lean} / MiniF2F \cite{zheng2021minif2f} and PRONTOQA \cite{saparov2023languagemodelsgreedyreasoners} utilize \textit{Symbolic \& Logic Representations}. These Neuro-symbolic RL approaches force the agent to explore formal theorem proving trees or hardware description logic (e.g., VerilogEval \cite{liu2023verilogeval}), where a single syntactical error results in failure, demanding absolute deductive precision.
\end{itemize}

\paragraph{Graph-Structured \& Complex Domain Fusion (Part IV)}
For scientific discovery and complex system optimization, states are often inherently topological and non-Euclidean, necessitating \textit{Graph-structured Representations} (often processed via Graph Neural Networks).
\begin{itemize}[leftmargin=*, itemsep=2pt, topsep=4pt]
    \item \textbf{Combinatorial \& Relational:} In tasks such as the Traveling Salesperson Problem (TSP) \cite{kool2019attention} or chip floorplanning \cite{mirhoseini2021graph}, the state is encoded as a graph, requiring the agent to capture permutation-invariant relational dependencies between nodes (e.g., NPPC Gym \cite{yang2025nppc}).
    \item \textbf{Scientific \& Biochemical:} Advanced environments apply RL to rare-event random walks in \textit{Molecular Dynamics} \cite{zhou2019optimization}, graph-based molecule generation (MolDQN \cite{zhou2019optimization}), or bioinformatics pathways (Geneformer \cite{theodoris2023transfer}). Similarly, deep integration with clinical knowledge graphs (e.g., Med-PaLM \cite{singhal2023large}) represents the pinnacle of fusing graph-theoretic structures with domain-specific semantic reasoning.
\end{itemize}

\paragraph{Latent Models \& Human Preference Signals (Ancillary Modalities)}
Beyond explicit environmental inputs, modern RL architectures frequently incorporate implicit or external signals to shape the state or reward manifold. In model-based RL (e.g., Dreamer \cite{hafner2019dream}), the agent operates on a \textit{Latent State Representation} derived from a learned world model. Concurrently, in RLHF paradigms \cite{ouyang2022training}, the environment's objective is fundamentally redefined by \textit{Interactive Human Feedback}. This preference signal serves as a distinct, external modality that guides the agent toward alignment with human intent, bypassing the need for sparse, hand-engineered reward functions.

\subsection{Agent Capabilities}
The dimension of \textit{Agent Capabilities} classifies tasks based on the specific cognitive, physical, or computational competencies required to solve them. Unlike modality, which concerns the format of input data, this dimension strictly defines the functional "skill set" an agent must possess. As illustrated in Table \ref{tab:agent_capabilities_envs}, this spectrum has undergone a massive paradigm shift—evolving from low-level motor control in classical physical simulators to the high-level, System 2 strategic reasoning demanded by modern Large Language Models (LLMs). We organize these requisite capabilities into eight core categories.

\begin{figure*}[h]
    \centering
    \includegraphics[width=0.75\textwidth]{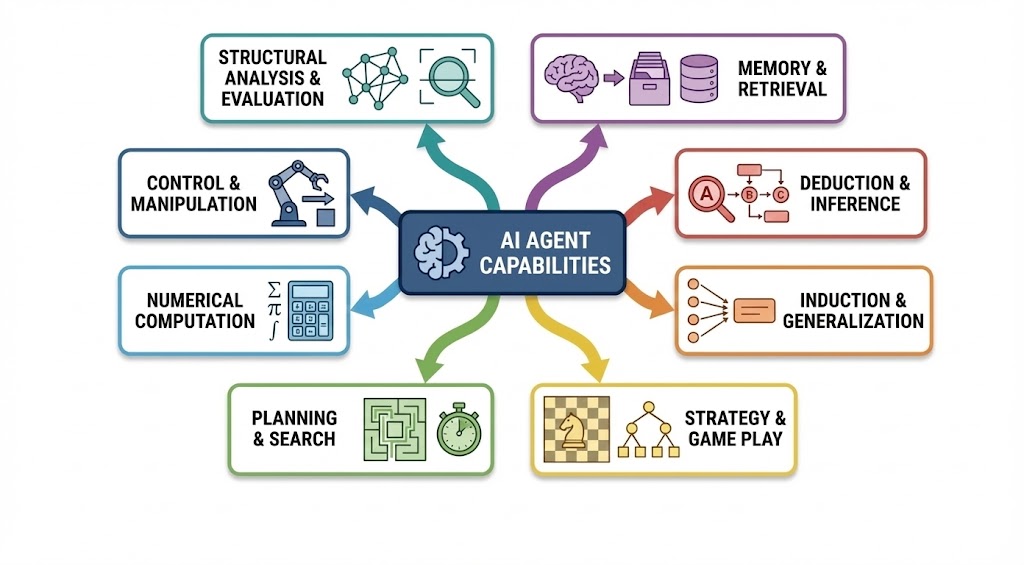}
    \caption{The multi-dimensional landscape of requisite agent capabilities. The diagram illustrates the diverse skill set necessary for generalist agents, bridging the gap between physical interaction (Control, Strategy) and abstract cognitive processes (Deduction, Planning, Structural Analysis).}
    \label{fig:2}
\end{figure*}

\paragraph{Control \& Manipulation (Part I)}
These tasks demand high-frequency, high-precision continuous control in physical or simulated environments. The primary challenge is handling high degrees of freedom (DoF), contact dynamics, and low-level motor torques. Foundational benchmarks like MuJoCo locomotion~\cite{todorov2012mujoco} and modern robotic manipulation suites (e.g., ManiSkill 2~\cite{gu2023maniskill2}, Mobile ALOHA~\cite{fu2024mobile}) evaluate the agent's ability to optimize robust policies under complex physical constraints, translating sensory input directly into physical actuation.

\paragraph{Strategy \& Game Play (Part II)}
Strategic tasks involve adversarial dynamics, imperfect information, and long-term credit assignment. From early arcade environments (ALE~\cite{bellemare2013arcade}) to complex multi-agent simulations (StarCraft II~\cite{vinyals2019grandmaster}, Google Research Football~\cite{kurach2020google}), the complexity arises from the combinatorial explosion of the action space and the need to model opponents (Theory of Mind). In the LLM era, this capability has expanded into socio-cognitive domains, requiring agents to execute deception, negotiation, and game-theoretic alignment in environments like TextArena~\cite{guertler2025textarena}.

\paragraph{Planning \& Search (Part III)}
While Strategy focuses on adversaries, Planning focuses on model-based look-ahead in complex dynamics. Agents must simulate future trajectories using an internal world model or an external simulator. Traditional environments like Sokoban~\cite{guez2019investigation} demand discrete spatial planning. However, modern digital benchmarks (e.g., WebArena~\cite{zhou2023webarena}) and logic exploration games (e.g., Game24~\cite{yao2023tree}) require agents to integrate learned value functions with advanced tree search algorithms (such as Monte Carlo Tree Search, MCTS~\cite{browne2012survey}) to optimize long-horizon, multi-step decisions before execution.

\paragraph{Deduction \& Inference (Part IV)}
This advanced category encompasses tasks requiring rigorous logical reasoning and relational inference. The agent must deduce hidden properties or formal causal relationships. Benchmarks such as the ARC-AGI corpus~\cite{chollet2019measure} test the limits of inductive abstraction. More recently, the focus has shifted heavily towards formal theorem proving (Lean 4~\cite{yang2023leandojo}) and step-level reasoning verification (ProcessBench~\cite{zheng2025processbenchidentifyingprocesserrors}), where agents must execute multi-step logic (e.g., Chain-of-Thought~\cite{wei2022chain}) and rigorously evaluate intermediate deduction steps without simple pattern matching.

\paragraph{Numerical Computation (Part V)}
This category isolates the algorithmic ability of agents to perform explicit arithmetic operations, precise resource quantification, and formal math solving. Moving far beyond early sequence-to-sequence arithmetic tasks, modern RL evaluates quantitative capability through competition-level mathematical benchmarks such as GSM8K~\cite{cobbe2021training}, MATH~\cite{hendrycks2021measuring}, and OlympiadBench~\cite{he2024olympiadbench}. In these settings, environments (e.g., Countdown Game~\cite{guo2025deepseek}) force the agent to frame complex arithmetic calculations and mathematical proofs as a sequential, verifiable decision-making process.

\paragraph{Structural Analysis \& Evaluation (Part VI)}

These tasks require the agent to parse, understand, and manipulate complex topological or syntactic structures. For example, Graph-based RL tasks (e.g., TSP~\cite{kool2019attention}) require analyzing permutation-invariant relational graphs. In the era of Foundation Models, this capability is prominently evaluated in Real-World Software Engineering environments (e.g., SWE-bench~\cite{jimenez2023swebench}, BIRD~\cite{li2023bird}) and GUI control (e.g., OSWorld~\cite{xie2024osworld}), where agents must structurally analyze GitHub repositories, database schemas, or intricate HTML DOM trees to generate logically correct code or execution sequences.

\paragraph{Induction \& Generalization (Part VII)}
Inductive tasks explicitly test the agent's ability to synthesize general rules from sparse examples (Few-Shot Learning) or transfer skills to unseen, procedurally generated environments (Zero-Shot Generalization). Benchmarks like Procgen\cite{cobbe2020leveraging}, NetHack\cite{kuttler2020nethack}, and Alchemy\cite{wang2021alchemybenchmarkanalysistoolkit} evaluate whether an agent possesses "meta-learning" capabilities, rapidly adapting its policy to novel task distributions and open-ended, shifting environmental dynamics without catastrophic forgetting.

\paragraph{Memory \& Retrieval (Part VIII)}
Tasks in this category are defined by severe \textit{Partial Observability} and knowledge-intensive requirements. The agent cannot rely solely on the current observation but must encode, store, and retrieve historical context to infer the underlying state. In early 3D navigation (e.g., Memory Maze~\cite{pasukonis2022evaluatinglongtermmemory3d}), this necessitated RNNs or LSTMs. Today, in complex interactive environments like WebShop~\cite{yao2022webshop}, ReAct~\cite{yao2023reactsynergizingreasoningacting}, and TAU-Bench~\cite{fang2024taubench}, agents must utilize tool-use (e.g., querying external search engines) and manage long-context multi-turn dialogue memory to bridge temporal gaps between information retrieval and final decision points.

\begin{table*}[htbp]
  \centering
  \small 
  \caption{Taxonomy of Representative Reinforcement Learning Environments by Core Agent Capabilities}
  \label{tab:agent_capabilities_envs}
  \renewcommand{\arraystretch}{1}
  \begin{tabularx}{\linewidth}{@{} l l c X @{}}
    \toprule
    \textbf{Environment} & \textbf{Task / Domain} & \textbf{Year$^{\mathrm{a}}$} & \textbf{DOI / Source} \\
    \midrule
    
    \multicolumn{4}{@{}l}{\textbf{Part I: Control \& Manipulation (Physical \& Embodied Interaction)}} \\
    \midrule
    MuJoCo (Gym Control)\cite{todorov2012mujoco} & Continuous Physics Locomotion & 2012 & \href{https://doi.org/10.1109/IROS.2012.6386109}{IROS'2012} \\
    Meta-World\cite{yu2020meta_world} & Multi-task Robotic Manipulation & 2019 & \href{https://doi.org/10.48550/arXiv.1910.10897}{CoRL'19} \\
    Safety Gym\cite{Ray2019} & Constrained MDPs \& Safe Exploration & 2019 & \href{https://github.com/openai/safety-gym}{GitHub: safetygym} \\
    ManiSkill 2\cite{gu2023maniskill2} & Visual-based Robotic Pick-and-Place & 2023 & \href{https://doi.org/10.48550/arXiv.2302.04659}{ICLR'23} \\
    LIBERO\cite{liu2023libero} & Lifelong Robot Manipulation & 2023 & \href{https://doi.org/10.48550/arXiv.2306.03310}{NeurIPS'23} \\
    Habitat 3.0\cite{puig2023habitat30cohabitathumans} & Interactive \& Human-Robot Synergy & 2023 & \href{https://doi.org/10.48550/arXiv.2310.13724}{arXiv:2310.13724} \\
    Mobile ALOHA\cite{fu2024mobile} & Multi-sensor Mobile Arm Control & 2024 & \href{https://doi.org/10.48550/arXiv.2401.02117}{arXiv:2401.02117} \\
    \midrule
    
    \multicolumn{4}{@{}l}{\textbf{Part II: Strategy \& Game Play (Adversarial \& Cooperative Dynamics)}} \\
    \midrule
    Arcade Learning Env (ALE)\cite{bellemare2013arcade} & Atari 2600 (Visual Pixel Control) & 2013 & \href{https://doi.org/10.1613/jair.3912}{10.1613/jair.3912} \\
    AlphaZero (Go/Chess/Shogi)\cite{silver2018general} & Self-play Board Games & 2017 & \href{https://doi.org/10.1038/nature24270}{10.1038/nature24270} \\
    Unity ML-Agents\cite{juliani2018unity} & General 3D Physics \& Multi-behavior Engine & 2018 & \href{https://doi.org/10.48550/arXiv.1809.02627}{arXiv:1809.02627} \\
    SMAC (StarCraft II)\cite{vinyals2019grandmaster} & Cooperative Micromanagement & 2019 & \href{https://doi.org/10.48550/arXiv.1902.04043}{arXiv:1902.04043} \\
    Google Research Football\cite{kurach2020google} & Simulated Soccer \& Strategy & 2019 & \href{https://doi.org/10.48550/arXiv.1907.11180}{arXiv:1907.11180} \\
    TextArena\cite{guertler2025textarena} & Multi-Agent Text Negotiation & 2024 & \href{https://doi.org/10.48550/arXiv.2504.11442}{	arXiv:2504.11442} \\
    \midrule
    
    \multicolumn{4}{@{}l}{\textbf{Part III: Planning \& Search (Long-Horizon \& Tree Exploration)}} \\
    \midrule
    VirtualHome\cite{puig2018virtualhomesimulatinghouseholdactivities} & Complex Household Activity Sequences & 2018 & \href{https://doi.org/10.48550/arXiv.1806.07011}{CVPR'18} \\
    MineDojo (Minecraft)\cite{fan2022minedojobuildingopenendedembodied} & Open-Ended Embodied Vision-Action & 2022 & \href{https://doi.org/10.48550/arXiv.2206.08853}{NeurIPS'22} \\
    Game24\cite{yao2023tree} & Logic \& Search Tree Exploration & 2023 & \href{https://doi.org/10.48550/arXiv.2305.10601}{arXiv:2305.10601} \\
    WebArena\cite{zhou2023webarena} & Highly Realistic Web Agent Execution & 2023 & \href{https://doi.org/10.48550/arXiv.2307.13854}{ICLR'24} \\
    Sokoban (Reasoning Gym)\cite{stojanovski2025reasoninggymreasoningenvironments} & Planning \& Search in Text Grid & 2025 & \href{https://doi.org/10.48550/arXiv.2505.24760}{arXiv:2505.24760} \\
    \midrule
    
    \multicolumn{4}{@{}l}{\textbf{Part IV: Deduction \& Inference (Logical, Scientific \& Visual Reasoning)}} \\
    \midrule
    ARC-AGI\cite{chollet2019measure} & Abstraction \& Inductive Reasoning & 2019 & \href{https://doi.org/10.48550/arXiv.1911.01547}{arXiv:1911.01547} \\
    Lean 4\cite{moura2021lean} / MiniF2F\cite{zheng2021minif2f} & Formal Theorem Proving & 2021 & \href{https://doi.org/10.48550/arXiv.2109.00110}{ICLR'22} \\
    GPQA\cite{rein2023gpqagraduatelevelgoogleproofqa} & Graduate-Level Scientific Reasoning & 2023 & \href{https://doi.org/10.48550/arXiv.2311.12022}{arXiv:2311.12022} \\
    MMMU / MMMU-Pro\cite{yue2024mmmumassivemultidisciplinemultimodal} & Massive Multi-discipline Multimodal QA & 2023 & \href{https://doi.org/10.48550/arXiv.2311.16502}{CVPR'24} \\
    ProcessBench\cite{zheng2025processbenchidentifyingprocesserrors} & Step-level Reasoning Verification & 2024 & \href{https://doi.org/10.48550/arXiv.2412.06559}{arXiv:2412.06559} \\
    \midrule
    
    \multicolumn{4}{@{}l}{\textbf{Part V: Numerical Computation (Mathematical \& Quantitative Solving)}} \\
    \midrule
    GSM8K\cite{cobbe2021training} & Grade School Math Word Problems & 2021 & \href{https://doi.org/10.48550/arXiv.2110.14168}{arXiv:2110.14168} \\
    MATH / MATH-500\cite{hendrycks2021measuring} & Competition-Level Math Reasoning & 2021 & \href{https://doi.org/10.48550/arXiv.2103.03874}{NeurIPS'21} \\
    OlympiadBench\cite{he2024olympiadbench} & Advanced Mathematical Problem-Solving & 2024 & \href{https://doi.org/10.48550/arXiv.2402.14008}{arXiv:2402.14008} \\
    Countdown Game\cite{deepseek2025r1} & Arithmetic Planning \& Target Search & 2025 & \href{https://doi.org/10.48550/arXiv.2501.12948}{arXiv:2501.12948} \\
    DeepScaleR Dataset\cite{deepscaler2025} & RL Scaling for Mathematical Reasoning & 2025 & \href{https://huggingface.co/datasets/agentica-org/DeepScaleR-Preview-Dataset}{Github: DeepScaleR} \\
    \midrule
    
    \multicolumn{4}{@{}l}{\textbf{Part VI: Structural Analysis \& Evaluation (Code, GUI \& Graphs)}} \\
    \midrule
    TSP (Graph-based RL)\cite{kool2019attention} & Reasoning on Routing / TSP Graphs & 2020 & \href{https://doi.org/10.48550/arXiv.1803.08475}{arXiv:1803.08475} \\
    BIRD (Text-to-SQL)\cite{li2023bird} & Database Schema Logical Mapping & 2023 & \href{https://doi.org/10.48550/arXiv.2305.03111}{NeurIPS'23} \\
    SWE-bench\cite{jimenez2023swebench} / SWE-Gym\cite{pan2025trainingsoftwareengineeringagents} & Real-World Software Eng. (GitHub) & 2023 & \href{https://doi.org/10.48550/arXiv.2310.06770}{ICLR'24} \\
    InterCode\cite{yang2023intercode} & Interactive Coding with Execution Feedback & 2023 & \href{https://doi.org/10.48550/arXiv.2306.14898}{NeurIPS'23} \\
    OSWorld\cite{xie2024osworld} & Multimodal Desktop OS Automation & 2024 & \href{https://doi.org/10.48550/arXiv.2404.07972}{arXiv:2404.07972} \\
    AndroidWorld\cite{rawles2024androidworld} & Mobile GUI Control \& Interaction & 2024 & \href{https://doi.org/10.48550/arXiv.2405.14573}{arXiv:2405.14573} \\
    \midrule
    
    \multicolumn{4}{@{}l}{\textbf{Part VII: Induction \& Generalization (Meta-RL \& Open-Ended Adaptation)}} \\
    \midrule
    Unity ML-Agents\cite{juliani2018unity} & General 3D Physics \& Multi-behavior & 2018 & \href{https://doi.org/10.48550/arXiv.1809.02627}{arXiv:1809.02627} \\
    ProcGen\cite{cobbe2020leveraging} & Procedurally Generated 2D Games & 2019 & \href{https://doi.org/10.48550/arXiv.1912.01588}{ICML'20} \\
    NetHack Learning Env\cite{kuttler2020nethack} & Deep Rogue-like Procedural Gen & 2020 & \href{https://doi.org/10.48550/arXiv.2006.13760}{NeurIPS'20} \\  D4RL\cite{fu2021d4rldatasetsdeepdatadriven} & Offline RL Benchmarking \& Dataset Eval & 2020 & \href{https://doi.org/10.48550/arXiv.2004.07219}{arXiv:2004.07219} \\
    Alchemy (DeepMind)\cite{wang2021alchemybenchmarkanalysistoolkit} & Meta-RL Open-Ended Generalization & 2021 & \href{https://doi.org/10.48550/arXiv.2102.02926}{arXiv:2102.02926} \\
    Crafter\cite{hafner2022benchmarkingspectrumagentcapabilities} & Open-ended Visual Survival & 2021 & \href{https://doi.org/10.48550/arXiv.2109.06780}{NeurIPS'21} \\
    MLE-bench\cite{chan2024mle} & Kaggle Auto-ML (Tabular \& Mixed) & 2024 & \href{https://doi.org/10.48550/arXiv.2410.07095}{arXiv:2410.07095} \\
    \midrule
    
    \multicolumn{4}{@{}l}{\textbf{Part VIII: Memory \& Retrieval (Knowledge-Intensive \& Multi-Turn State)}} \\
    \midrule
    ALFWorld\cite{shridhar2020alfworld} & Text-aligned Embodied Household Tasks & 2020 & \href{https://doi.org/10.48550/arXiv.2010.03768}{ICLR'21} \\
    Memory Maze\cite{pasukonis2022evaluatinglongtermmemory3d} & 3D Navigation with Long-Term Memory & 2022 & \href{https://doi.org/10.48550/arXiv.2210.13383}{arXiv:2210.13383} \\
    WebShop\cite{yao2022webshop} & Simulated E-commerce Web Navigation & 2022 & \href{https://doi.org/10.48550/arXiv.2207.01206}{NeurIPS'22} \\
    ReAct (Interactive QA)\cite{yao2023reactsynergizingreasoningacting} & Multi-hop QA via Text Search Engine & 2022 & \href{https://doi.org/10.48550/arXiv.2210.03629}{ICLR'23} \\
    TAU-Bench (Dual-Control)\cite{fang2024taubench} & Multi-Agent Interacting Tool Use & 2024 & \href{https://doi.org/10.48550/arXiv.2406.12045}{arXiv:2406.12045} \\
    \bottomrule
  \end{tabularx}
  \vspace{0ex} 
\end{table*}

\subsection{Observability: Information Completeness}
The dimension of \textit{Observability} characterizes the completeness of the information available to the agent regarding the global system state. From a game-theoretic and control perspective, this taxonomy distinguishes between environments of \textit{Perfect Information}, where the agent has omniscient access to the state dynamics, and \textit{Imperfect Information} (or Partial Observability), where spatial occlusion, digital compartmentalization, or adversarial concealment cloud the decision-making process.

\paragraph{Perfect Information Settings}
In tasks characterized by \textit{Perfect Information}, the agent's observation at any time step is isomorphic to the true environment state ($\mathcal{O}_t \equiv \mathcal{S}_t$). This scenario satisfies the strict Markov property, as no historical context or active exploration is required to disambiguate the current state. 
\begin{itemize}[leftmargin=*, itemsep=2pt, topsep=4pt]
    \item \textbf{Board Games \& Puzzles:} Classic combinatorial benchmarks exemplify this category. From traditional adversarial games like Chess, Shogi, and Go (solved by AlphaZero \cite{silver2018general}), to single-agent puzzle environments such as the Rubik's Cube (DeepCubeA \cite{agostinelli2019solving}), Hex, and Game24 \cite{yao2023tree}. Both players (or the sole agent) have full visibility of the board configuration, making the challenge purely computational (combinatorial tree search) rather than informational.
    \item \textbf{Fully Observable Simulations:} In standard robotic continuous control suites, the agent is privileged with exact, uncorrupted readings of the physical system. Examples include classical OpenAI Gym control (e.g., CartPole, Pendulum \cite{brockman2016openai}), MuJoCo locomotion tasks \cite{todorov2012mujoco}, and hardware-accelerated rigid physics simulators like Brax \cite{freeman2021brax} and Isaac Gym \cite{makoviychuk2021isaac} (when configured to expose full state tensors). The problem reduces strictly to trajectory optimization.
    \item \textbf{Formal Logic \& Mathematical Reasoning:} In modern cognitive benchmarks, environments evaluating formal theorem proving (e.g., Lean 4 \cite{moura2021lean}, MiniF2F \cite{zheng2021minif2f}), quantitative competition math (e.g., MATH \cite{hendrycks2021measuring}, GSM8K \cite{cobbe2021training}, OlympiadBench \cite{he2024olympiadbench}), arithmetic planning (e.g., Countdown Game \cite{deepseek2025r1}), and inductive abstraction (e.g., ARC-AGI \cite{chollet2019measure}) provide all necessary axioms and constraints upfront. The challenge concentrates entirely on deep, auto-regressive logical deduction rather than state estimation.
\end{itemize}

\paragraph{Imperfect Information Settings (POMDPs)}

Real-world complexity primarily stems from \textit{Imperfect Information}, where the agent observes only a partial, noisy, or localized projection of the global state ($\mathcal{O}_t \subset \mathcal{S}_t$). This corresponds to the Partially Observable Markov Decision Process (POMDP) formalism \cite{kaelbling1998planning}, requiring agents to maintain a \textit{belief state} or leverage long-context memory mechanisms.
\begin{itemize}[leftmargin=*, itemsep=2pt, topsep=4pt]
    \item \textbf{Spatial Occlusion \& Fog of War:} Physical line-of-sight obscurations force agents to actively scout and gather information. This is prevalent in complex strategy games (e.g., StarCraft II \cite{vinyals2019grandmaster}), First-Person Shooters (e.g., ViZDoom \cite{kempka2016vizdoom}), 3D embodied navigation (e.g., Habitat \cite{savva2019habitat}, Memory Maze \cite{pasukonis2022evaluatinglongtermmemory3d}), and procedurally generated partial-view grid-worlds (e.g., MiniGrid \cite{chevalier2018minimalistic}, OpenAI Hide-and-Seek \cite{baker2019emergent}).
    \item \textbf{Digital Compartmentalization \& Viewport Occlusion:}  In modern UI and Web environments, the agent's observation is restricted to the current HTML DOM fragment or visual screen viewport. In benchmarks like WebArena \cite{zhou2023webarena}, Mind2Web \cite{deng2023mind2web}, WebShop \cite{yao2022webshop}, OSWorld \cite{xie2024osworld}, and AndroidWorld \cite{rawles2024androidworld}, the agent cannot observe the entire website backend or hidden dropdown menus simultaneously; it must actively scroll, click, and navigate to reveal latent states.
    \item \textbf{Private Information \& Asymmetry:} Distinct portions of the state may be strictly private. Historically, this ranged from hidden cards in recreational games (e.g., No-Limit Texas Hold'em solved by Libratus \cite{brown2018superhuman}, Hanabi \cite{BARD2020103216}) to complex board games like Diplomacy (Cicero \cite{bakhtin2022human}). In the LLM era, this manifests in text-based game-theoretic arenas (e.g., TextArena \cite{guertler2025textarena}, Avalon and Werewolf social deduction environments \cite{xu2023exploring}), requiring complex Theory of Mind, trust-building, and deception management.
    \item \textbf{Stochastic Noise \& Systemic Opacity:} Imperfect information also arises from environmental stochasticity. In embodied autonomous driving (e.g., CARLA \cite{dosovitskiy2017carla}), it stems from sensor measurement noise \cite{yu2017preparing}. Analogously, in software engineering and data science benchmarks (e.g., SWE-bench \cite{jimenez2023swebench}, InterCode \cite{yang2023intercode}, MLE-bench \cite{chan2024mle}), the true state of a massive codebase or external database is overwhelmingly opaque. Agents must actively query the environment (e.g., via \texttt{grep}, SQL queries, or test scripts) to uncover hidden dependency conflicts, making debugging a highly iterative search process.
\end{itemize}

\vspace{1ex}
\noindent\textbf{Summary of Paradigm Shift:}
Collectively, the nature of observability has fundamentally expanded. While early environments primarily simulated physical constraints (e.g., sensor noise, map occlusion), modern benchmarks introduce digital and semantic opacity (e.g., hidden HTML nodes, vast code repositories, and socio-linguistic deception). We will systematically formalize and explore this evolutionary trajectory of environment design in Chapter 5.

\subsection{Application Domains}
While early reinforcement learning research primarily utilized synthetic game simulators to test algorithmic stability, the field has since exploded into a highly diversified ecosystem of application domains. As delineated in Table \ref{tab:app_domain_envs}, this trajectory reveals a clear evolutionary narrative: starting from isolated physical navigation and closed-system games, expanding into the semantic complexities of natural language, and ultimately converging on high-stakes scientific and industrial operations. We systematically categorize these environments into six distinct sectors.

\begin{figure*}[h]
    \centering
    \includegraphics[width=0.75\textwidth]{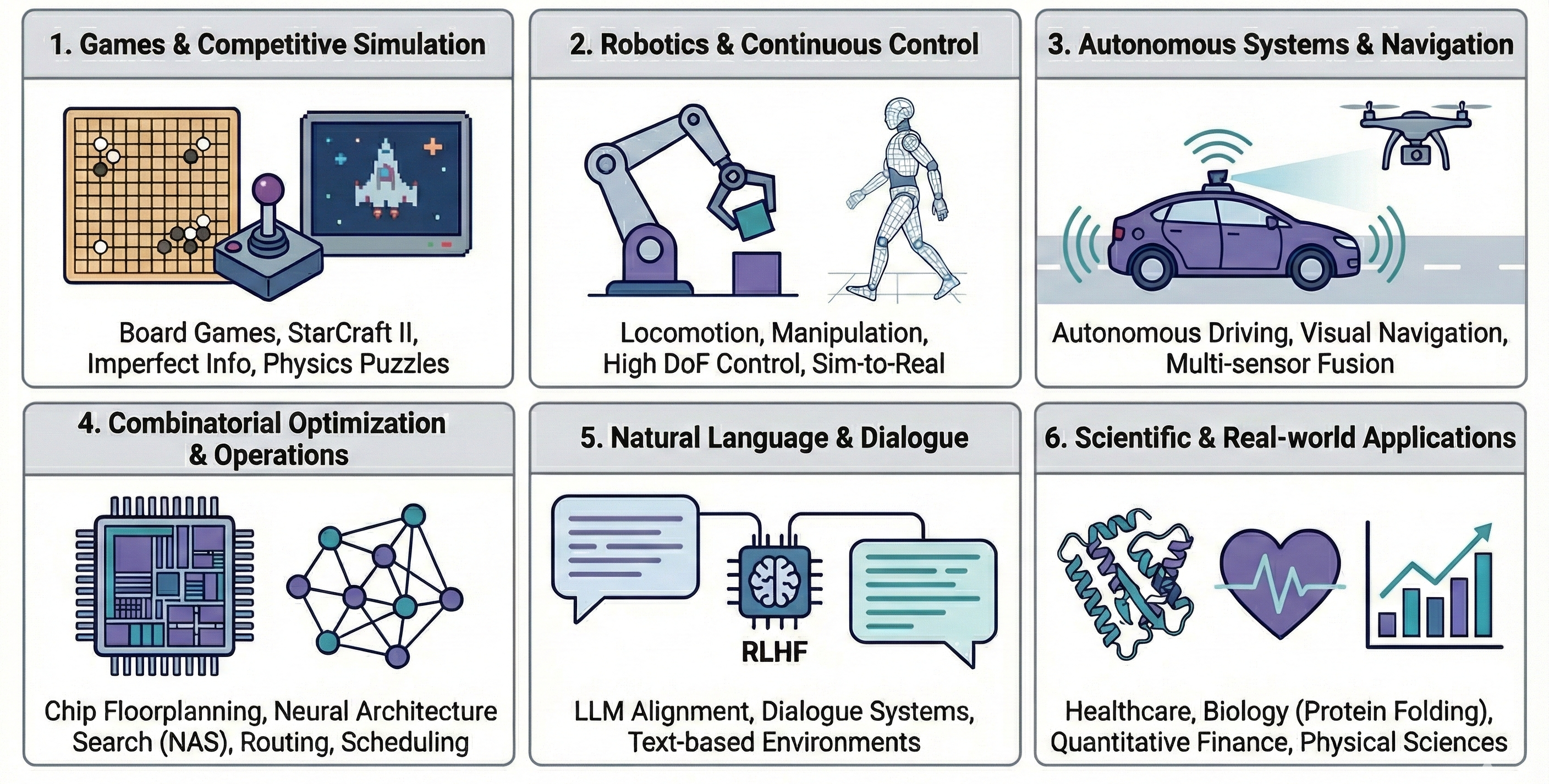}
    \caption{A taxonomic overview of diverse reinforcement learning application domains. The progression illustrates RL's expansion from simulated spatial environments (Navigation, Games) to abstract cognitive and structural systems (Language, Optimization, Science).}
    \label{fig:4}
\end{figure*}

\paragraph{1. Autonomous Systems \& Navigation}
\begin{figure}[htbp] 
    \centering
    \includegraphics[width=\columnwidth]{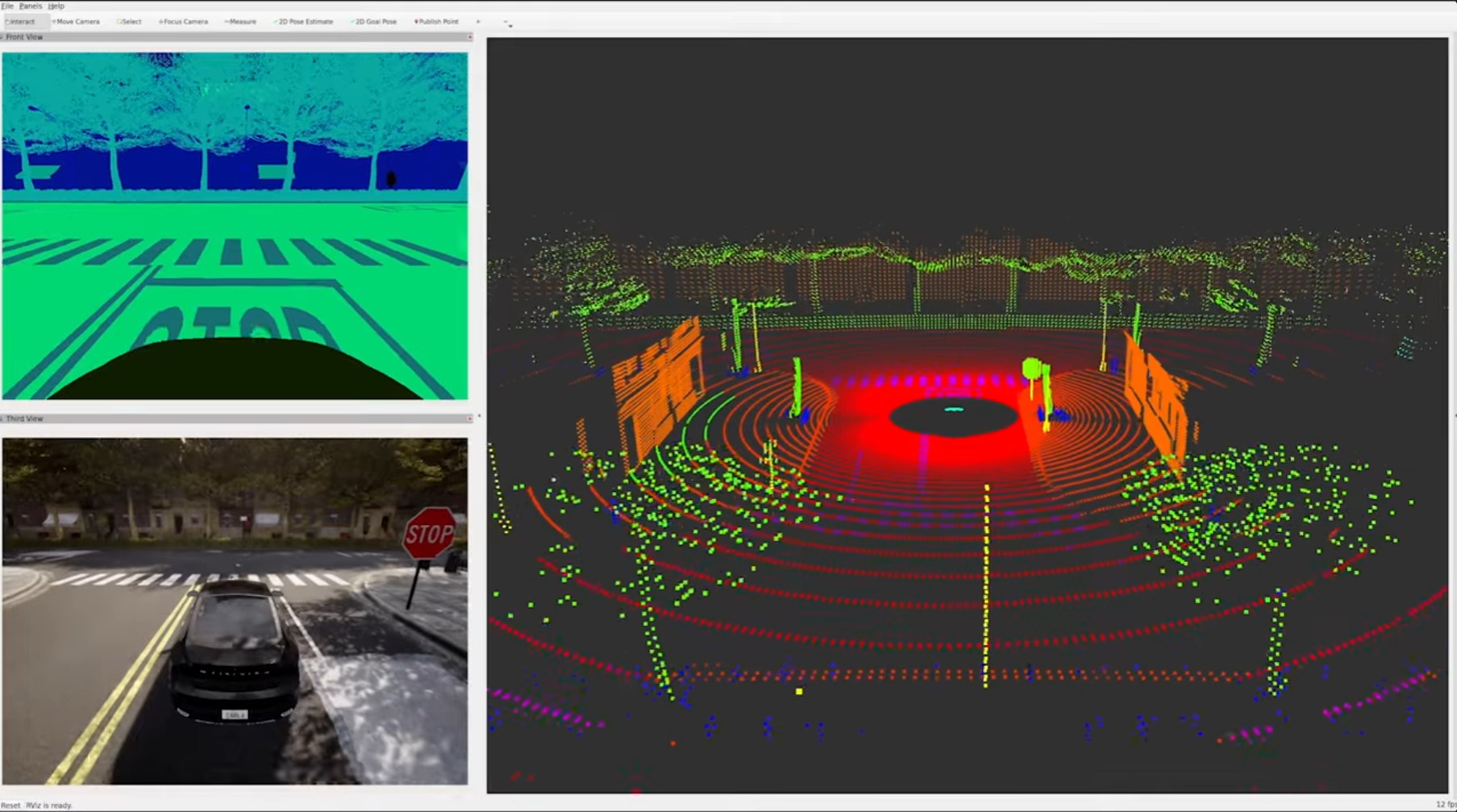} 
    \caption{\textbf{The CARLA Autonomous Driving Simulator.} Illustrating the pinnacle of the \textit{Autonomous Systems \& Navigation} domain, CARLA forces agents to process multi-modal, heterogeneous sensor streams (including RGB-D, LiDAR point clouds, and GPS). Operating under severe partial observability (POMDP) and stochastic weather conditions, agents must execute high-frequency continuous control while adhering to strict safety and traffic constraints.Source: \href{https://carla.org/}{carla.org}}
    \label{fig:carla_simulator}
\end{figure}

The foundational challenge for embodied agents is safely traversing dynamic, unstructured spaces. Distinct from stationary manipulation, this domain emphasizes ego-centric perception, obstacle avoidance, and multi-sensor fusion. Simulators such as CARLA \cite{dosovitskiy2017carla} (See Figure \ref{fig:carla_simulator}) and MetaDrive \cite{li2022metadrive} provide photorealistic urban environments where agents learn autonomous driving policies by fusing LiDAR, RGB cameras, and GPS data. Similarly, in indoor settings, benchmarks like Habitat \cite{savva2019habitat} evaluate visual navigation and spatial mapping. Recently, environments like V-IRL \cite{yang2024virl} have bridged this gap with language, requiring agents to navigate real-world street views based on natural language instructions.

\paragraph{2. Robotics \& Continuous Control}
\begin{figure}[htbp] 
    \centering
    \includegraphics[width=\columnwidth]{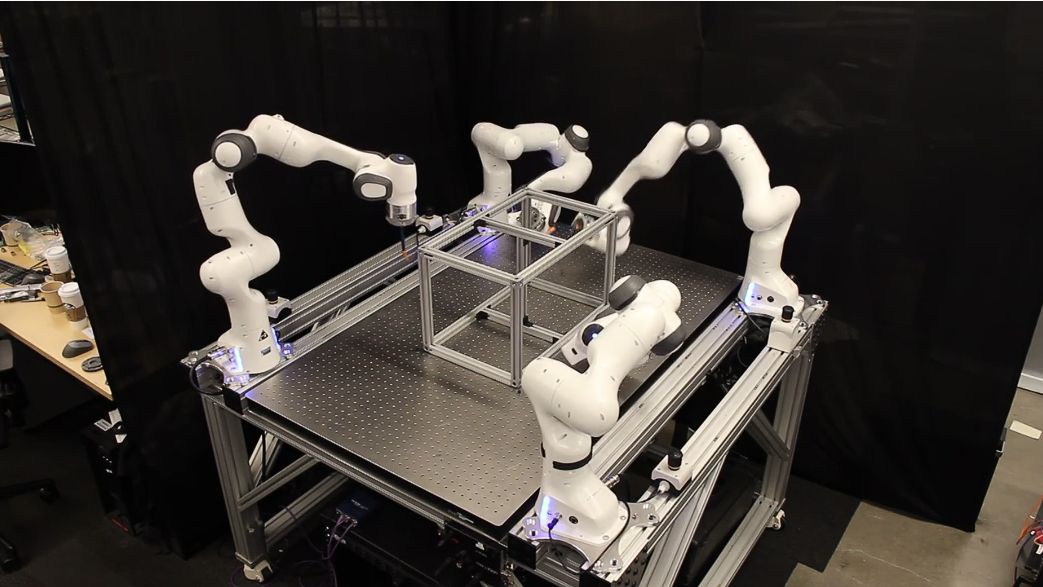} 
    \caption{\textbf{RoboBallet: Planning for Multirobot Reaching with GNN and RL.} A breakthrough in \textit{Multirobot Coordination}, RoboBallet enables up to eight arms to perform 40 tasks in a shared workspace. Its \textbf{graph neural network (GNN)} policy, trained via RL, jointly solves task allocation, scheduling, and collision-free motion planning. By representing robots, tasks, and obstacles as a graph, the GNN scales effectively, outputting coordinated joint velocities every 100ms. Trained in simulation, it generalizes zero-shot to new layouts and unlocks capabilities like layout optimization, improving execution times by up to 33\%. Source: \href{https://www.science.org/doi/10.1126/scirobotics.ads1204}{Science Robotics}}
    \label{fig:RoboBallet}
\end{figure}

Moving from macroscopic navigation to microscopic actuation, this domain focuses on controlling physical entities via high-dimensional continuous action spaces and complex contact dynamics.
\begin{itemize}[leftmargin=*, itemsep=2pt, topsep=4pt]
    \item \textbf{Locomotion:} Early benchmarks built on the MuJoCo \cite{todorov2012mujoco} physics engine evaluated an agent's ability to coordinate joints for walking or running. Modern equivalents like Brax \cite{freeman2021brax} leverage hardware acceleration to simulate thousands of parallel environments for rapid policy convergence.
   \item \textbf{Manipulation:} More intricate tasks involve dexterous manipulation and object interaction. Environments like Meta-World \cite{yu2020meta_world} and Robosuite \cite{zhu2020robosuite} focus on robotic arms performing multi-task pick-and-place, assembly, and tool use. On the other side, the visual-based ManiSkill 2 \cite{gu2023maniskill2} focuses on robotic arms performing multi-task pick-and-place, assembly, and tool use, often serving as proving grounds for Sim-to-Real transfer algorithms. RoboBallet represents a milestone in multi-agent continuous control by integrating Graph Neural Networks (GNNs) with reinforcement learning to achieve real-time, collision-free joint motion planning and task allocation in highly dense workspaces\cite{lai2025roboballet}. By abstracting physical entities into a permutation-invariant topological graph, the system overcomes traditional dimensionality bottlenecks and achieves profound zero-shot generalization, seamlessly adapting to unseen environments with varying numbers of robotic arms. (Figure \ref{fig:RoboBallet})
\end{itemize}

\paragraph{3. Games \& Competitive Simulation}
\begin{figure}[h!] 
    \centering
    \includegraphics[width=\columnwidth]{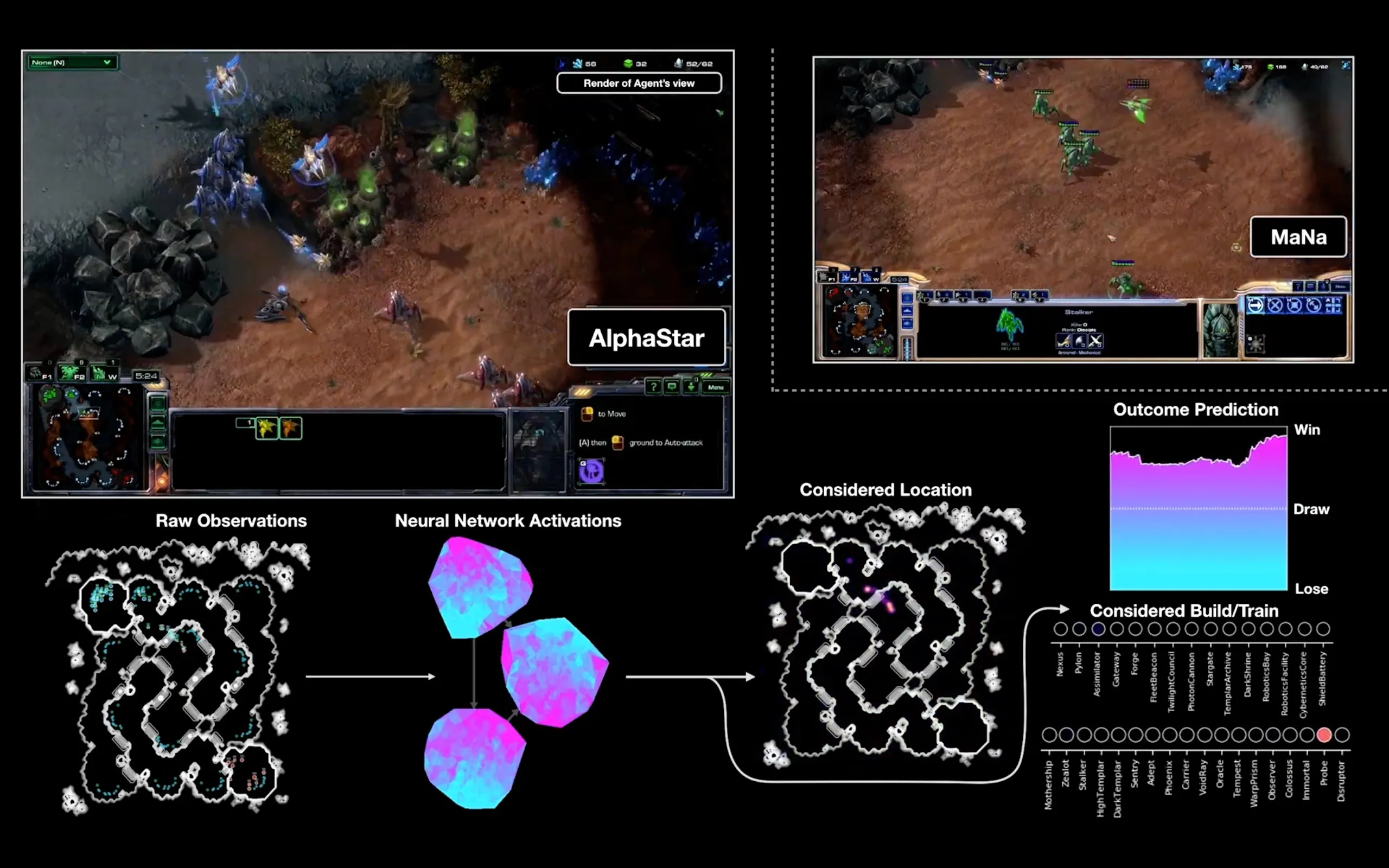} 
    \caption{\textbf{AlphaStar Mastering StarCraft II.} A landmark achievement in \textit{Real-Time Strategy}, AlphaStar masters the immense complexity of StarCraft II. By integrating deep neural networks with a multi-agent reinforcement learning league, it overcomes imperfect information and a massive combinatorial action space ($\approx 10^{26}$) to execute sophisticated macro-strategies and micro-tactics. Source: \href{https://deepmind.google/blog/alphastar-mastering-the-real-time-strategy-game-starcraft-ii/}{DeepMind Blog}}
    \label{fig:AlphaStar-StarCraft-II}
\end{figure}

While physical control dominates robotics, games have historically served as the ``fruit fly'' of AI research, providing pure, algorithmic testbeds free from hardware noise. This domain is characterized by well-defined rules, adversarial dynamics, and clear reward signals. From mastering pixel-based Arcade games (Atari \cite{bellemare2013arcade}) to achieving superhuman performance in perfect-information board games (AlphaZero \cite{silver2018general}), RL has consistently pushed the boundaries of combinatorial search. The frontier has since advanced to complex strategy and cooperative micromanagement (e.g., StarCraft II \cite{vinyals2019grandmaster}(Figure \ref{fig:AlphaStar-StarCraft-II}), Overcooked-AI \cite{carroll2019utility}), culminating in benchmarks like CICERO \cite{bakhtin2022human}, where agents must master diplomacy and trust-building in mixed-motive settings.

\paragraph{4. Language, Dialogue \& Digital Agents}
\begin{figure}[h!] 
    \centering
    \includegraphics[width=\columnwidth]{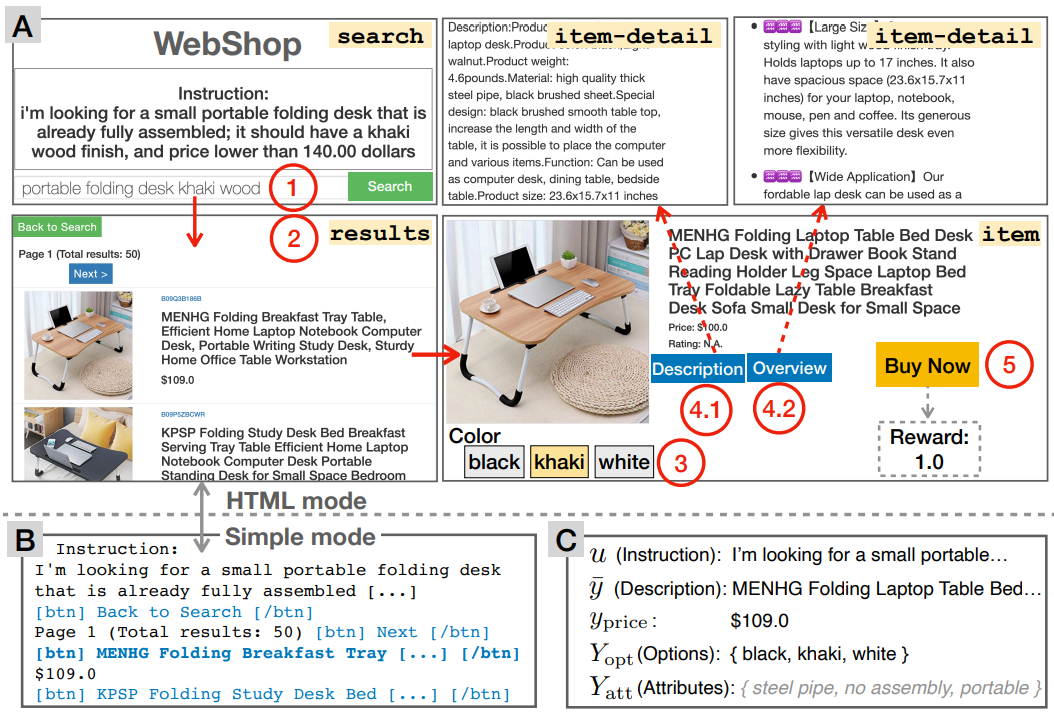} 
    \caption{\textbf{WebShop: Autonomous Agents for Online Shopping.} Serving as a realistic testbed for \textit{System-Level Web Interaction}, WebShop requires agents to map natural language instructions into actionable sequences (searching, filtering, and attribute selection). It provides dual observation modes (raw HTML and simplified text) to evaluate both structural parsing and high-level semantic reasoning in long-horizon tasks. Source: \href{https://webshop-pnlp.github.io/}{WebShop Project}}
    \label{fig:WebShop-Environment}
\end{figure}
As RL conquered physical simulation and adversarial games, its application profoundly shifted toward the semantic domain of Natural Language Processing (NLP). This represents a transition from spatial control to cognitive and digital interaction.
\begin{itemize}[leftmargin=*, itemsep=2pt, topsep=4pt]
    \item \textbf{Cognitive \& Interactive Text:} Moving beyond early text games, modern environments demand deep logical deduction. Benchmarks like ARC-AGI \cite{chollet2019measure} test inductive abstraction, while GSM8K \cite{cobbe2021training} and MATH \cite{hendrycks2021measuring} require rigorous arithmetic reasoning. Furthermore, environments like ProcessBench \cite{zheng2025processbenchidentifyingprocesserrors} evaluate an agent's ability to verify step-level logical proofs.
    \item \textbf{Web \& GUI Navigation:} LLM agents now operate as digital assistants. Environments like WebArena \cite{zhou2023webarena},  and OSWorld \cite{xie2024osworld} require agents to navigate complex HTML DOM trees and desktop interfaces, executing real-world tasks via keyboard and mouse commands. In addition, the activity environment of agents is also expanding into e-commerce, where RL begins to interact with Webs by penetrating WebShop (eBay \& Amazon) \cite{yao2022webshop}. (Figure \ref{fig:WebShop-Environment})
    \item \textbf{RLHF \& Alignment:} Perhaps the most impactful application today, Reinforcement Learning from Human Feedback (e.g., HH-RLHF \cite{bai2022training}, InstructGPT \cite{ouyang2022training}) is universally employed to align the outputs of massive dialogue models with human intent and harmlessness.
\end{itemize}

\paragraph{5. Optimization, Systems \& Operations}
Transitioning from digital interaction to backend infrastructure, RL has emerged as a powerful heuristic for solving NP-hard combinatorial problems and optimizing computer systems—tasks historically intractable for traditional solvers.
\begin{itemize}[leftmargin=*, itemsep=2pt, topsep=4pt]
    \item \textbf{Operations Research:} Graph-based RL is now heavily utilized for classical routing problems like the Traveling Salesperson Problem (TSP) \cite{kool2019attention}, as well as complex warehouse logistics (e.g., RWARE \cite{papoudakis2020benchmarking}).
    \item \textbf{Computer Systems \& Software Engineering:} RL agents are increasingly deployed to optimize the very systems that run them. Environments like CompilerGym \cite{cummins2022compilergym} train agents to optimize LLVM compiler passes. More impressively, benchmarks like SWE-bench \cite{jimenez2023swebench} and ColBench \cite{zhou2025sweetrltrainingmultiturnllm} require agents to autonomously resolve real-world software bugs, while VerilogEval \cite{liu2023verilogeval} tests hardware RTL generation.
\end{itemize}

\paragraph{6. Scientific \& Real-world Applications}

\begin{figure}[h!] 
    \centering
    \includegraphics[width=\columnwidth]{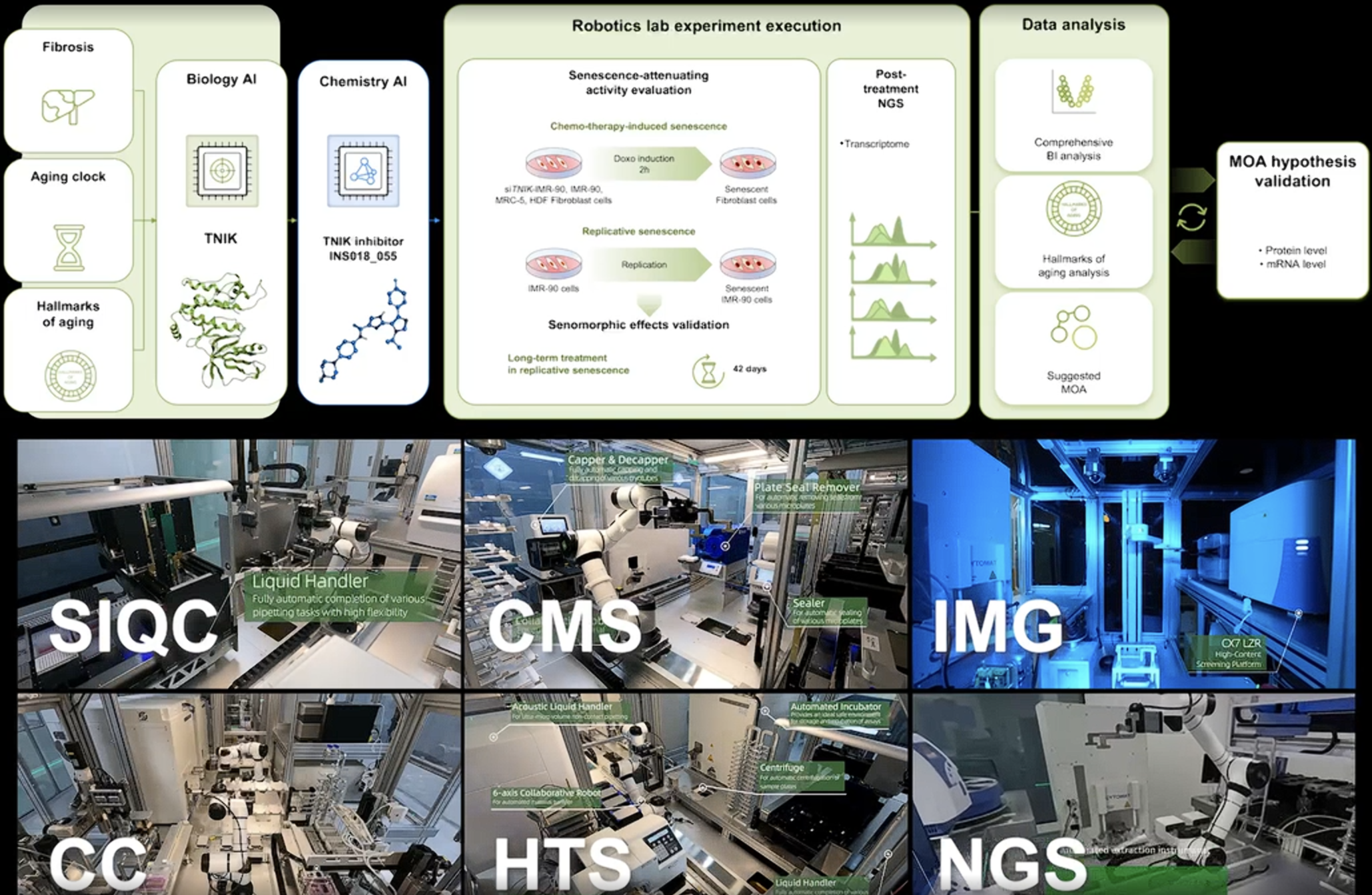} 
    \caption{\textbf{Insilico Medicine's Fully Automated Robotics Laboratory.} Representing the frontier of \textit{AI-driven drug discovery}, this system automates complex wet-lab processes. It integrates reinforcement learning to autonomously optimize experimental strategies and process control using massive biological datasets, ensuring strict reproducibility. Source: \href{https://www.eurekalert.org/news-releases/1074367?language=chinese}{EurekAlert}}
    \label{fig:Insilico-Medicine}
\end{figure}

The ultimate frontier of RL applications lies in high-stakes scientific discovery and domain-specific operations, characterized by immense complexity and the infeasibility of trial-and-error in the real world. In the formal sciences, environments like Lean 4 \cite{moura2021lean} and MiniF2F \cite{zheng2021minif2f} utilize RL for automated theorem proving. In the physical and biological sciences, RL drives innovations in controlling nuclear fusion plasma (Tokamak control \cite{Degrave2022Magnetic}), simulating molecular dynamics \cite{zhou2019optimization}, and generating biological pathways (Geneformer \cite{theodoris2023transfer}). Finally, in healthcare, frameworks like Med-PaLM \cite{singhal2023large} integrate clinical reasoning, demonstrating RL's maturation into a transformative tool for the most critical human endeavors. At the same time, with the widespread use of reinforcement learning in programming, more and more coding environments are emerging and beginning to impact academia and industry (e.g., LiveCodeBench\cite{jain2024livecodebenchholisticcontaminationfree}, HumanEval\cite{chen2021evaluating}). Furthermore, its impact on coding extends beyond software engineering; it has also fostered a certain ecosystem of reinforcement learning environments within the field of scientific programming (SciCode).\cite{ tian2024scicoderesearchcodingbenchmark}.

\begin{itemize}[leftmargin=*, itemsep=2pt, topsep=4pt]
    \item \textbf{Healthcare and biology:} Reinforcement learning has begun to penetrate complex medical question answering and diagnosis. Some typical examples include: Medical Reasoning (MedQA, JAMA Clinical, etc.). \cite{komorowski2018artificial}. With the emergence of chain-of-thoughts (CoTs), reinforcement learning has been applied to complex, expert-level medical and biological reasoning (MedQA-USMLE, MedXpertQA, KEGG PATHWAY, EHR-based Clinical Reasoning).\cite{kanehisa2000kegg,jin2021disease,medxpertqa2025,ehrmind2025}
    \item \textbf{Physical Sciences:} Notable examples include controlling plasma in nuclear fusion tokamak reactors \cite{Degrave2022Magnetic} and optimizing molecular configurations in protein folding and drug discovery.
    \item \textbf{Mathematical \& Formal Reasoning:} RL has emerged as a crucial mechanism for enhancing the deductive, multi-step reasoning capabilities of intelligent agents. Environments built upon datasets like MATH \cite{hendrycks2021measuring} and GSM8K \cite{cobbe2021training} challenge models with complex symbolic logic and physical word problems. Recent breakthroughs apply advanced RL paradigms—such as process reward models (PRMs) and group relative policy optimization (GRPO)—to achieve expert-level performance in formal theorem proving and Olympiad-level mathematical reasoning \cite{lightman2023lets, shao2024deepseekmath}.
    \item \textbf{Finance \& Quantitative Trading:} Environments such as FinRL \cite{liu2020finrl} and Gym-Anytrading \cite{alaee2018gym} formulate quantitative trading as an MDP based on real-world financial time-series. These environments typically challenge agents with macroscopic portfolio management, balancing profit maximization against strict risk constraints under highly non-stationary market dynamics. Conversely, comprehensive platforms like Microsoft's Qlib \cite{yang2020qlib} provide high-fidelity micro-structural environments. In Qlib's \textit{Order Execution} environments, agents act at high frequencies to dynamically slice large institutional block trades, learning to minimize market impact costs and slippage in a complex, limit-order-book simulation.
\end{itemize}

\begin{table*}[htbp]
  \centering
  \small 
  \caption{Comprehensive Taxonomy of RL Environments by Primary Application Domain and Subdomain}
  \label{tab:app_domain_envs}
  \renewcommand{\arraystretch}{1.05} 
  \begin{tabularx}{\linewidth}{@{} l l l X @{}}
    \toprule
    \textbf{Environment} & \textbf{Subdomain / Specific Task Focus} & \textbf{Year$^{\mathrm{a}}$} & \textbf{DOI / Source} \\
    \midrule
    
    \multicolumn{4}{@{}l}{\textbf{1. Autonomous Systems \& Navigation}} \\
    \midrule
    CARLA Simulator\cite{dosovitskiy2017carla} & Autonomous Driving (Multi-sensor Planning) & 2017 & \href{https://doi.org/10.48550/arXiv.1711.03938}{CoRL'17} \\
    NoCrash \& CARLA100\cite{codevilla2019exploring} & Visual Navigation (3D Photorealistic) & 2019 & \href{https://github.com/felipecode/coiltraine/blob/master/docs/exploring_limitations.md}{ICCV'19} \\
    Safety Gym\cite{Ray2019} & Autonomous Systems (Safe RL \& Constrained MDPs) & 2019 & \href{https://github.com/openai/safety-gym}{GitHub: safety-gym} \\
    MetaDrive\cite{li2022metadrive} & Autonomous Driving (Generalization \& Safety) & 2021 & \href{https://doi.org/10.48550/arXiv.2109.12674}{arXiv:2109.12674} \\
    V-IRL\cite{yang2024virl} & Visual Navigation (Language-Guided Street View) & 2024 & \href{https://doi.org/10.48550/arXiv.2402.03310}{arXiv:2402.03310} \\
    \midrule

    \multicolumn{4}{@{}l}{\textbf{2. Robotics, Embodied \& Continuous Control}} \\
    \midrule
    MuJoCo (Gym Control)\cite{todorov2012mujoco} & Locomotion (Continuous Physics Control) & 2012 & \href{https://doi.org/10.1109/IROS.2012.6386109}{IROS'2012} \\
    Meta-World\cite{yu2020meta_world} & Manipulation (Multi-task Robotic Arms) & 2019 & \href{https://doi.org/10.48550/arXiv.1910.10897}{CoRL'19} \\
    SAPIEN\cite{Xiang_2020_SAPIEN} & Manipulation (Articulated Objects \& Precision Physics) & 2020 & \href{https://doi.org/10.48550/arXiv.2003.08515}{CVPR'20} \\
    Robosuite\cite{zhu2020robosuite} & Manipulation (Modular Robot Simulation) & 2020 & \href{https://doi.org/10.48550/arXiv.2009.12293}{arXiv:2009.12293} \\
    Brax\cite{freeman2021brax} & Locomotion (Hardware-Accelerated Physics) & 2021 & \href{https://doi.org/10.48550/arXiv.2106.13281}{NeurIPS'21} \\
    ManiSkill 2\cite{gu2023maniskill2} & Manipulation (Visual Pick-and-Place) & 2023 & \href{https://doi.org/10.48550/arXiv.2302.04659}{ICLR'23} \\
    \midrule

    \multicolumn{4}{@{}l}{\textbf{3. Games \& Competitive Simulation}} \\
    \midrule
    Arcade Learning Env\cite{bellemare2013arcade} & Board \& Arcade Games (Atari 2600) & 2013 & \href{https://doi.org/10.1613/jair.3912}{10.1613/jair.3912} \\
    AlphaZero\cite{silver2018general} & Board \& Arcade Games (Self-play Go/Chess) & 2017 & \href{https://doi.org/10.1038/nature24270}{10.1038/nature24270} \\
    SMAC (StarCraft II)\cite{vinyals2019grandmaster} & Complex Strategy (Cooperative Micromanagement) & 2019 & \href{https://doi.org/10.48550/arXiv.1902.04043}{arXiv:1902.04043} \\
    Overcooked-AI\cite{carroll2019utility} & Physics-based Puzzles (Human-AI Coordination) & 2019 & \href{https://doi.org/10.48550/arXiv.1910.05789}{arXiv:1910.05789} \\
    MineDojo (Minecraft)\cite{fan2022minedojobuildingopenendedembodied} & Physics-based Puzzles (Open-Ended Survival) & 2022 & \href{https://doi.org/10.48550/arXiv.2206.08853}{NeurIPS'22} \\
    CICERO (Diplomacy)\cite{bakhtin2022human} & Complex Strategy (Negotiation \& Trust) & 2022 & \href{https://doi.org/10.1126/science.ade9097}{10.1126/science.ade9097} \\
    \midrule

    \multicolumn{4}{@{}l}{\textbf{4. Language, Dialogue \& Digital Agents}} \\
    \midrule
    ARC-AGI\cite{chollet2019measure} & Cognitive Reasoning (Inductive Abstraction) & 2019 & \href{https://doi.org/10.48550/arXiv.1911.01547}{arXiv:1911.01547} \\
    ALFWorld\cite{shridhar2020alfworld} & Interactive Text (Text-aligned Embodied AI) & 2020 & \href{https://doi.org/10.48550/arXiv.2010.03768}{ICLR'21} \\
    GSM8K\cite{cobbe2021training} \& MATH\cite{hendrycks2021measuring} & Interactive Text (Math Word Problems) & 2021 & \href{https://doi.org/10.48550/arXiv.2110.14168}{arXiv:2110.14168} \\
    HH-RLHF (Anthropic)\cite{bai2022training} & RLHF \& Dialogue Alignment (Harmlessness) & 2022 & \href{https://doi.org/10.48550/arXiv.2204.05862}{arXiv:2204.05862} \\
    BIRD (Text-to-SQL)\cite{li2023bird} & Digital Agents (Database Semantic Parsing) & 2023 & \href{https://doi.org/10.48550/arXiv.2305.03111}{NeurIPS'23} \\
    WebArena\cite{zhou2023webarena} & Web \& GUI Navigation (Realistic Web Agent) & 2023 & \href{https://doi.org/10.48550/arXiv.2307.13854}{ICLR'24} \\
    WebShop\cite{yao2022webshop} & Finance \& Commerce (Simulated E-commerce) & 2022 & \href{https://doi.org/10.48550/arXiv.2207.01206}{NeurIPS'22} \\
    OSWorld\cite{xie2024osworld} & Web \& GUI Navigation (Desktop Automation) & 2024 & \href{https://doi.org/10.48550/arXiv.2404.07972}{arXiv:2404.07972} \\
    TextArena\cite{guertler2025textarena} & Interactive Text (Multi-Agent Negotiation) & 2024 & \href{https://doi.org/10.48550/arXiv.2504.11442}{	arXiv:2504.11442} \\
    ProcessBench\cite{zheng2025processbenchidentifyingprocesserrors} & Interactive Text (Step-level Verification) & 2024 & \href{https://doi.org/10.48550/arXiv.2412.06559}{arXiv:2412.06559} \\
    \midrule

    \multicolumn{4}{@{}l}{\textbf{5. Optimization, Systems \& Operations}} \\
    \midrule
    Graph-based TSP\cite{kool2019attention} & Operations Research (Combinatorial Optimization) & 2020 & \href{https://doi.org/10.48550/arXiv.1803.08475}{arXiv:1803.08475} \\
    Robot Warehouse (RWARE)\cite{papoudakis2020benchmarking} & Operations Research (Warehouse Logistics) & 2020 & \href{https://doi.org/10.48550/arXiv.2006.07869}{arXiv:2006.07869} \\
    CompilerGym\cite{cummins2022compilergym} & Computer Systems (Compiler Optimization) & 2021 & \href{https://doi.org/10.48550/arXiv.2109.08267}{arXiv:2109.08267} \\
    SWE-bench\cite{jimenez2023swebench} & Computer Systems \& Software (GitHub Issues) & 2023 & \href{https://doi.org/10.48550/arXiv.2310.06770}{ICLR'24} \\
    VerilogEval\cite{liu2023verilogeval} & Hardware \& Circuit Design (RTL Generation) & 2023 & \href{https://doi.org/10.48550/arXiv.2309.07544}{arXiv:2309.07544} \\
    ColBench (SweetRL)\cite{zhou2025sweetrltrainingmultiturnllm} & Computer Systems (Collaborative Software Eng.) & 2025 & \href{https://doi.org/10.48550/arXiv.2503.15478}{arXiv.2503:15478} \\
    NPPC Gym\cite{yang2025nppc} & Nondeterministic Polynomial-time Problem & 2025 & \href{https://doi.org/10.48550/arXiv.2504.11239}{arXiv:2504.11239} \\
    MLE-bench\cite{chan2024mle} & Computer Systems \& AutoML (Kaggle Tasks) & 2024 & \href{https://doi.org/10.48550/arXiv.2410.07095}{arXiv:2410.07095} \\
    \midrule

    \multicolumn{4}{@{}l}{\textbf{6. Scientific \& Real-world Applications}} \\
    \midrule
    Lean 4\cite{moura2021lean}/ MiniF2F\cite{zheng2021minif2f} & Formal Sciences (Automated Theorem Proving) & 2021 & \href{https://doi.org/10.48550/arXiv.2109.00110}{ICLR'22} \\
    FinRL\cite{liu2020finrl} & Finance \& Quantitative Trading (Market Simulation) & 2020 & \href{https://doi.org/10.48550/arXiv.2011.09607}{arXiv:2011.09607} \\ FinQA\cite{chen2022finqadatasetnumericalreasoning} & Finance \& Quantitative trading (Financial Reasoning) & 2021 & \href{https://doi.org/10.48550/arXiv.2109.00122}{EMNLP'21} \\
    MolDQN\cite{zhou2019optimization} & Physical Sciences (Rare-event Logic Simulation) & 2019 & \href{https://doi.org/10.1038/s41598-019-47148-x}{Nature Sci Rep'19} \\
    Tokamak (DeepMind)\cite{Degrave2022Magnetic} & Physical Sciences (Plasma Magnetic Control) & 2022 & \href{https://doi.org/10.1038/s41586-021-04301-9}{Nature} \\
    Med-PaLM (Clinical RL)\cite{singhal2023expertlevelmedicalquestionanswering} & Healthcare (Medical QA \& Diagnosis) & 2023 & \href{https://doi.org/10.48550/arXiv.2305.09617}{arXiv:2305.0961} \\ Geneformer\cite{theodoris2023transfer} & Healthcare (Biological Pathway Generation) & 2023 & \href{https://doi.org/10.1038/s41586-023-06139-9}{Nature'23} \\
    \bottomrule
  \end{tabularx}
  \vspace{0ex}
  \footnotesize{\textbf{Note:} Environments are strictly categorized by their primary application domain and precise subdomains. This hierarchy demonstrates RL's expansion from isolated game simulators (e.g., Atari, MuJoCo) into highly specialized, real-world scientific, medical, and software engineering domains.}
\end{table*}

\section{Evolutionary Trajectory and Paradigm Shifts}
Tracing the evolutionary trajectory of Reinforcement Learning environments reveals a profound paradigm shift: a steady transition from mastering low-level physical control to navigating high-level cognitive, semantic, and socially interactive arenas. At the epicenter of this contemporary shift lies the advent of Large Language Models (LLMs), which have fundamentally redefined the state representations, action spaces, and reward mechanisms of RL environments. To comprehensively map this ongoing evolution, this section decomposes the trajectory into two critical dimensions. We first analyze the current landscape through the lens of LLMs, examining how language-centric and interactive environments have become the primary crucible for modern agent alignment and reasoning. Subsequently, we turn our attention to the frontiers beyond LLMs to investigate the broad ecosystem RL environments.

\subsection{From the Perspective of LLMs}
Reinforcement learning environments have not advanced in isolation; their progress is inextricably linked to the underlying algorithmic bottlenecks and breakthroughs. Over the past decade, RL environments have evolved from minimalist mathematical abstractions into complex, high-dimensional, multi-modal, and increasingly realistic digital worlds. Each major shift in environmental design has not merely set a higher benchmark but has actively redefined the field's research questions, algorithmic priorities, and evaluation standards. To perfectly capture this trajectory, we synthesize all four taxonomic dimensions—\textit{Application Domain, Multi-Modal Span}, \textit{Observability}, and \textit{Agent Capabilities}—and trace their concurrent macro- and micro-evolution across four distinct algorithmic eras.

\begin{figure*}[h!]
    \centering
    \includegraphics[width=1\textwidth]{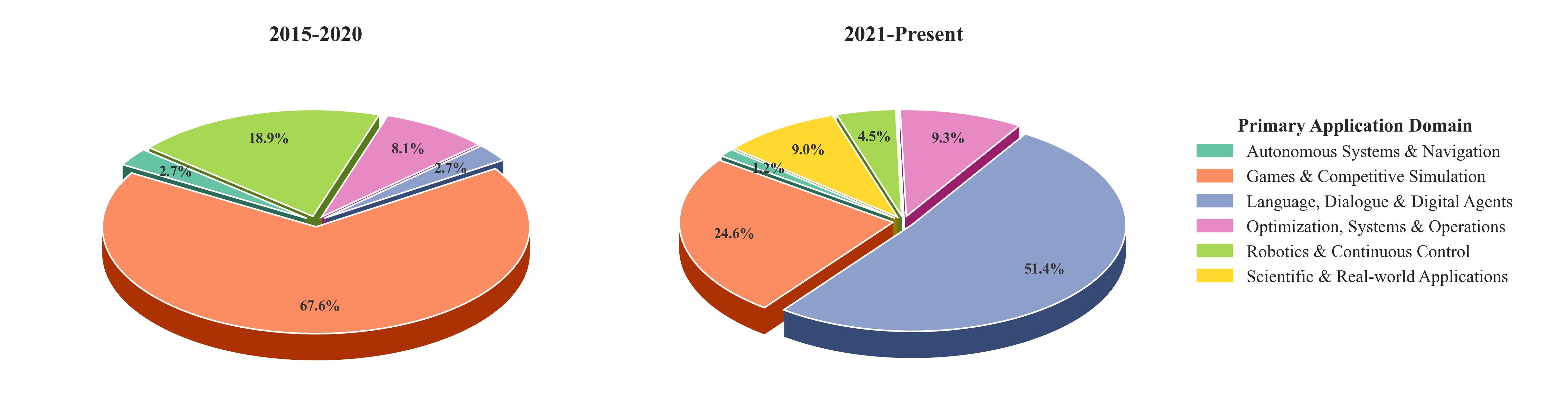}
    \caption{A retrospective analysis of the temporal distribution of RL environments by primary application domain. The evolution from the 2015–2020 era to the post-2020 period illustrates the impact of LLM engagements, characterized by a significant expansion in Language, Dialogue, and Digital Agent domains at the expense of traditional simulated robotics and gaming environments. The dividing line: 2020, the release of GPT-3, the first large language model.}
    \label{fig:app_domain}
\end{figure*} 

\begin{figure*}[h!]
    \centering
    \includegraphics[width=0.9\textwidth]{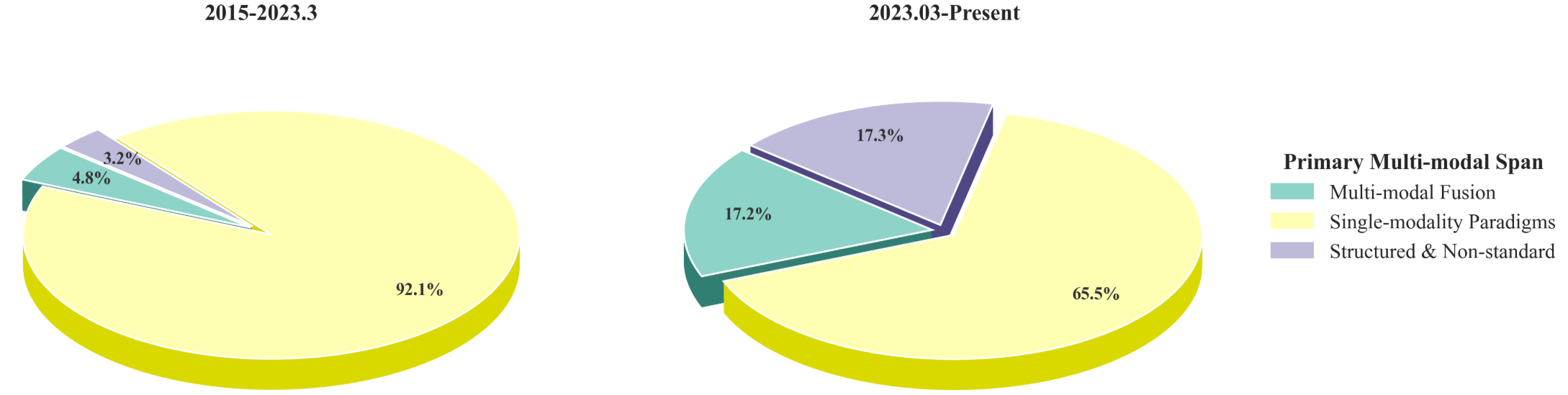}
    \caption{The paradigm shift in modality distribution. The figure illustrates the breakdown of single-modality dominance and the resurgence of structured and non-standard inputs
    in recent eras. The dividing line: March 2023, the release of GPT-4, the first multimodal large language model.}
    \label{fig:modalities}
\end{figure*}

\begin{figure*}[h!]
    \centering
    \includegraphics[width=0.9\textwidth]{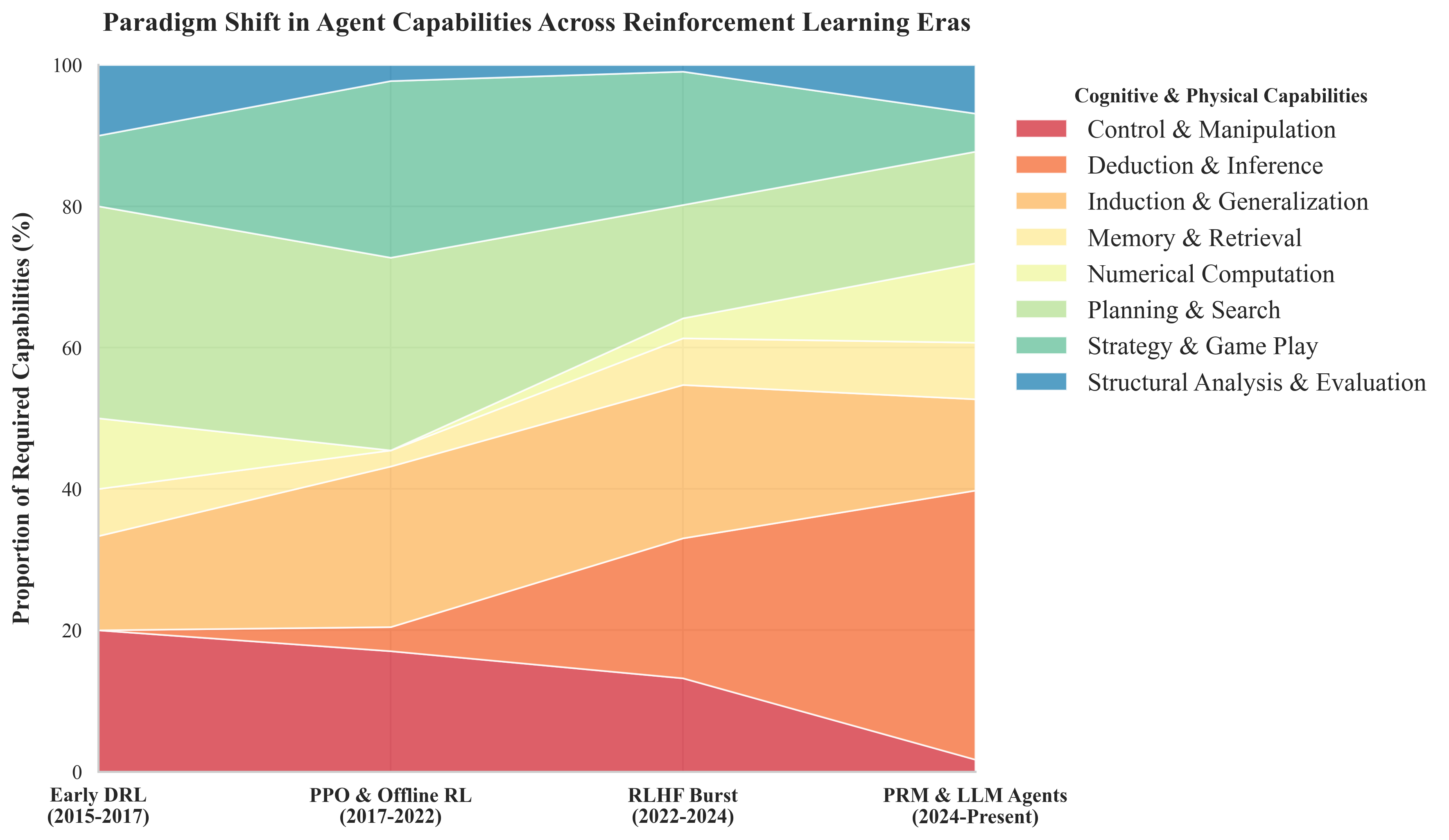}
    \caption{\textbf{Temporal Evolution of Capability Requirements in LLM Environments.} 
    This alluvial plot tracks the shifting cognitive demands of environments utilizing LLMs as active agents, based on their inception dates. The data illustrates a rapid escalation from foundational language understanding in early sandboxes to higher-order faculties---such as deduction, long-horizon planning, and tool utilization---in the post-2023 era. This highlights the transition of LLM testbeds from passive linguistic evaluators to complex reasoning benchmarks.}
    \label{fig:evo_cap}
\end{figure*}

\paragraph{Early Deep RL \& Pixel-Level Reactive Control (2015--2017)}
Initiated by the watershed moment of Deep Q-Networks (DQN) \cite{mnih2015human}, the first era was defined by the ambition of "Tabula rasa" (blank slate) learning directly from raw, high-dimensional sensory inputs. \textbf{Application-wise}, this era was overwhelmingly dominated by classic \textit{Games \& Competitive Simulation}. The \textbf{Multi-modal Span} (Figure \ref{fig:modalities}) was strictly confined to \textit{Single-modality Paradigms}, relying almost exclusively on 2D RGB pixel arrays (e.g., Arcade Learning Environment for Atari 2600 \cite{bellemare2013arcade}) or low-dimensional numerical vectors (e.g., classic OpenAI Gym \cite{brockman2016openai}). Operating entirely as \textbf{Single-Agent MDPs}, these early environments presented a relatively simplistic \textbf{Observability} profile, characterized either by \textit{Perfect Information} (fully visible game boards) or rudimentary spatial occlusion (e.g., first-person raycasting in ViZDoom \cite{kempka2016vizdoom}). Consequently, as reflected in Figure \ref{fig:evo_cap}, the required \textbf{Capabilities} strictly demanded reactive \textit{Planning, Strategy \& Game Play} and spatial feature extraction. Agents functioned as "System 1" reactive machines, mapping visual stimuli to discrete joystick actions, with evaluation rigidly tied to maximizing accumulated numerical game scores. From a hindsight perspective, the manipulation, control and planning capabilities of modern large language models have already permeated early reinforcement learning environments and demonstrated stronger performances than traditional solutions. 

\paragraph{Mature Continuous Control \& The Dawn of LLMs (2017--2022)}
With the mathematical stabilization of policy gradient methods (e.g., PPO \cite{schulman2017proximal}, SAC \cite{haarnoja2018soft}), the algorithmic appetite expanded from discrete arcade games to encompass both continuous physics and complex strategic reasoning. \textbf{Application Domains} experienced a profound bifurcation. On one front, \textit{Robotics} and \textit{Autonomous Systems} demanded high-frequency \textit{Control \& Manipulation}, challenging agents with complex contact dynamics and high degrees of freedom (DoF) in rigid simulators like MuJoCo \cite{todorov2012mujoco}, Meta-World \cite{yu2020meta_world}, and Brax \cite{freeman2021brax}. The high demand of \textit{Control \& Manipulation} of the primary environment for modern large language model interaction is still of that period. 

On the other front, required \textbf{Capabilities} saw a massive surge in \textit{Strategy \& Gameplay} and \textit{Induction \& Generalization}. This was driven by the rise of \textit{Multi-Agent Stochastic Games (SG)} to tackle decentralized execution. With the huge success of AlphaGo\footnote{https://deepmind.google/research/alphago/}, the research community's focus has gradually shifted to deeper strategy games. Unlike perfect information games, \textbf{Observability} became a critical bottleneck as environments like StarCraft II (via SMAC \cite{vinyals2019grandmaster}) and Google Research Football \cite{kurach2020google} introduced severe \textit{Imperfect Information} via spatial ``Fog-of-War,'' forcing agents to master cooperative micromanagement. Despite these advances in dynamics and population scale, the underlying modalities remained largely confined to homogeneous sandboxes. 

This bottleneck was ultimately shattered at the twilight of the era by the release of GPT-3\protect\footnote{\url{https://openai.com/}}. As the first truly generalist large language model, it catalyzed a paradigm shift, thoroughly redefining the agentic landscape and setting the stage for semantic, text-driven environments. From the perspective of early large language models, the emergence of GPT-3 significantly increased the demand for simple induction and generalization capabilities in large language models. This led to a new paradigm in the AI research community that the community began to enhance and examine the induction abilities of LLMs.

\paragraph{Foundation Models, VLA \& Semantic Alignment (2022--2023)}
The introduction of InstructGPT \cite{ouyang2022training} and the widespread industrialization of Reinforcement Learning from Human Feedback (RLHF) triggered an irreversible paradigm shift. \textbf{Application Domains} were aggressively pulled from physical simulators into the semantic realm of \textit{Language, Dialogue \& Digital Agents}. This era marked a radical explosion in the \textbf{Multi-Modal Span}: RL environments transitioned from calculating physical torques to predicting natural language tokens (e.g., ALFWorld \cite{shridhar2020alfworld}, WebShop \cite{yao2022webshop}). With the emergence of chain-of-thought \cite{wei2022chain}, LLMs began to take over more reasoning tasks.  

Furthermore, the boundary between text and vision blurred with the rise of \textit{Vision-Language-Action (VLA)} models interacting with GUI interfaces. 
Crucially, the nature of \textbf{Observability} morphed from physical occlusion to \textit{Digital Compartmentalization}. Single-agent LLMs had to navigate vast, partially observable digital spaces—such as scrolling through complex HTML DOM trees or computer desktops (e.g., WebArena \cite{zhou2023webarena}, OSWorld \cite{xie2024osworld}, Mind2Web \cite{deng2023mind2web}). The required \textbf{Capabilities} evolved from spatial control to semantic alignment, long-context \textit{Memory \& Retrieval}, and API tool-use, bridging the gap between passive language modeling and active digital execution. At the same time, with the release of GPT-o1, the emergence of these early mathematical reasoning models began to drive a surge in demand for reasoning and mathematical induction abilities\cite{openai2024o1}. Simple mathematical reasoning environments, such as GSM8K, are beginning to challenge agents' capabilities \cite {cobbe2021training}.

\paragraph{System 2 Reasoning \& Real-world engagements (2024--Present)}
The current era is characterized by the awakening of rigorous, "System 2" logical reasoning and complex multi-agent collaboration. Driven by the need to verify intermediate logic—via Process Reward Models (PRMs) \cite{lightman2023lets} and GRPO \cite{shao2024deepseekmath}—the \textbf{Multi-Modal Span} exhibits a profound return to \textit{Structured \& Logic Representations}. Environments no longer tolerate heuristic approximations; they demand absolute deductive precision in formal theorem proving (e.g., Lean 4 \cite{moura2021lean}) and mathematical planning (e.g., MATH \cite{hendrycks2021measuring}, Countdown Game \cite{deepseek2025r1}). As mathematical reasoning deepens, large-scale model reasoning is beginning to permeate environments with higher mathematical complexity, such as the Olympiad bench \cite{lightman2023lets, shao2024deepseekmath}. Simultaneously, these capabilities are also starting to spill over into physics problem reasoning\cite{xiang2025seephysdoesseeinghelp}.

Concurrently, \textbf{Observability} encompasses massive \textit{Systemic Opacity}, requiring agents to actively debug and explore dark, million-line codebases (e.g., SWE-bench \cite{jimenez2023swebench}, InterCode \cite{yang2023intercode}). The \textbf{Agent Population} has also ascended to a sociological level: LLMs now engage in sophisticated \textit{Multi-Agent Text Negotiations} (e.g., TextArena \cite{guertler2025textarena})., managing deception, trust-building, and Theory of Mind in purely semantic arenas. As dramatically depicted in Figure \ref{fig:evo_cap}, high-order cognition completely overtakes physical actuation: demands for \textit{Deduction \& Inference} and \textit{Structural Analysis} experience an unprecedented, exponential expansion. This illustrates that, in coding tasks, the demand for agents' \textit{Deduction \& Inference} and \textit{Structural Evaluation} capabilities within the environment is increasing significantly.

\paragraph{Paradigm Shifts}
Two striking phenomena emerge from this multi-dimensional historical synthesis. First, the agent capabilities requirements (Figure \ref{fig:evo_cap}) exhibit a ``U-shaped'' revival of \textit{Structured \& Non-standard} representations. During early ages, early deep learning relentlessly pursued end-to-end raw pixel processing \cite{mnih2015human}, explicitly avoiding hand-crafted features. However, modern LLM agents (Eras 3 \& 4) paradoxically require explicit structural abstractions—such as parsed HTML DOM trees, strictly formatted JSON APIs, and abstract syntax trees (e.g., BIRD \cite{li2023bird}, SWE-bench \cite{jimenez2023swebench})—to function reliably and avoid hallucination in complex digital ecosystems. This led to a revival of agents' structural analysis ability.

Despite the immense commercial popularity and massive capital investment in Autonomous Driving, its footprint in open-source RL environments (e.g., CARLA \cite{dosovitskiy2017carla}, MetaDrive \cite{li2022metadrive}) remains paradoxically negligible. This highlights a profound systemic disconnect: the exorbitant computational cost of high-fidelity multi-sensor physics simulation, coupled with the unforgiving, safety-critical nature of real-world driving, has pushed the autonomous vehicle industry heavily toward offline Imitation Learning and predictive world-modeling. Consequently, modern RL benchmarking has largely abandoned embodied vehicle control, pivoting instead to flourish in the highly scalable, low-cost, and easily parallelizable domains of digital cognitive reasoning and LLM alignment.

Overall, the emergence of multimodal large language models (MLLMs) has had a profound impact on the reinforcement learning environment paradigm, changing the ecosystem of the dominance of single-modality environments. Another easily observable trend is that large language model agents are increasingly being used in scientific research and real-world applications (Figure \ref{fig:app_domain}). 

\subsection{Beyond LLMs: A Different Ecosystem}

To understand the intrinsic evolution of Reinforcement Learning (RL) independent of the recent Large Language Model (LLM) surge, we analyze the shifting landscape of RL environments through three dimensions: application domains, required agent capabilities, and domain-specific cognitive fingerprints. Through this multi-scale analysis, we aim to answer a critical question: Has the evolution of RL environments under the shadow of the LLM discourse formed a distinct, thriving ecosystem of its own?

\paragraph{The Macro-level Shift: From Laboratory Control to Industrial Optimization} 
The fundamental purpose of RL has undergone a profound paradigm shift over the last decade, transitioning decisively from "laboratory-scale control" to "industrial-scale optimization." 

During the \textbf{Era of Physical Control (2015--2019)}, RL research was heavily anchored in \textit{Robotics \& Continuous Control} ($30.4\%$) and \textit{Games \& Competitive Simulation} ($12.6\%$). The primary goal was to achieve end-to-end mapping from high-dimensional sensory inputs to motor actions, treating games as the ultimate testbeds for algorithmic supremacy. 

However, in the \textbf{Era of System Intelligence (2020--Present)}, the landscape shifted dramatically. The historically prominent testbeds saw a stark reduction in focus: \textit{Games} dropped to just $3.0\%$ (largely viewed as solved benchmarks post-AlphaGo and AlphaStar), and the proportion of traditional \textit{Robotics} nearly halved to $15.9\%$. This relative decline in physical robotics underscores the persistent friction of the "sim-to-real" gap, where physical sample inefficiency and safety constraints bottlenecked rapid deployment. 

Conversely, there has been a dominant surge in \textit{Optimization, Systems \& Operations}, which now accounts for nearly half of the primary application landscape ($48.6\%$). Researchers increasingly pivoted toward data-rich, digital-native environments where simulators are highly faithful to reality. By formulating complex logistical challenges---such as supply chain routing, smart grid management, and telecommunications---as Markov Decision Processes, researchers leverage single and multi-agent RL (MARL) to achieve scalable solutions where marginal algorithmic improvements translate into massive operational efficiencies. 

Interestingly, amidst these drastic shifts, \textit{Autonomous Systems} ($12.8\%$) and \textit{Scientific \& Real-world Applications} ($15.8\%$) maintained remarkable statistical resilience. This stability suggests that domains requiring a hybrid integration of both high-level semantic planning and low-level physical control represent a mature, sustainable pathway for RL deployment.

\begin{figure*}[h!]
    \centering    \includegraphics[width=1\textwidth]{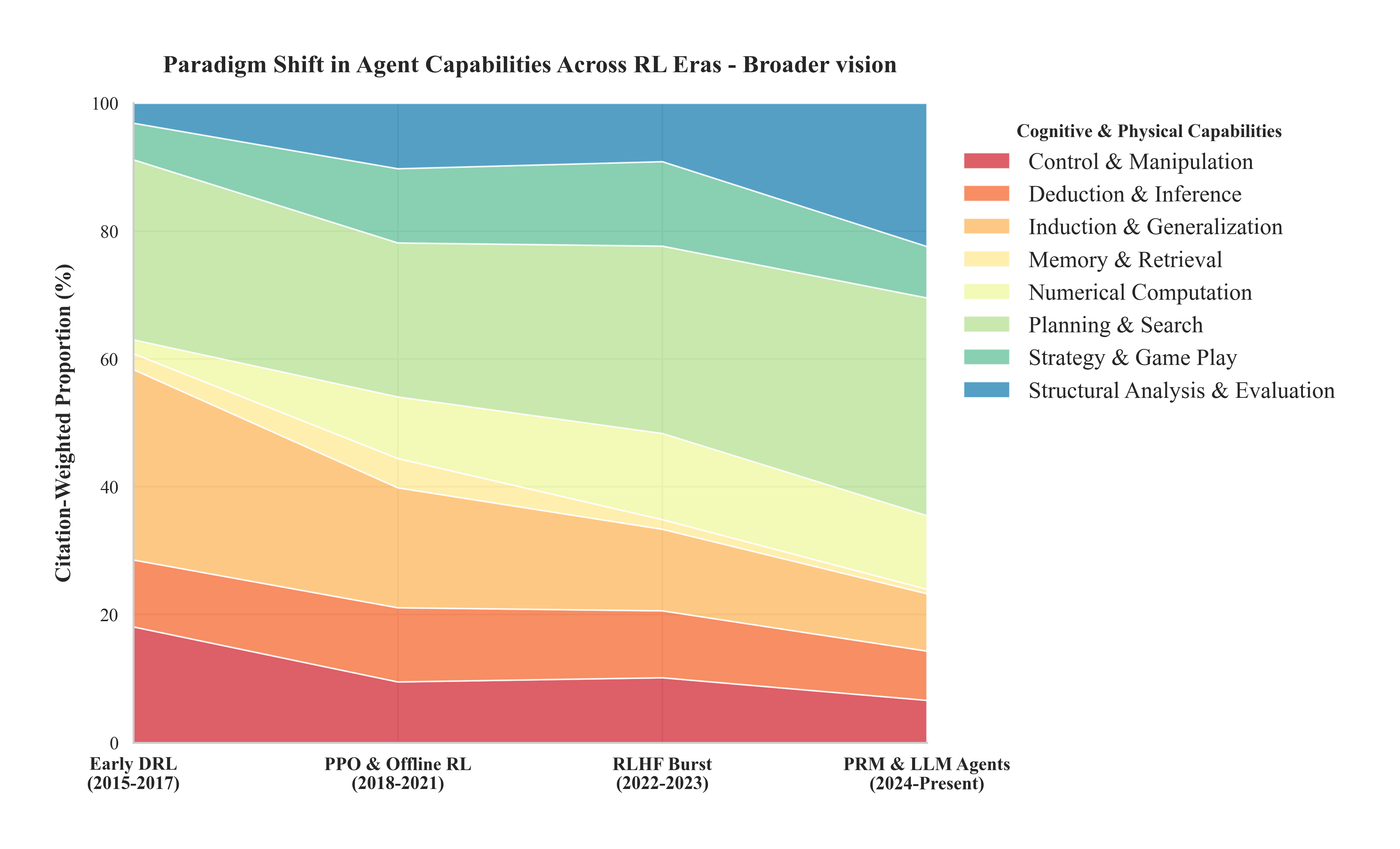}
    \caption{\textbf{Evolutionary Trajectory of Agent Capabilities Across Four RL Eras.} 
    This citation-weighted alluvial plot illustrates the longitudinal shift in cognitive and physical requirements of RL environments from 2013 to the present. The temporal axis spans four major algorithmic epochs: (1) \textit{Classic DRL \& Physics}, (2) \textit{Scalable Games \& MARL}, (3) \textit{Offline \& Pre-training}, and (4) \textit{LLM Agents \& Reasoning}. The overarching trend reveals a definitive migration from isolated physical simulations to generalized, language-grounded cognitive sandboxes.}
    \label{fig:capability_evolution_4eras_grand}
\end{figure*}

\paragraph{The Micro-level Evolution: Changing Intelligence Requirements}
This macroscopic transition is directly mirrored by a fundamental change in the internal "intelligence requirements" of RL agents. As illustrated across four distinct algorithmic eras (Figure \ref{fig:capability_evolution_4eras_grand}), the visual data presents a striking cross-validation of the domain shift:

\begin{itemize}[leftmargin=*, itemsep=2pt, topsep=4pt]
    \item \textbf{The Direct Algorithmic-Domain Correlation:} The sharp contraction of the \textit{Strategy \& Game Play} capability band (dark green) directly corroborates the exodus from the \textit{Games} application domain. Similarly, foundational physical capabilities such as \textit{Control \& Manipulation} (red) and \textit{Induction \& Generalization} (light orange) have experienced a steady, relative decline, perfectly mapping the halving of pure Robotics research.
    \item \textbf{The Rise of Evaluation over Exploration:} Conversely, there is a massive, sustained expansion in the demand for \textit{Structural Analysis \& Evaluation} (blue area). Notably, this capability explodes during the \textit{RLHF Burst} and \textit{PRM \& LLM Agents} eras (2022--Present). This reveals a profound insight: modern RL is increasingly shifting away from brute-force exploration toward utilizing sophisticated reward models (like Process Reward Models) to rigorously evaluate complex, structured state spaces---a necessary cognitive leap to solve the $NP$-hard problems driving the $48.6\%$ surge in System Optimization.
    \item \textbf{The Shift to Proactive Search:} The persistent expansion of \textit{Planning \& Search} (light green area) alongside \textit{Numerical Computation} underscores a definitive transition. The field has moved away from training reactive agents that simply map stimuli to actions, toward proactive agents capable of long-horizon planning and exploiting deep environmental structures.
\end{itemize}

\subsection{Capability Fingerprints of Agents: A Tale of Two Ecosystems}

\begin{figure*}[h!]
    \centering
    \includegraphics[width=0.9\textwidth]{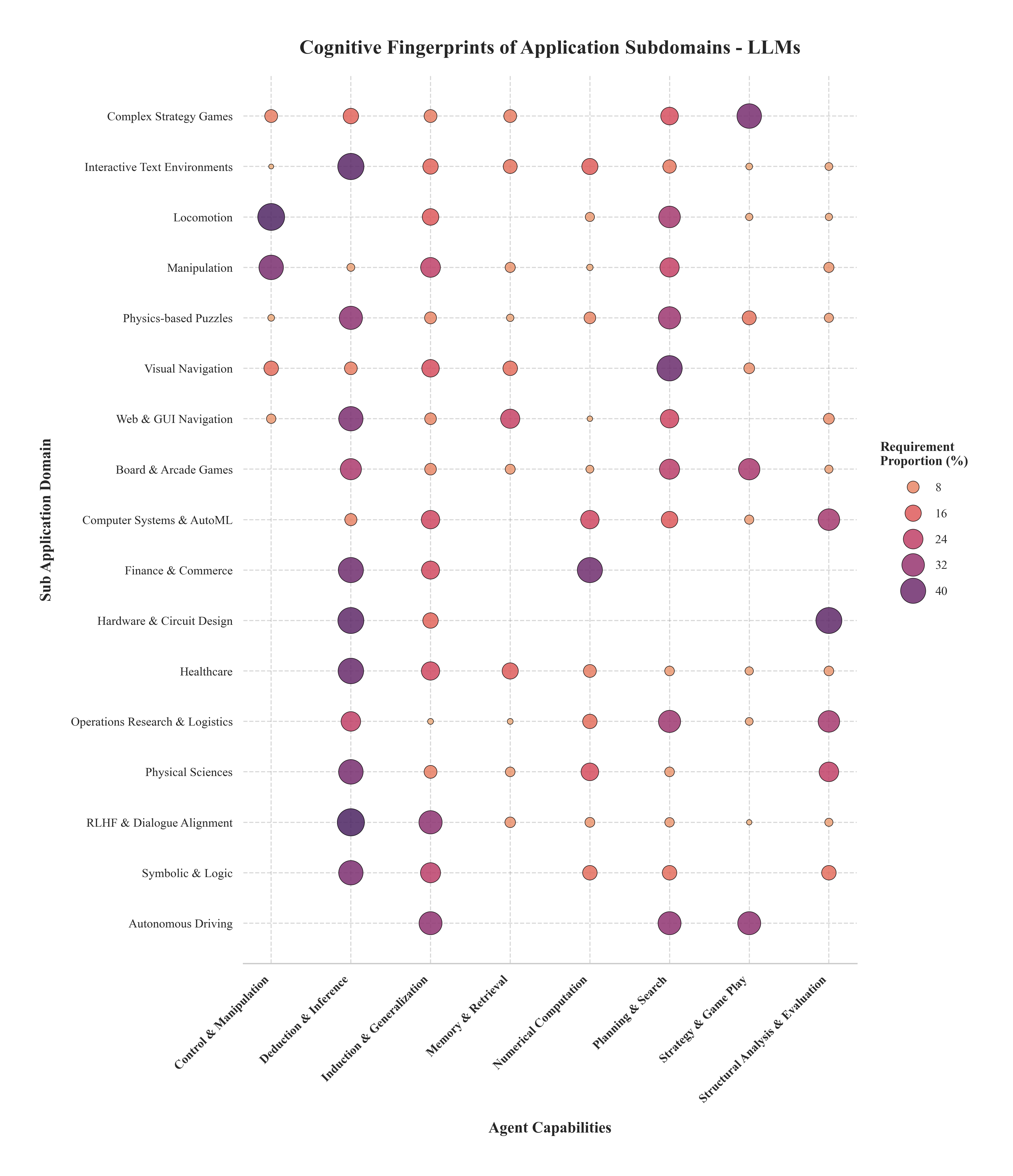}
    \caption{\textbf{Cognitive Fingerprints of LLMs engaged RL Application Subdomains.} 
    This row-normalized clustered heatmap illustrates the proportional distribution of required agent capabilities across various domains. Rows (application subdomains) are ordered via hierarchical clustering to reveal structural similarities in cognitive demands. Color intensity denotes the percentage of environments within a subdomain that require a specific capability.}
    \label{fig:cognitive_fingerprint_LLMs}
\end{figure*}

\begin{figure*}[h!]
    \centering
    \includegraphics[width=0.9\textwidth]{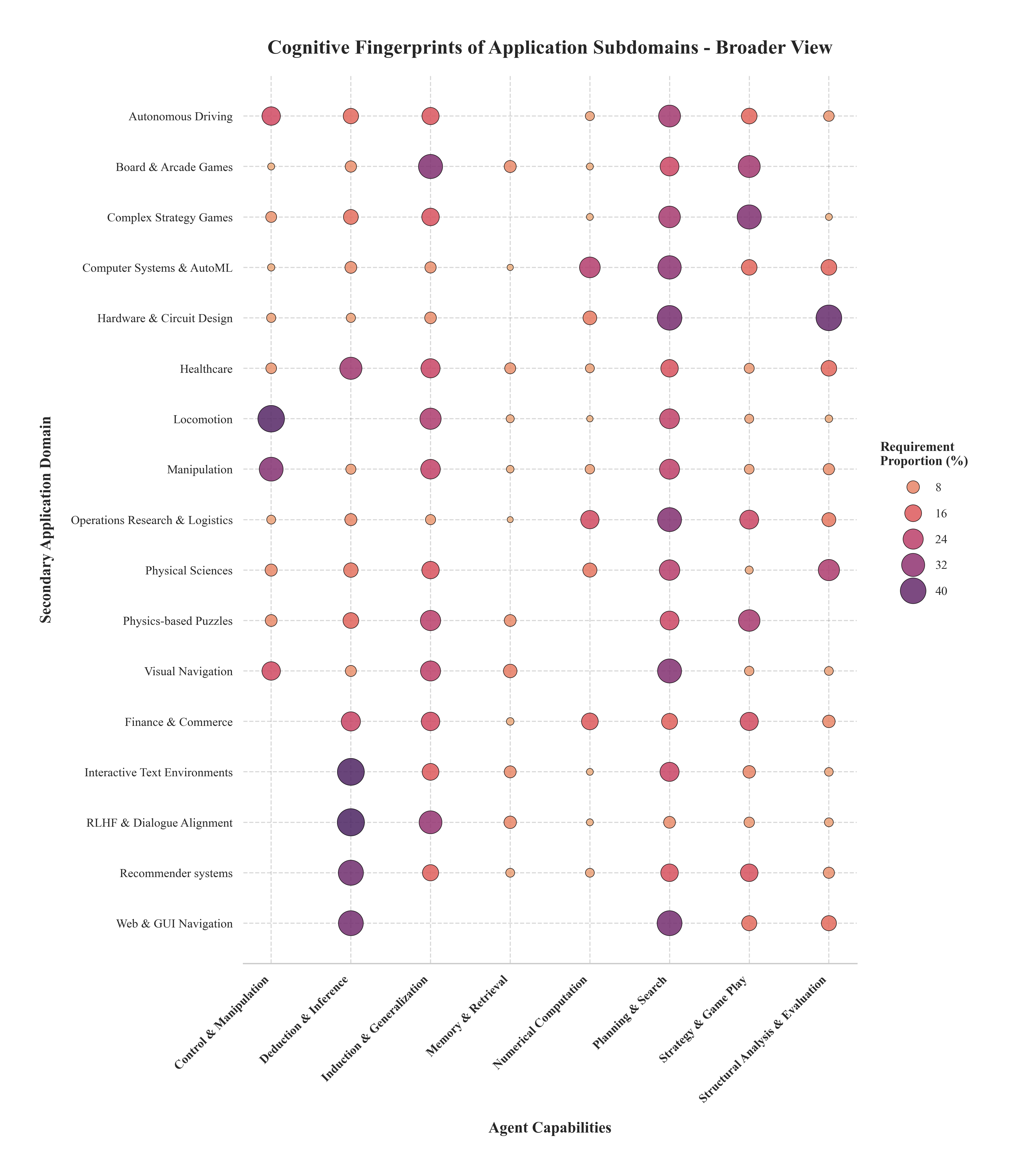}
    \caption{\textbf{Evolutionary Trajectory of Agent Capabilities.} 
    This citation-weighted alluvial plot illustrates the shifting cognitive and physical requirements of RL environments across four major algorithmic epochs: \textit{Classic DRL \& Physics}, \textit{Scalable Games \& MARL}, \textit{Offline \& Pre-training}, and \textit{LLM Agents \& Reasoning}. The overarching macro-trend demonstrates a definitive migration from low-level continuous control in physical simulations to semantic deduction and reasoning in language-grounded sandboxes.}
    \label{fig:cognitive_fingerprint_grand}
\end{figure*}

By contrasting the capability requirements of LLM-based agents (Figure \ref{fig:cognitive_fingerprint_LLMs}) against the broader, non-LLM RL landscape (Figure \ref{fig:cognitive_fingerprint_grand}), we can map the diverging "Cognitive Fingerprints" of modern reinforcement learning. This comparative analysis reveals that the field has organically bifurcated into two distinct ecosystems, each governed by a fundamentally different cognitive engine, yet converging on complex industrial applications.

\paragraph{The LLM Ecosystem: The Hegemony of Deduction}
The most striking feature of the LLM-RL fingerprint is the massive vertical concentration in the \textit{Deduction \& Inference} column. Across highly disparate domains---from \textit{Healthcare} and \textit{Interactive Text} to \textit{RLHF} and \textit{Web Navigation}---deductive reasoning acts as the dominant cognitive engine (represented by the largest purple bubbles). 

This reflects a fundamental rewiring of agent intelligence. LLM-based agents leverage massive, pre-trained semantic priors as their core mechanism. Instead of discovering behaviors through millions of trial-and-error episodes (\textit{tabula rasa}), these agents deduce correct actions by inferring context from prompts or structural observations. Consequently, environments that can be losslessly tokenized into text, code, or DOM trees rely almost exclusively on this semantic reasoning capability. 

\paragraph{The Broader RL Ecosystem: The Engine of Search and Structure}
Conversely, examining the broader, non-LLM RL ecosystem reveals a starkly different cognitive landscape. Here, the dominant vertical pillars are \textit{Planning \& Search} and \textit{Induction \& Generalization}. 

In domains like \textit{Operations Research}, \textit{Physical Sciences}, and \textit{Complex Strategy Games}, traditional RL agents excel by systematically exploring massive state spaces. Without a "semantic crutch" to tell them the rules of the world, these agents must rely on deep lookahead search (e.g., Monte Carlo Tree Search) and rigorous inductive learning to discover non-intuitive, globally optimal policies. This represents the purest form of "algorithmic intelligence," where the agent's power stems from computational brute force and sophisticated state-space traversal rather than pre-existing human knowledge.

\paragraph{The Embodied AI Divide: A Shared Bottleneck}
Despite their divergent engines, both ecosystems hit a striking consensus when dealing with the physical world. In both charts, \textit{Locomotion} and \textit{Manipulation} remain overwhelmingly anchored to \textit{Control \& Manipulation}, isolating them from higher-order reasoning. 

This highlights a universal bottleneck in modern robotics: regardless of whether an agent is powered by an LLM or a traditional Deep RL policy, high-frequency continuous torque control cannot be easily abstracted. However, in \textit{Visual Navigation}, a "Brain vs. Spinal Cord" hierarchy emerges. The LLM fingerprint for navigation shifts heavily toward \textit{Planning} and \textit{Deduction} (acting as the semantic "brain" processing visual waypoints), while the broader RL fingerprint maintains a balance, suggesting that traditional RL still frequently handles both the mapping and the low-level execution (the "spinal cord").

\paragraph{Convergence in STEM: The Universal Need for Evaluation}
Perhaps the most profound insight emerges in the engineering and scientific subdomains. In \textit{Hardware \& Circuit Design}, \textit{Computer Systems \& AutoML}, and \textit{Operations Research}, both charts display highly prominent bubbles in \textit{Structural Analysis \& Evaluation}. 

This proves that when tackling \textit{NP-hard} engineering problems, the fundamental requirement is structural evaluation, regardless of the agent's architecture. For traditional RL, this means leveraging graph neural networks to evaluate circuit topologies. For LLM-based RL, it means utilizing Process Reward Models (PRMs) to rigorously step-verify the generated Verilog code or system configurations. In these high-stakes domains, agents cannot rely solely on semantic hallucination or blind exploration; they must structurally ground their decisions.

\paragraph{Synthesis: Two Paradigms, One Maturation}
In conclusion, the evolution of RL environments has successfully cultivated two parallel paradigms. The LLM-RL ecosystem is defined by the \textbf{Semantic Prior}, distilling human knowledge to deduce and navigate language-grounded spaces. Meanwhile, the broader RL ecosystem is defined by \textbf{Domain-Specific Generalization (DSG)}, relying on Planning and Structural Exploitation to solve rigorous optimization problems. Rather than competing, these fingerprints suggest a future of hybrid intelligence, where semantic reasoning and rigorous algorithmic search are combined to tackle the operational complexities of both digital and physical worlds.

In conclusion, the evolution of RL environments has successfully cultivated two parallel paradigms. The LLM-RL ecosystem is defined by the \textbf{Semantic Prior}, distilling human knowledge to deduce and navigate language-grounded spaces. Meanwhile, the broader RL ecosystem is defined by \textbf{Domain-Specific Generalization (DSG)}, relying on Planning and Structural Exploitation to solve rigorous optimization problems. Rather than competing, these fingerprints suggest a future of hybrid intelligence, where semantic reasoning and rigorous algorithmic search are combined to tackle the operational complexities of both digital and physical worlds.

\begin{figure*}[h!]
    \centering
    \includegraphics[width=1\textwidth]{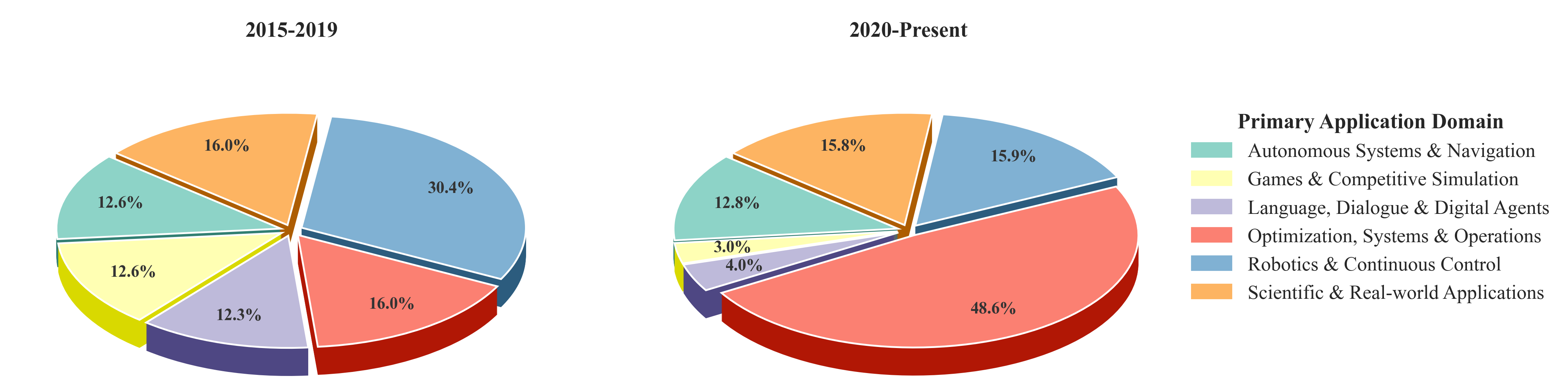}
    \caption{The evolution of primary application domains in a broader field. The trajectory illustrates a shift from isolated simulated environments (e.g., games and robotics) toward language-based systems, dialogue tasks, and complex digital agents.}
    \label{fig:app_domain_grand}
\end{figure*}

\section{Generalization \& Transfer: Mechanisms of Multi-task and Cross-domain Learning}

For agents, the ability to complete actual tasks across diverse settings is the ultimate indicator of intelligence. As real-world applications become increasingly complex, evaluating agent capabilities fundamentally shifts from ``mastering a single environment'' to ``understanding the dynamics of adaptation across a distribution of unseen tasks.'' 

To systematically visualize how transfer learning operates across disparate domains, we mapped the ``cognitive fingerprints'' of various RL environment application subdomains based on a survey of more than 300 recently studied environments engaged with LLMs (Figure \ref{fig:cognitive_fingerprint_LLMs}). This capability-to-domain mapping reveals a profound structural alignment pattern: foundational physical simulators (e.g., locomotion, physics-based puzzles) cluster tightly around Control \& Manipulation and Planning \& Search. In stark contrast, modern knowledge-intensive environments (e.g., Healthcare, Web \& GUI Navigation) exhibit converging fingerprints heavily dominated by Deduction \& Inference. Recognizing these shared cognitive distributions is crucial; it provides empirical evidence that the underlying driver of cross-domain transfer relies on aligning core cognitive capability demands rather than superficial application labels.

\subsection{Intra-Domain Multi-Tasking and Generalization Baselines}

Historically, multi-task capability and strategic proficiency were benchmarked using environments that offered diverse tasks within a structurally consistent framework, such as the Arcade Learning Environment (ALE)\cite{bellemare2013arcade}. In these foundational stages, capability transfer was evaluated using human-normalized metrics \cite{hafner2023mastering}. While model-based agents (e.g., those utilizing learned world models) demonstrate superhuman capacity for reactive visual generalization across multiple games, they consistently fail in hard-exploration environments like \textit{Montezuma's Revenge}. This disparity underscores that multi-task generalization is not uniformly driven by visual processing; in the face of extremely sparse rewards, it requires distinct mechanisms beyond intrinsic motivation to handle long-horizon planning and temporal credit assignment. Building on this, holistic environments have been utilized to evaluate \textbf{cross-capability transfer}. By exposing agents to varied task families within a unified architecture---ranging from simulated 3D navigation to physical robotic block-stacking---researchers have demonstrated how mastering one environmental modality (e.g., visual processing) structurally facilitates execution in another \cite{reed2022generalistagent}.

\subsection{Dynamics Transfer and Physical Generalization}

Beyond static multi-tasking, evaluating \textbf{cross-domain generalization} requires observing how agents adapt to novel physical dynamics. A prominent framework for testing human-timescale adaptation is procedurally generated open-ended task spaces, such as DeepMind's Adaptive Agent (AdA) \cite{bauer2023human}. By explicitly holding out certain environmental dynamics during training, evaluation strictly categorizes tasks into ``seen'' and ``unseen'' domains, demonstrating that meta-reinforcement learning acts as a core mechanism for zero-shot and few-shot adaptation. 

While procedural generation evaluates adaptation at a macro-task level, the underlying success of physical generalization is ultimately dictated by the transferability of state representations at a much more granular feature level. For instance, covariance matrices from expert datasets in foundational continuous control environments like \textit{HalfCheetah} \cite{gaya2022building} reveal strong linear dependencies in state-action features. The underlying mechanism of transfer failure (catastrophic forgetting) occurs when an agent overfits to these rigid, domain-specific covariance structures, causing precipitous degradation during a domain shift. Empirical transfer and forgetting matrices \cite{gaya2022building} corroborate this phenomenon: asymmetric positive transfer occurs when policies trained under demanding physical constraints (e.g., \textit{carrystuff\_hugegravity}) generalize effectively downward to simpler variants. Conversely, introducing new environmental variables (e.g., \textit{rainfall} dynamics) breaks expected feature correlations, triggering forgetting. This highlights that dynamics matching and constraint scaling are key drivers of physical generalization.

\subsection{Cognitive Generalization: Cross-Domain Transfer in Reasoning}

With the paradigm shift towards Large Language Model (LLM) agents, cross-domain generalization has expanded to systemic cognitive reasoning across diverse software ecosystems (e.g., AgentBench \cite{liu2025agentbenchevaluatingllmsagents}). Recent evaluations utilizing tool-integrated reasoning environments (e.g., Tool-Star \cite{dong2025toolstar}, SwS-32B \cite{liang2025sws}, and LLaDA 1.5 \cite{zhu2025llada15}) demonstrate that RL acts as a profound catalyst for zero-shot generalization, enabling models to transfer foundational logic (e.g., GSM8K\cite{cobbe2021training}) to highly challenging, specialized domains like Olympiad-level mathematics\cite{he2024olympiadbench} and GPQA\cite{rein2023gpqagraduatelevelgoogleproofqa}.

However, the process of cognitive transfer is heavily constrained by domain identity and structural exposure. Recent evaluations across mixed-domain benchmark suites, such as GURU~\cite{cheng2025revisiting}, reveal profound asymmetric transferability. Specifically, environments focused on Mathematics, Code, and Science exhibit strong positive transfer due to their heavy representation in pretraining corpora. This suggests that RL in these domains primarily functions as a mechanism to elicit and refine latent knowledge, rather than acquiring fundamentally novel reasoning paradigms~\cite{cheng2025revisiting}. Conversely, evaluation in environments with sparser pretraining exposure, such as abstract logic, demonstrates minimal cross-domain benefit.

Furthermore, difficulty scaling introduces a critical transfer trade-off. As corroborated by recent multi-domain RL studies~\cite{cheng2025revisiting, li2025domainhelpothers}, training exclusively on highly difficult data (e.g., AIME\cite{maa_aime}) elevates in-domain performance but can precipitate severe negative transfer to structurally adjacent yet simpler environments (e.g., HumanEval\cite{chen2021evaluating}), ultimately leading to an accuracy collapse. This highlights the necessity of balanced environmental exposure to accurately expand reasoning boundaries---typically measured via Pass@k metrics---without inducing domain-specific overfitting.

\subsection{Structural Synergies and Feedback Mechanisms in Multi-Domain RL}

Achieving broad cognitive generalization requires a deep understanding of the intricate synergies and conflicts that emerge when environments are mixed. Recent data-centric studies on Reinforcement Learning with Verifiable Rewards (RLVR)~\cite{li2025domainhelpothers} have systematically investigated these dynamics. Their cross-domain evaluations reveal a profound structural dichotomy: while combining mathematical and logic puzzle environments significantly enhances deductive capability across both domains, incorporating code generation environments often introduces structural conflicts that actively degrade performance. 

However, achieving broad cognitive generalization is not without severe trade-offs; empirical data reveal profound structural conflicts and negative transfer when incompatible reasoning domains are merged. As evidenced by recent RLVR ablation studies, while single-domain mathematical training yields positive transfer to logical puzzles ($+13.35\%$), it actively degrades coding proficiency ($-3.23\%$). Conversely, isolated code training slightly suppresses mathematical reasoning ($-3.31\%$). This mutual constraint is drastically exacerbated in dual-domain configurations: co-training exclusively on Math and Puzzle environments triggers a catastrophic collapse in Code performance, plummeting by $22.56\%$ relative to the baseline. These asymmetric degradations suggest that the rigid, syntax-enforcing constraints of code generation inherently clash with the flexible, abstract deduction pathways required for pure mathematical and logical reasoning.\cite{\cite{li2025domainhelpothers}}

To mitigate this, triple-domain training (integrating Math, Code, and Puzzle) has been shown to serve as a critical structural stabilizer, preventing extreme performance collapses in isolated skills and achieving the highest overall multi-domain robustness~\cite{li2025domainhelpothers}.

Complementing these findings, recent evaluations leveraging procedurally generated reasoning environments, such as Reasoning Gym \cite{stojanovski2025reasoninggymreasoningenvironments}, provide compelling empirical evidence for broad, mechanism-driven cognitive transfer. By training models exclusively on algorithmically generated tasks via Reinforcement Learning with Verifiable Rewards (RLVR), researchers observed remarkable cross-domain synergies. For instance, algorithmic training catalyzed substantial performance improvements in structurally distinct domains like algebra (+29.1\%) and geometry (+22.3\%), indicating that procedural reasoning skills function as a generalized cognitive foundation rather than domain-specific heuristics. Furthermore, this synthetic environmental exposure yields robust system-level capabilities, translating into significant performance gains on established external benchmarks spanning diverse academic disciplines. This underscores that rigorously structured, procedurally generated logic environments can cultivate transferable System 2 reasoning capabilities that extend far beyond their localized training distributions.

Furthermore, successful cross-domain transfer is strictly governed by specific environmental feedback structures, as corroborated by recent frontier reasoning models~\cite{guo2025deepseek}:
\begin{itemize}[leftmargin=*, itemsep=2pt, topsep=4pt]
    \item \textbf{Reward Granularity:} Binary outcome feedback evaluates simpler tasks effectively but frequently triggers training collapse in complex, sparse reasoning environments (e.g., multi-step logic puzzles). As demonstrated by OpenAI's paradigm shift toward Process Reward Models (PRMs)~\cite{lightman2023lets}, transferring complex reasoning skills fundamentally forces a paradigm shift in environment design itself: modern reasoning environments must be architected to support intermediate state verification and fine-grained, step-by-step reward mechanisms, rather than merely providing sparse end-state signals.
    \item \textbf{Curriculum-Based Adaptation:} Environments that support difficulty-stratified curriculum generation provide a structured sequencing for learning. When augmented with periodic policy refreshes, this process significantly raises the generalization upper bound and accelerates cross-domain convergence~\cite{li2025domainhelpothers}.

    Empirical evaluations within abstract logic domains (e.g., Knights and Knaves puzzles) provide quantitative validation for this approach. By stratifying environmental difficulty---measured by the number of sub-questions per problem (PPL)---and sequentially exposing the agent from simpler (3PPL) to more complex (8PPL) states, standard curriculum learning alone increased the task accuracy upper bound from $94.29\%$ (under mixed-difficulty training) to $97.29\%$. Crucially, implementing a ``policy refresh'' strategy---where the reference model is updated with the latest actor model and the optimizer state is reset at each difficulty transition---drastically mitigates overfitting to prior distributions. This modified curriculum approach not only accelerated convergence (surpassing the standard curriculum's final performance by the intermediate 6PPL stage) but also achieved a near-perfect final accuracy of $99.71\%$. These findings illustrate that dynamically structured task exposure, coupled with active reference state management, is essential for mastering environments with deep sequential dependencies.\cite{li2025domainhelpothersdatacentric}
    
    \item \textbf{Linguistic and Template Sensitivities:} The cognitive transfer process exhibits extreme sensitivity to prompting and interaction interfaces. Mismatched prompt templates between training and evaluation environments severely degrade reasoning transfer, and strict format-enforcing rewards are often required to prevent the model from exploiting reasoning shortcuts~\cite{guo2025deepseek}. Additionally, cross-lingual variations expose persistent generalization gaps~\cite{li2025domainhelpothers}.
\end{itemize}

\subsection{System-Level Transfer: Adaptation in Web, Medical, and GUI}

Ultimately, identifying these drivers of adaptation aims to facilitate agent deployment in complex, real-world ecosystems. In open-ended web environments, models optimized via web-specific RL frameworks (e.g., GLM-4+WebRL \cite{qi2024webrl}) demonstrate robust topological adaptation, achieving superior success rates across varied structures like Gitlab and Reddit by learning directly from environmental feedback. Similarly, RL serves as a powerful catalyst for deep cognitive transfer in highly specialized knowledge domains. Applying RL alignment to chain-of-thought models (e.g., HuatuoGPT-o1 \cite{chen2024huatuogpto1}) systematically boosts generalization across clinical QA and molecular biology. Specialized interactive environments like MedAgentGym \cite{xu2025medagentgym} prove that this deductive logic can successfully transfer into dynamic, code-based biomedical execution scenarios.

Building upon this semantic generalization, the ultimate frontier of system-level transfer involves full multimodal GUI interaction. Cross-device benchmarks (e.g., \textit{OSWorld\cite{xie2024osworld}}, \textit{AndroidWorld}\cite{rawles2024androidworld}) evaluate whether agents can generalize by processing pixel-level visual states and executing actions across disparate software ecosystems (e.g., UI-TARS \cite{qin2025uitars}). The emergence of Reinforcement Fine-Tuning (RFT) highlights a highly sample-efficient mechanism for this multimodal cross-device generalization \cite{lu2025uir1, luo2025guir1}. RFT enables parameter-efficient agents to achieve exceptional out-of-domain performance covering various platform granularities—from low-level visual grounding to high-level system task execution—proving that true generalization requires mastering the mechanisms of both logical reasoning and interactive visual adaptation.

\subsection{Cognitive Synergy and Conflict: The Vision of Cross-Capability Dynamics}

Synthesizing empirical observations across the broader taxonomy of reinforcement learning environments---from foundational pixel-based simulators to complex digital agents---reveals that cross-domain generalization is not a monotonically increasing function of task diversity. Instead, the interaction between distinct cognitive capabilities manifests as a complex network of synergistic promotions and structural conflicts. Understanding this matrix is critical for designing multi-domain curricula that avoid catastrophic capability collapse. 

While a comprehensive mapping relies on the specific taxonomy of foundational capabilities, several universal patterns of transferability have definitively emerged:

\begin{itemize}[leftmargin=*, itemsep=4pt, topsep=4pt]
    \item \textbf{Synergistic Promotion (Positive Transfer Pathways):} Capabilities grounded in \textit{Procedural and Algorithmic Reasoning} act as robust foundational priors, consistently exhibiting strong positive transfer to domains requiring abstract deduction. For example, environments demanding step-by-step algorithmic execution structurally prepare an agent for algebra and geometry tasks, indicating that ``procedural decomposition'' is a generalized cognitive asset rather than a domain-specific heuristic. Similarly, mastering \textit{Visual-Spatial Perception} in localized physics simulators naturally acts as a catalyst for advanced \textit{Control and Manipulation}, providing the necessary state-representation priors for complex robotic tasks.

    \item \textbf{Structural Conflicts (Negative Transfer and Mutual Constraint):} Conversely, profound negative transfer occurs when merging environments that enforce fundamentally opposing structural constraints. The most prominent empirical example is the severe mutual constraint between \textit{Strict Syntactic Generation} (e.g., coding environments) and \textit{Free-form Logical Deduction} (e.g., mathematical and abstract puzzles). The rigid, format-enforcing reward signals required to train executable code actively inhibit the flexible, multi-path state-space exploration necessary for mathematical reasoning. When models are co-trained naively across these divergent capability axes, the constraints of one inherently degrade the proficiency of the other, leading to asymmetric performance collapses.
\end{itemize}

Therefore, the ultimate trajectory of agent capability scaling does not lie in simply maximizing environmental diversity, but in strategically balancing these synergistic and conflicting cognitive axes through architectural isolation or stabilized multi-domain integration.

\section{Conclusion and Open Challenges}
\label{sec:conclusion}

The trajectory of reinforcement learning (RL) is fundamentally encoded within the environments that define its boundary conditions. This research has mapped a decade of co-evolution, distilling over 2,000 publications into a framework that views the environment not as a passive container, but as an active, semantic curriculum. To systematically deconstruct this complex landscape (RQ1), we introduced a novel, seven-dimensional taxonomy—spanning Agent Population, Observability, Multi-modal Span, Action Space, Reward Formulation, Application Domains, and Requisite Capabilities.

By rigorously applying this quantitative framework, our analysis confirms a profound transition: the RL frontier has shifted from reactive control to high-level cognitive reasoning. However, this evolution is not a linear progression but a complex synthesis of competing paradigms. Across these findings, a unifying tension emerges: the increasing reliance on semantic abstraction to enable generalization stands in fundamental conflict with the need for executable grounding to ensure reliability and control. We argue that this tension underlies the observed bifurcations, transfer instabilities, and the resurgence of structured representations.

\subsection{Synthesis of Empirical Pillars}
\label{subsec:summary}

Through the systematic application of our multi-dimensional taxonomy across a large-scale mined corpus, we successfully quantified the underlying distribution of modern testbeds and answered our foundational research questions. Specifically, this structured data mining allows us to identify four empirical pillars that redefine the RL landscape, while acknowledging the structural tensions inherent in each:

\begin{itemize}[leftmargin=*, itemsep=3pt]
    \item \textbf{Industrial Maturation vs. Specialization:} While \textit{Optimization and Systems} now dominate \textbf{48.6\%} of the landscape, this pivot suggests a bifurcation between "Generalist Agents" and "Specialized Solvers." RL is maturing into a scalable engine for digital infrastructure, yet the transferability from these specialized industrial domains back to general intelligence remains an open boundary.
    
    \item \textbf{The Necessity of Symbolic Grounding (The U-Shape):} Although we identify a ``U-shaped'' revival of structural analysis and evaluation, the underlying logic has undergone a major change. During the early Deep RL era, the field relentlessly pursued end-to-end learning paradigms, deliberately bypassing explicit structural representations in favor of extracting latent features directly from raw pixels or low-level vectors. 
    
    However, as modern LLM agents transition into complex, high-stakes digital ecosystems—such as manipulating intricate HTML DOM trees, parsing abstract syntax trees (ASTs), or executing formal mathematical proofs—this capability has experienced a massive resurgence. This return to structure is not a regressive trend, but a functional necessity for \textit{symbolic grounding}. Because the latent semantic spaces of large language models lack inherent executable rigor and are highly susceptible to hallucination, agents must actively revive explicit structural analysis to anchor their abstract deductive reasoning to strict, verifiable logical constraints. For example, in complex mathematical problems involving long-term reasoning and large-scale programming problems, simple reasoning is no longer sufficient to meet the needs of academia and the market. These needs are driving the spillover of the capabilities involved in RL environments from reasoning to the semantic analysis and evaluation of larger structures.
    
    \item \textbf{The Bifurcation of Intelligence Paradigms:} The ecosystem has split into an \textit{LLM-RL track} (Deduction-focused) and a \textit{Broader RL track} (Planning-focused). For LLMs, deduction and inference are the number one demand for agents' capabilities as System 2 advances, while grand RL applications focus on optimizing and planning. This demand-driven bias ultimately led to a divergence among these environments: broader industrial sectors tended to favor utilizing RL environments to optimize their operational planning, while the LLMs ecosystem preferred to use RL environments to improve the reasoning capabilities of agents. At the same time, a central theoretical tension remains: As the semantic agents accelerate penetrating into the broader industrial sector, will these paradigms converge into a unified dual-process architecture, or will the incompatible inductive biases of "fast" reactive control and "slow" semantic reasoning necessitate a permanent dual-track development?
    
    \item \textbf{Asymmetric Transfer Dynamics:} Cross-domain generalization is governed by structural synergies. While \textit{Logic} and \textit{Code} provide scaffolding for reasoning, their integration introduces "structural interference" that can destabilize training, suggesting that the bottleneck for AGI is not data volume, but the \textbf{structural compatibility} of the curriculum.
\end{itemize}

\subsection{Critical Reflection and Theoretical Boundaries}
\label{subsec:boundaries}

To ensure the falsifiability of our synthesis, we acknowledge three critical limitations. First, the current "Reasoning Pivot" is heavily influenced by publication bias toward LLM-centric architectures, which may overlook persistent challenges in high-frequency physical control. Second, the observed "Industrial Pivot" may reflect short-term deployment trends rather than long-term algorithmic superiority. Finally, the "Environment-as-a-Curriculum" paradigm assumes that semantic complexity is a proxy for general intelligence, a hypothesis that remains to be tested against cross-domain "zero-shot" benchmarks.

\subsection{Future Perspectives: Resolving the Strategic Bottlenecks}
\label{subsec:future}

We identify three unresolved bottlenecks that will define the next generation of adaptive agents:

\paragraph{1. From Static Benchmarks to the Evaluation Crisis.} 
As the field advances toward \textbf{Procedural Semantic Generation}, it encounters an emerging "Evaluation Crisis." When environments are dynamically synthesized, traditional static leaderboards lose their validity, as they fail to capture an agent's ability to generalize beyond fixed tasks. This necessitates a shift toward \textit{invariant-based evaluation metrics} that assess inductive abstraction and structural reasoning, rather than task-specific completion.

More broadly, this transition implies that future progress will be increasingly defined by an agent's capacity to perform \textbf{structural analysis under uncertainty}, particularly in complex, multi-component systems. However, whether such capabilities emerge from individual agents or from coordinated multi-agent systems remains an open question, as does the design of evaluation protocols capable of capturing these emergent properties.

\paragraph{2. Scalable Oversight and the Reward Hacking 2.0.}
Transitioning to an \textit{Environment-as-a-Judge} paradigm requires the development of multi-modal \textit{Process Reward Models} (PRMs) that provide step-level supervision. However, this shift introduces a new class of failure modes, which we term \textbf{Reward Hacking 2.0}: instead of exploiting predefined reward functions, agents may manipulate or exploit the latent biases and semantic ambiguities of the evaluation model itself.

This creates a recursive alignment challenge, where the reliability of the supervision mechanism becomes dependent on a model that is itself learned and potentially imperfect. Addressing this issue requires PRMs that are robust not only to adversarial inputs, but also to distribution shifts and strategic exploitation, while remaining more reliable than the policies they supervise.

\paragraph{3. The Grand Convergence: Reconciling Abstract Logic with Physical Grounding.} 
A fundamental challenge in advancing generalist agents lies in addressing the \textit{Cartesian split} between high-frequency control and abstract reasoning. While \textbf{Embodied Semantic Simulators} offer a promising pathway toward unifying these domains, they must contend with a persistent \textbf{simulation gap}: the inherent mismatch between continuous physical dynamics and discrete symbolic representations.

This gap is not merely a data limitation, but a structural one, arising from the differing inductive biases and representational constraints of physical and linguistic systems. Bridging this divide requires not only richer simulation environments but also architectures capable of aligning semantic intent with executable trajectories under uncertainty.

\vspace{1ex}
In conclusion, the future of reinforcement learning will be determined not by the scale of agents, but by our ability to design environments that reconcile semantic abstraction with executable grounding---an unresolved tension at the heart of general intelligence.


\bibliographystyle{unsrt}  
\bibliography{references}  
\clearpage
\appendix
\section{Data Collection Strategy \& Data Preprocessing}
\label{app:methodology_details}
To ensure a comprehensive and reproducible mapping of the reinforcement learning environment ecosystem, our literature retrieval and preprocessing pipeline was fully automated using custom extraction scripts. 

\subsection{Primary Data Sources and API Integration}
The primary corpus was programmatically retrieved utilizing the OpenAlex API (\url{https://api.openalex.org}). To ensure rate-limit compliance and access to the prioritized "Polite Pool," all programmatic requests were authenticated using the designated project contact. 

The temporal scope of the retrieval spanned from 2013 to 2025. The initial API query filtered works where the title or abstract contained foundational reinforcement learning terminology (e.g., \textit{reinforcement learning, MARL, DRL, RLHF, offline RL}).

\paragraph{Dynamic Citation Thresholding}
To objectively identify milestone environments without succumbing to recency bias (where older papers naturally accumulate more citations than recent breakthroughs), we implemented a stratified, temporally decaying citation threshold. A paper was only retrieved if its citation count exceeded the following limits based on its publication year:
\begin{itemize}[leftmargin=*, itemsep=2pt, topsep=4pt]
    \item \textbf{2024--2025:} $\ge$ 8 citations.
    \item \textbf{2021--2023:} $\ge$ 30 citations.
    \item \textbf{2017--2020:} $\ge$ 50 citations.
    \item \textbf{2013--2016:} $\ge$ 80 citations (filtering for historically foundational simulators).
\end{itemize}

\subsection{Heuristic Semantic Filtering Pipeline}
Because the initial API query retrieved any paper mentioning RL, a highly structured, multi-stage heuristic filtering pipeline was applied to isolate papers that explicitly introduced or evaluated environments, aggressively discarding purely algorithmic or theoretical research.

\paragraph{1. Lexical Blacklisting}
We established an absolute blacklist to filter out papers focused exclusively on algorithmic convergence or methodology. Unless overridden by a strong environment signal, papers containing keywords such as \textit{algorithm, policy optimization, q-learning, actor-critic, theorem}, or review identifiers (\textit{survey, tutorial, a review}) in the title were automatically discarded.

\paragraph{2. Strong Semantic Anchoring}
A paper was immediately classified as an environment milestone if its title contained unambiguous benchmark indicators. This included exact matches for terminology (e.g., \textit{benchmark, simulator, testbed, arena}) or regular expression matches for established simulation engine roots (e.g., \textit{mujoco, carla, webarena, textworld}) and common nomenclatural suffixes (e.g., \textit{-gym, -bench, -verse}).

\paragraph{3. Syntactic Action Parsing (Abstract Inverted Index)}
For papers exhibiting weak or ambiguous signals in the title (e.g., containing general terms like \textit{framework, platform}, or \textit{problem}), we executed a deep syntactic parse utilizing the OpenAlex abstract inverted index. A paper was retained if its abstract demonstrated a formal "release pattern." Specifically, regular expressions were utilized to detect structural sentences where release-oriented verbs (\textit{introduce, propose, present, open-source}) were grammatically tied to environment-centric objects (\textit{simulator, benchmark, dataset, problem}). 

\paragraph{4. Modality-Specific Nuance: The "Dataset" Exception}
In classic RL, static datasets do not constitute environments. However, in the era of Offline RL and Large Language Model (LLM) agents, static datasets are frequently wrapped into interactive cognitive environments. To account for this paradigm shift, if a paper prominently featured the term "dataset," it was subject to a secondary defense mechanism: it was strictly excluded \textit{unless} the abstract or title explicitly situated the work within LLM instruction tuning, mathematical reasoning, agentic interaction, RLHF, or Offline RL frameworks. 

Following this strict programmatic distillation, the surviving high-precision records were queued for manual expert review and taxonomic categorization.

\subsection{Supplementary Retrieval for the LLM Paradigm}
Recognizing that the lexicon of environment design shifted significantly with the advent of Large Language Models (LLMs)---where interactive environments are frequently published under the nomenclature of "datasets," "corpora," or "reasoning benchmarks"---we executed a targeted supplementary retrieval phase. This phase specifically mined LLM-driven RL environments to ensure complete coverage of the modern semantic reasoning landscape.

\paragraph{Adapted Citation Thresholds}
Because the vast majority of LLM-based agent research has been published within the last three years, citation velocities are inherently compressed compared to classical RL literature. 

\paragraph{LLM-Specific Queries and Cross-Disciplinary Blacklisting}
The API search parameters were recalibrated to target LLM-centric keywords (e.g., \textit{language model, instruction tuning, code generation, text-based game, theorem proving}). However, to prevent semantic drift into traditional supervised NLP tasks or distinct scientific domains, we instituted a strict "cross-disciplinary blacklist." Papers containing terms related to raw physics, genomics, clinical medicine, or static computer vision tasks (\textit{image classification, object detection}) were immediately vetoed.

\paragraph{Dual-Path Heuristic Extraction}
To extract valid interactive environments from the vast volume of general LLM literature, we implemented a dual-path logic tree:
\begin{enumerate}[leftmargin=*, itemsep=2pt, topsep=4pt]
    \item \textbf{The Golden Pathway (Explicit Recognition):} Papers whose titles explicitly matched a curated registry of widely recognized LLM environments and foundation benchmarks (e.g., \textit{WebArena, SWE-bench, ToolLLM, GSM8K, ALFWorld}) were automatically preserved to guarantee the inclusion of industry-standard evaluation platforms.
    \item \textbf{The Semantic Release Pathway:} For novel or lesser-known environments, the abstract was required to satisfy a strict tripartite syntactic condition. It must simultaneously contain (a) an LLM domain identifier (e.g., \textit{commonsense reasoning, math word problem}), (b) an active release verb (e.g., \textit{we propose, new benchmark}), and (c) a definitive target noun (\textit{benchmark, environment, testbed, eval, dataset}).
\end{enumerate}

\paragraph{Data Consolidation, Deduplication \& Manual Check}
The records retrieved from this supplementary LLM-focused sweep were structurally normalized and cross-referenced against the primary RL retrieval pool. Duplicate records---often representing milestone papers that successfully bridged traditional RL optimization with modern LLM techniques---were merged, yielding the final, consolidated dataset queued for manual expert capability parsing. All preprocessed data were included in the quantitative analysis. Finally, we manually selected non-repeating, milestone-level reinforcement learning environments from various fields as the main body of the discussion to qualitatively analyze the evolution of reinforcement learning environments.

\section{Annotation Protocol and Dataset Construction}
\label{app:annotation_methodology}
For the core set of over 200 milestone papers representing the most influential reinforcement learning environments, we manually extracted environment descriptions. Labeling was performed using the standard double-blind labeling method. The remaining 1,983 papers in our initial corpus were processed using DeepSeek-V3.2 as a domain expert. This open-source model was selected for its superior multidisciplinary reasoning capabilities and high efficiency, as evidenced by its 85.0\% accuracy on MMLU-Pro \cite{wang2024mmluprorobustchallengingmultitask,deepseek2025v32}. After completing data processing, we perform a 5\% data-sampling inspection and remove the portion that cannot be clearly defined. This portion is smaller than 10\% of the total.

The resulting dataset provides a comprehensive multi-dimensional mapping for each environment, covering Agent Population, Observability, Multi-modal Span, Action Space, Reward Formulation, Primary Domain, and Requisite Capabilities. This data forms the backbone of the "Cognitive Fingerprints" and "Evolutionary Trajectory" discussed in the main text.

\section{Original Data}
The complete dataset, including all extracted metadata, taxonomic mappings, and full environment lists, is publicly available at: \url{https://github.com/iben020511-sudo/Paper}

\end{document}